\documentclass[10pt,twocolumn,letterpaper]{article}

\usepackage{cvpr}              % To produce the CAMERA-READY version
\usepackage{graphicx}
\usepackage{amsmath}
\usepackage{amssymb}
\usepackage{booktabs}
\usepackage{balance}

\usepackage[pagebackref,breaklinks,colorlinks]{hyperref}
\usepackage[capitalize]{cleveref}
\crefname{section}{Sec.}{Secs.}
\Crefname{section}{Section}{Sections}
\Crefname{table}{Table}{Tables}
\crefname{table}{Tab.}{Tabs.}

\newcommand{\myparagraph}[1]{\vspace{0.05in}\noindent\textbf{#1}}

\usepackage[dvipsnames]{xcolor}
\definecolor{PurpleColor}{rgb}{0.5,0,0.7}
\definecolor{OrangeColor}{rgb}{0.914,0.541,0.0.141}
\definecolor{GreenColor}{rgb}{0.137,0.573,0.565}

\newcommand{\smpl}{\mbox{SMPL}\xspace}
\newcommand{\smplx}{\mbox{SMPL-X}\xspace}

\newcommand{\modelname}{BEDLAM\xspace}

\newcommand{\numBodies}{271\xspace}
\newcommand{\numBoxes}{{1M}\xspace}
\newcommand{\numClothTextures}{1691\xspace}
\newcommand{\numSkinTextures}{100\xspace}
\newcommand{\numOutfits}{111\xspace}
\newcommand{\numWigs}{27\xspace}
\newcommand{\numMotions}{2311\xspace}
\newcommand{\numFrames}{{380K}\xspace}
\newcommand{\numMen}{109\xspace}
\newcommand{\numWomen}{162\xspace}
\newcommand{\group}{{3-10}\xspace}
\newcommand{\avgPeople}{{1-10}\xspace}
\newcommand{\numScenes}{8\xspace}
\newcommand{\numHDRI}{95\xspace}

\newcommand{\numClips}{10K\xspace}
\newcommand{\numTrain}{750K\xspace}
\newcommand{\numValid}{200K\xspace}
\newcommand{\numTest}{50K\xspace}

\newcommand{\fps}{30\xspace}

\def\cvprPaperID{5126} % *** Enter the CVPR Paper ID here
\def\confName{CVPR}
\def\confYear{2023}

\begin{document}

\title{BEDLAM: A Synthetic Dataset of\\ Bodies Exhibiting Detailed Lifelike Animated Motion }

\author{
Michael J. Black$^{1,}$\footnotemark[1] 
\quad Priyanka Patel$^{1,}$\footnotemark[1]
\quad Joachim Tesch$^{1,}$\footnotemark[1]
\quad Jinlong Yang$^{2,}$\footnotemark[1] $^,$\footnotemark[2] \\
 \\
 $^1$Max Planck Institute for Intelligent Systems, T{\"u}bingen, Germany \quad
 $^2$Google %(This work was done at MPI-IS.)
}

%\maketitle
\newcommand{\teaserCaption}{
{\bf BEDLAM: }
Bodies Exhibiting Detailed Lifelike Animated Motion.
}

\twocolumn[{
    \renewcommand\twocolumn[1][]{#1}
    \maketitle
    \centering
    \vspace{-0.5em}
    \begin{minipage}{1.00\textwidth}
        \centering
        \includegraphics[trim=000mm 000mm 000mm 000mm, clip=False, width=\linewidth]{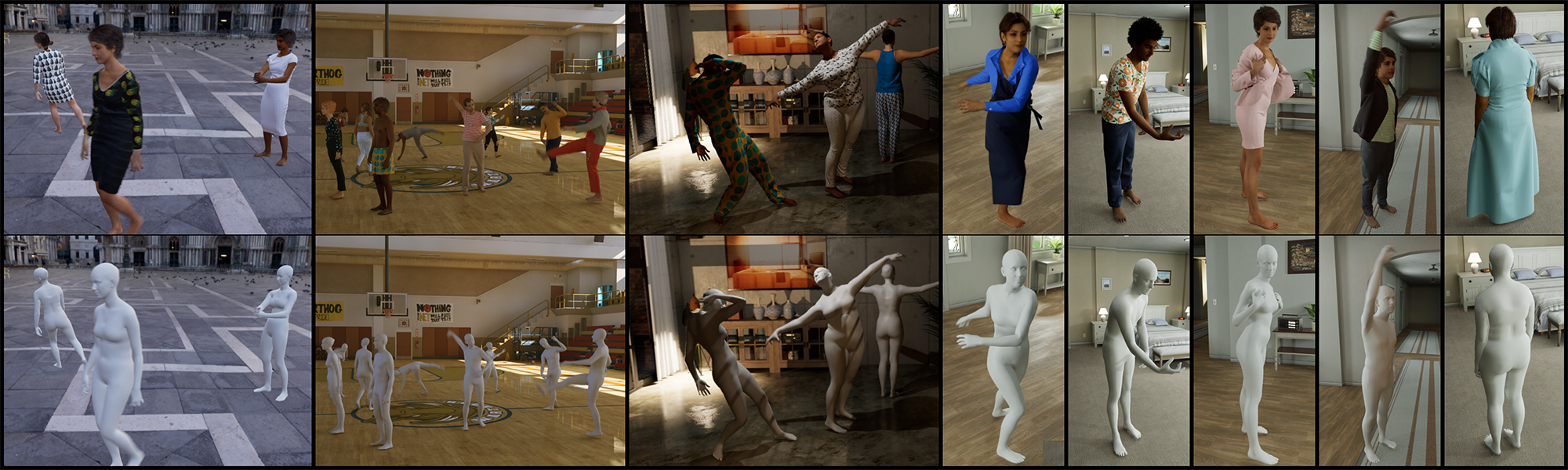}
    \end{minipage}
    \vspace{-0.5 em}
    \captionof{figure}{{\bf BEDLAM} is a large-scale synthetic video dataset designed to train and test algorithms on the task of 3D human pose and shape estimation (HPS). BEDLAM contains diverse body shapes, skin tones, and motions. Beyond previous datasets, BEDLAM has SMPL-X bodies with hair and realistic clothing animated using physics simulation. With BEDLAM's realism and scale, we find that synthetic data is sufficient to train regressors to achieve state-of-the-art HPS accuracy on real-image datasets without using any real training images.
    }
    \label{fig:teaser}
    \vspace{2.2em}
}]
%%%%%%%%% ABSTRACT
\begin{abstract}
\renewcommand{\thefootnote}{\fnsymbol{footnote}}
\footnotetext[1]{The authors contributed equally and are listed alphabetically.}
\footnotetext[2]{This work was performed when JY was at MPI-IS.}
We show, for the first time, that neural networks trained only on synthetic data achieve state-of-the-art accuracy on the problem of 3D human pose and shape (HPS) estimation from real images.
Previous synthetic datasets have been small, unrealistic, or lacked realistic clothing.
Achieving sufficient realism is non-trivial and we show how to do this for full bodies in motion.
Specifically, our BEDLAM dataset contains monocular RGB videos with ground-truth 3D bodies in \smplx format. 
It includes a diversity of body shapes, motions, skin tones, hair, and clothing.
The clothing is realistically simulated on the moving bodies using commercial clothing physics simulation.
We render varying numbers of people in realistic scenes with varied lighting and camera motions.
We then train various HPS regressors using BEDLAM and achieve state-of-the-art accuracy on real-image benchmarks despite training with synthetic data.
We use BEDLAM to gain insights into what model design choices are important for accuracy.
With good synthetic training data, we find that a basic method like HMR approaches the accuracy of the current SOTA method (CLIFF). 
BEDLAM is useful for a variety of tasks and all images, ground truth bodies, 3D clothing, support code, and more are available for research purposes.
Additionally, 
we provide detailed information about our synthetic data generation pipeline, enabling others to generate their own datasets.
See the project page: \url{https://bedlam.is.tue.mpg.de/}.
\end{abstract}

%%%%%%%%% BODY TEXT
\vspace{-0.1in}
\section{Introduction}
\label{sec:intro}

The estimation of 3D human pose and shape (HPS) from images has progressed rapidly since the introduction of HMR~\cite{Kanazawa2018_hmr}, which uses a neural network to regress SMPL~\cite{SMPL:2015} pose and shape parameters from an image.
A steady stream of new methods have improved the accuracy of the estimated 3D bodies~\cite{feng2021pixie, Kolotouros2019_spin, pymaf, li2022cliff, pare,lin2021mesh, sun2022putting}.
The progress, however, entangles two things: improvements to the architecture and improvements to the training data. 
This makes it difficult to know which matters most. 
To answer this, we need a  dataset with real ground truth 3D bodies and not simply 2D joint locations or pseudo ground truth.
To that end, we introduce a new, realistic, synthetic dataset called BEDLAM (Bodies Exhibiting Detailed Lifelike Animated Motion) and use it to analyze the current state of the art (SOTA).
\cref{fig:teaser} shows example images from BEDLAM along with the ground-truth SMPL-X \cite{Pavlakos2019_smplifyx} bodies.

Theoretically, synthetic data has many benefits.
The ground truth is ``perfect" by construction, compared with existing image datasets.
We can ensure diversity of the training data across skin tones, body shapes, ages, etc., so that HPS methods are inclusive.
The data can also be easily repurposed to new cameras, scenes, and sensors.
% (IR, range, etc.).
Consequently, there have been many attempts to create synthetic datasets to train HPS methods. 
While prior work has shown synthetic data is useful, it has not been sufficient so far.
This is likely due to the lack of realism and diversity in existing synthetic datasets.

In contrast, BEDLAM provides the realism necessary to test whether
``synthetic data is all you need".
Using BEDLAM, we evaluate different network architectures, backbones, and training data and find that {\em training only using synthetic data} produces methods that generalize to real image benchmarks, obtaining SOTA accuracy on both 3D human pose and 3D body shape estimation.
Surprisingly, we find that even basic methods like HMR \cite{Kanazawa2018_hmr}  achieve SOTA performance on real images when trained on BEDLAM.

{\bf Dataset.} 
BEDLAM contains monocular RGB videos together with ground truth 3D bodies in \smplx format.
To create diverse data, we use \numBodies body shapes (\numMen men and \numWomen women), with \numSkinTextures skin textures from Meshcapade \cite{meshcapade} covering a wide range of skin tones.
In contrast to previous work, we add \numWigs different types of hair  (Reallusion \cite{reallusion}) to the head of \smplx.
To dress the body, we hired a professional 3D clothing designer to make \numOutfits outfits, which we drape and simulate on the body using CLO3D \cite{clo3d}.
We also texture the clothing using \numClothTextures artist-designed textures \cite{wowpatterns}.
The bodies are animated using \numMotions motions sampled from AMASS \cite{AMASS:ICCV:2019}.
Because AMASS does not include hand motions, we replace the static hands with hand motions sampled from the GRAB dataset \cite{GRAB:2020}.
We render single people as well as groups of people (varying from \group) moving in a variety of 3D scenes (\numScenes) and HDRI panoramas (\numHDRI). 
We use a simple method to place multiple people in the scenes so that they do not collide and use simulated camera motions with various focal lengths.
The synthetic image sequences are rendered using  Unreal Engine 5   \cite{unreal} at \fps fps with motion blur.
In total, BEDLAM contains around \numFrames unique image frames with \avgPeople people per image, for a total of \numBoxes unique bounding boxes with people.

We divide BEDLAM into training, validation, and test sets with 75\%, 20\% and 5\% of the total bounding boxes respectively.  
While we make all the image data available, we withhold the SMPL-X ground truth from the test set and provide an automated evaluation server.
For the training and validation sets, we provide all the SMPL-X animations, the 3D clothing, skin textures, and all freely available assets. 
Where we have used commercial assets, we provide information about how to obtain the data and replicate our results.
We also provide the details necessary for researchers to create their own data.

{\bf Evaluation.}
With sufficient high-quality training data, fairly simple neural-network architectures often produce SOTA results on many vision tasks.
Is this true for HPS regression?
To tackle this question, we train two different baseline methods (HMR \cite{Kanazawa2018_hmr} and CLIFF \cite{li2022cliff}) on varying amounts of data and with different backbones; HMR represents the most basic method and CLIFF the recent SOTA.
Since BEDLAM provides
paired images with \smplx parameters, we train methods to directly regress these parameters; this simplifies the training compared with methods that use 2D training data.
We evaluate on natural-image datasets including 3DPW~\cite{vonMarcard2018} and RICH~\cite{Huang:CVPR:2022}, a laboratory dataset (Human3.6M \cite{ionescu2013human36m}), as well as two datasets that evaluate body shape accuracy (SSP-3D \cite{sengupta2020straps} and HBW \cite{Shapy:2022}).

Surprisingly, despite its age, we find that training HMR on synthetic data produces results on 3DPW that are better than many recently published results and are close to CLIFF.
We find that the backbone has a large impact on accuracy, and pre-training on COCO is significantly better than pre-training on ImageNet or from scratch.
We perform a large number of experiments in which we train with just synthetic data, just real data, or synthetic data followed by fine tuning on real data.
We find that there is a significant benefit to training on synthetic data over real data and that fine tuning with real data offers only a small benefit.

A key property of BEDLAM is that it contains realistically dressed people with ground truth body shape.
Consequently, we compare the performance of methods trained on BEDLAM with 
two SOTA methods for body shape regression: SHAPY \cite{Shapy:2022}  and Sengupta et al.\cite{sengupta2021hierarchicalICCV} using both the HBW and SSP-3D datasets.
CLIFF trained with BEDLAM does well on both datasets, achieving the best overall of all methods tested.
This illustrates how methods trained on BEDLAM generalize across tasks and datasets.

{\bf Summary.} We propose a large synthetic dataset of realistic moving 3D humans. We show that training on synthetic dataset alone, even with a basic network architecture, produces accurate 3D human pose and shape estimates on real data. 
BEDLAM enables us to perform an extensive meta-ablation study that illuminates which design decisions are most important.
While we focus on HPS, the dataset has many other uses in learning 3D clothing models and action recognition.
BEDLAM is available for research purposes together with an evaluation server and the assets needed to generate new datasets.

\section{Related work}
\label{sec:related}

There are four main types of data used to train HPS regressors:
(1) Real images from constrained scenarios with high-quality ground truth (lab environments with motion capture).
(2) Real images in-the-wild with 2D ground truth (2D keypoints, silhouettes, etc.).
(3) Real images in-the-wild with 3D pseudo ground truth (estimated from 2D or using additional sensors).
(4) Synthetic images with perfect ground truth.
Each of these has played an important role in advancing the field to its current state.
The ideal training data would have perfect ground truth 3D human shape and pose information together with fully realistic and highly diverse imagery.
None of the above fully satisfy this goal.
We briefly review 1-3 while focusing our analysis on 4.

\vspace{0.075in}
\noindent\textbf{Real Images.}
Real images are diverse, complex, and plentiful.
Most methods that use them for training rely on 2D keypoints, which are easy to manually label at scale \cite{lin2014coco,Andriluka14CVPR,MartinMartin21PAMI,iqbal2017poseTrack}.
Such data relies on human annotators who may not be consistent, and only provides 2D constraints on human pose with no information about 3D body shape.
In controlled environments, multiple cameras and motion capture equipment provide accurate ground truth \cite{sigal_ijcv_10b,ionescu2013human36m,BenShabat21WACV,Trumble17BMVC,Nibali21IVC,Joo19PAMI,Huang:CVPR:2022,PROX:2019,Bhatnagar22CVPR,Zhang20CVPR,Yu20CVPR,Li21ICCV,vonMarcard2018,cai2022humman}.
In general, the cost and complexity of such captures limits the number of subjects, the variety of clothing, the types of motion, and the number of scenes. 

Several methods fit 3D body models to images to get pseudo ground truth \smpl parameters \cite{Kolotouros2019_spin,joo2020eft,moon2022neuralannot}. 
Networks trained on such data inherit any biases of the methods used to compute the ground truth; e.g.~a tendency to estimate bent knees, resulting from a biased pose prior.
Synthetic data does not suffer such biases.

Most image datasets are designed for 3D pose estimation and only a few have addressed body shape.
SSP-3D \cite{sengupta2020straps} contains 311 in-the-wild images of 62 people wearing tight sports clothing with pseudo ground truth body shape.
Human Bodies in the Wild (HBW) \cite{Shapy:2022} uses 3D body scans of 35 subjects who are also photographed in the wild with varied clothing.
HBW includes 2543 photos with ``perfect" ground truth shape.
Neither dataset is sufficiently large to train a general body shape regressor.

In summary, real data for training HPS involves a fundamental trade off.
One can either have diverse and natural images with low-quality ground truth or limited variability with high-quality ground truth.

\vspace{0.075in}
\noindent\textbf{Synthetic.}
%Big picture:
Synthetic data promises to address the limitations of real imagery and there have been many previous attempts.
While prior work has shown synthetic data to be useful (e.g.~for pre-training), no prior work has shown it to be sufficient without additional real training data.
We hypothesize that this is due to the fact that prior datasets have either been too small or not sufficiently realistic.
To date, no state-of-the-art method is trained from synthetic data alone.

Recently, Microsoft has shown that a synthetic dataset of faces is sufficiently accurate to train high-quality 2D feature detection \cite{wood2022eccv}.
While promising, human bodies are more complex.
AGORA  \cite{patel2020agora} provides realistic images of clothed bodies from {\em static} commercial scans with \smplx ground truth.
SPEC \cite{Kocabas_SPEC_2021} extends AGORA to more varied camera views.
These datasets have limited avatar variation (e.g.~few obese bodies) and lack motion.

\textbf{Synthetic from real.}
Since creating realistic people using graphics is challenging, several methods {\em capture} real people and then render them synthetically in new scenes \cite{moulding,ds:mucomupots,Mehta17TDV}.
For example, MPI-INF-3DHP \cite{Mehta17TDV} captures 3D people, augments their body shape, and swaps out clothing before compositing the people on images.
Like real data, these capture approaches are limited in size and variety.
Another direction takes real images of people plus information about body pose and, using machine learning methods, synthesizes new images that look natural \cite{Rogez2016,ZanfirAAAI2020}.
\iffalse
Another approach takes real images and separate mocap data and composites new synthetic images by stitching real image patches using information about body kinematics \cite{Rogez2016}. While the images resemble natural images, the ground truth is approximate and lacks shape information. 
Zanfir et al.~\cite{ZanfirAAAI2020} combine image synthesis methods with 3D representations of the body and scene to produce images that look realistic but have ground truth body information.
\fi 
This is a promising direction but, to date, no work has shown that this is sufficient train HPS regressors.

\textbf{Synthetic data without clothing.}
Synthesizing images of 3D humans on image backgrounds has a long history \cite{sminchisecuSynth2006}.
We focus on more recent datasets for training HPS regressors for parametric 3D human body models like SCAPE \cite{anguelov2005scape} (e.g.~Deep3DPose \cite{chen2016synthesizing}) and SMPL\cite{SMPL:2015} (e.g.~SURREAL \cite{varol17_surreal}).
Both apply crude textures to the naked body and then render the bodies against random image backgrounds.
In  \cite{chen2016synthesizing,Hoffmann:GCPR:2019}, the authors use domain adaptation methods to reduce the domain gap between synthetic and real images.
In  \cite{varol17_surreal} the authors use synthetic data largely for pre-training, requiring fine tuning on real images.

Since realistic clothes and textures are hard to generate, several methods render 
\smpl silhouettes or part segments and then learn to regress HPS from these \cite{PavlakosZZD18,denserac,Rong00CL19}.
While one can generate an infinite amount of such data, these methods rely on a separate process to compute silhouettes from images, which can be error prone.
For example, STRAPS \cite{sengupta2020straps} uses synthetic data to regress body shape from silhouettes.

\begin{figure*}[!t]
\centerline{\includegraphics[width=\textwidth]{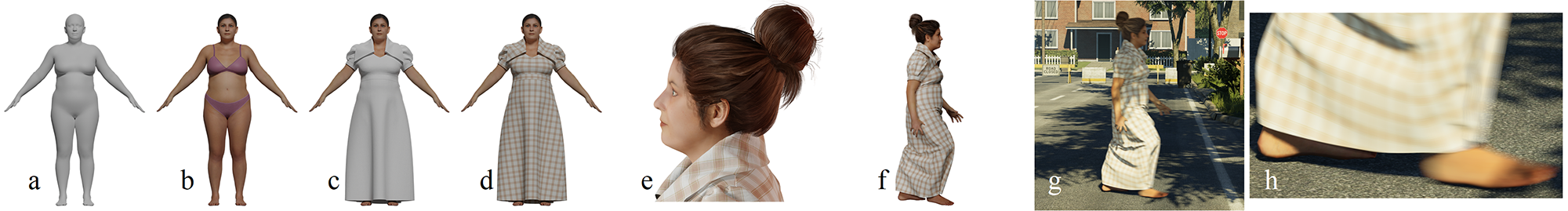}}
\vspace{-0.1in}
\caption{{\bf Dataset construction.} Illustration of each step in the process, shown for a single character.
Left to right:
(a) sampled body shape.
(b) skin texture.
(c) clothing simulation.
(d) cloth texture.
(e) hair.
(f) pose.
(g) scene and illumination.
(h) motion blur.
%(h) camera view.
%(i) illumination.
%(j) motion blur.
}
\vspace{-0.1in}
\label{fig:overview}
\end{figure*}

\textbf{Synthetic data with rigged clothing.}
Another approach renders commercial, rigged, body models for which the clothing deformations are not realistic. 
For example PSP-HDRI+ \cite{ebadi2022psphdri}, 3DPeople \cite{pumarola20193dpeople}, and JTA \cite{fabbri2018learning} use rigged characters but provide  only 3D skeletons so they cannot be used for body shape estimation.
The Human3.6M dataset \cite{ionescu2013human36m}  
includes mixed-reality data with rigged characters inserted into real videos.
There are only 5 sequences, 7.5K frames, and a limited number of rigged models, making it too small for training.
Multi-Garment Net (MGN) \cite{bhatnagar2019mgn} constructs a wardrobe from rigged 3D scans but renders them on images with no background.
Synthetic data has also been used to estimate ego-motion from head-mounted cameras
\cite{xu2019mo2cap2,hakada2022unrealego,tome2019xr}.
HSPACE \cite{bazavan2021hspace} uses 100 rigged people with 100 motions and 100 3D scenes.
To get more variety, they fit  GHUM \cite{xu2020ghum} to the scans and reshape them.
They train an HPS method \cite{zanfir2021thundr} on the data and note that ``models trained on synthetic data alone do not perform the best, not even when tested on synthetic data."
This statement is consistent with the findings of other methods 
and points to the need for increased diversity to achieve generalization.

\begin{figure}[t]
	\centerline{
    \includegraphics[trim=000mm 000mm 000mm 000mm, clip=true, width=1.00 \linewidth]{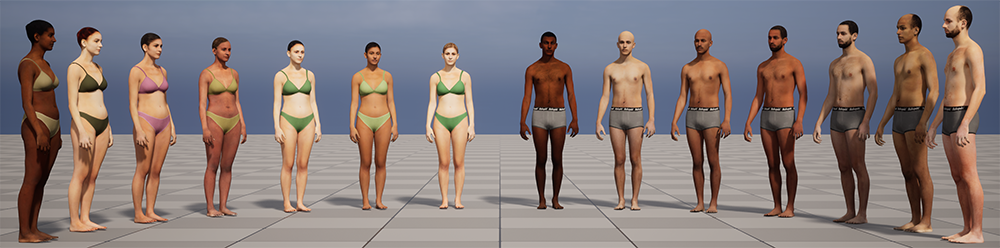}
    }
    \vspace{-0.1in}
	\caption{Skin tone diversity. Example body textures from 50 male and 50 female textures, covering a wide range of skin tones.}
	\label{fig:body_textures}
	\vspace{-0.15in}
\end{figure}

\textbf{Simulated clothing with images.}
Physics-based cloth simulation provides greater realism than rigged clothing and allows us to dress a wide range of bodies in varied clothing with full control.
The problem, however, is that physics simulation is challenging and this limits the size and complexity of previous datasets.
Liang and Lin \cite{liang2019shape} and Liu et al.~\cite{liu2019temporally}  
simulate 3D clothing draped on SMPL bodies. 
They render the people on image backgrounds with limited visual realism. 
BCNet \cite{DBLP:conf/eccv/JiangZHLLB20} uses both physics simulation and rigged avatars but the dataset is aimed at 3D clothing modeling more than HPS regression.
Other methods use a very limited number of garments or body shapes \cite{10.1111/cgf.13125,Wang:ACM:2018}.

\textbf{Simulated clothing without images.}
Several methods drape clothing on the 3D body to create datasets for learning 3D clothing deformations \cite{DRAPE2012,CLOTH3D,santesteban2022snug,patel20tailornet,tiwari20sizer}.
These datasets are limited in size and do not contain rendered images.

\textbf{Summary.}
The prior work is limited in one or more of these properties: body shapes, textures, poses, motions, backgrounds, clothing types, physical realism, cameras, etc.
As a result, these datasets are not sufficient for training HPS methods that work on real images.

\section{Dataset}
\label{sec:method}

Each step in the process of creating BEDLAM is explained below and illustrated in Fig.~\ref{fig:overview}.
Rendering is performed using Unreal Engine 5 (UE5) \cite{unreal}.
Additionally, the Sup.~Mat.~provides details about the process and all the 3D assets.
The Supplemental Video shows example sequences.

\subsection{Dataset Creation}

\paragraph{Body shapes.} 
We want a diversity of body shapes, from slim to obese.
We get 111 adult bodies in \smplx format from AGORA dataset.
These bodies mostly correspond to models with low BMI.
To  increase diversity, we sample an additional 80 male and 80 female bodies with $\mathrm{BMI}>30$ from the CAESAR dataset~\cite{CAESAR}. 
Thus we sample body shapes from a diverse pool of \numBodies body shapes in total.
The ground truth body shapes are represented with 11 shape components in the \smplx  gender-neutral shape space. See Sup.~Mat.~for more details about the body shapes.

\myparagraph{Skin tone diversity.}
HPS estimation will be used in a wide range of applications, thus
it is important that HPS solutions be inclusive.
Existing HPS datasets have not been designed to ensure diversity and this is a key advantage of synthetic data.
Specifically, we use 50 female and 50 male commercial skin albedo textures from Meshcapade~\cite{meshcapade}  with minimal clothing and a resolution of 4096x4096. These artist-created textures represent a total of seven ethnic groups  (African, Asian, Hispanic, Indian, Mideast, South East Asian and White) with multiple variations within each.
A few examples are shown in Fig.~\ref{fig:body_textures}.

\begin{figure}[t]
 \centerline{
 \includegraphics[trim=000mm 000mm 000mm 000mm, clip=true, width=1.00 \linewidth]{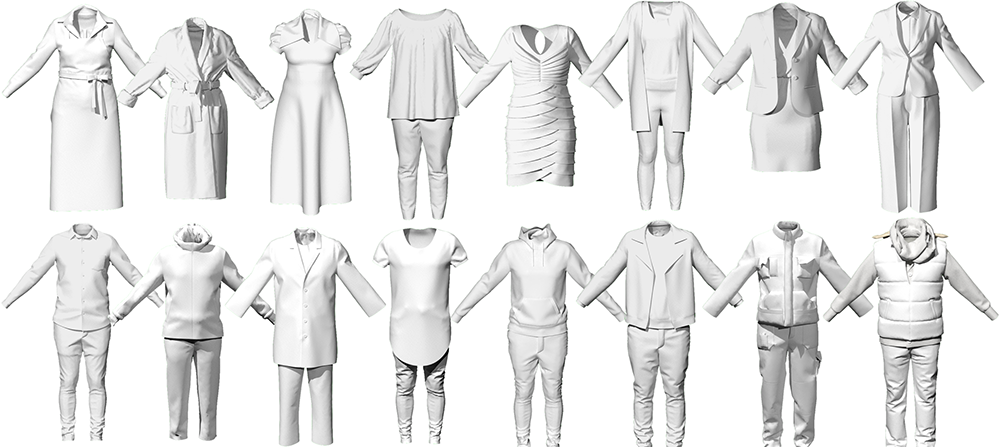}
 }
 \centerline{
 \includegraphics[trim=000mm 000mm 000mm 000mm, clip=true, width=1.00 \linewidth]{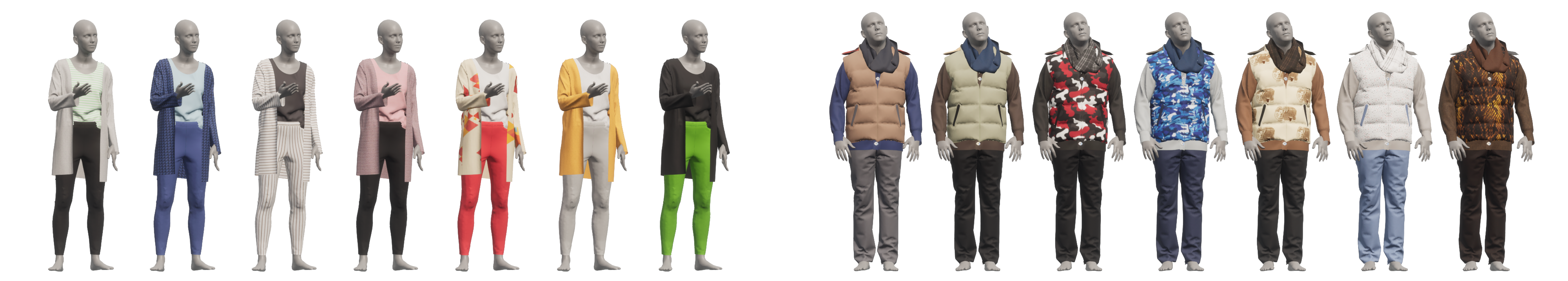}
 }
 \vspace{-0.1in}
 \caption{Diversity of clothing and texture. Top: samples from BEDLAM's \numOutfits outfits with real-world complexity. Bottom: each outfit has several clothing textures. Total: 1691.}
	\label{fig:outfits}
	\vspace{-0.2in}
\end{figure}

\myparagraph{3D Clothing and textures.}
A key limitation of previous synthetic datasets is the lack of diverse and complex 3D clothing with realistic physics simulation of the clothing in motion.
To address this, we hired a 3D clothing designer to create 111 unique real-world outfits, including but not limited to T-shirts, shirts, jeans, tank tops, sweaters, coats, duvet jackets, suits, gowns, bathrobes, vests, shorts, pants, and skirts. Unlike existing synthetic clothing datasets, our clothing designs have complex and realistic structure and details such as pleats, pockets, and buttons.
Example outfits are shown in Fig.~\ref{fig:outfits}.
We use commercial simulation software from CLO3D \cite{clo3d} to obtain realistic clothing deformations with various body motions for the bodies from the AGORA dataset (see Supplemental Video).
This 3D dataset is a unique resource that we will make available to support a wide range of research on learning models of 3D clothing.

Diversity of clothing appearance is also important.
For each outfit we design 5 to 27 clothing textures with different colors and patterns using WowPatterns \cite{wowpatterns}. In total we have \numClothTextures unique clothing textures (see Fig.~\ref{fig:outfits}).

\begin{figure}[t]
\centerline{\includegraphics[width=0.95\columnwidth]{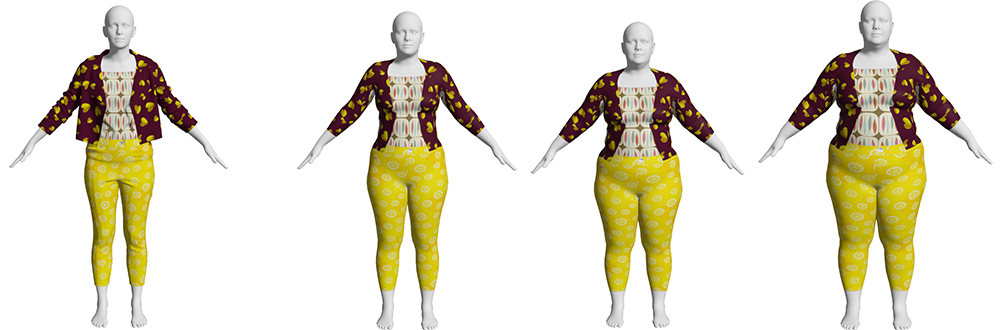}}
\vspace{-0.1in}
 \caption{Clothing as texture maps for high-BMI bodies.
 Left: example simulated clothing. Right: clothing texture mapped on bodies with BMIs of 30, 40, and 50. }
 \vspace{-0.15in}
	\label{fig:clothingSkins}
\end{figure}

For high-BMI bodies, physics simulation of clothing fails frequently due to the difficulty of garment auto-resizing and interpenetration between body parts. 
For such situations, we use clothing texture maps that look like clothing ``painted" on the body.
Specifically, we auto-transfer the textures of 1738 simulated garments onto the body UV-map using Blender.
We then render high-BMI body shapes using these textures (see Fig.~\ref{fig:clothingSkins}).

\begin{figure}[t]
\centerline{
 \includegraphics[trim=000mm 000mm 000mm 000mm, clip=true, width=1.00 \linewidth]{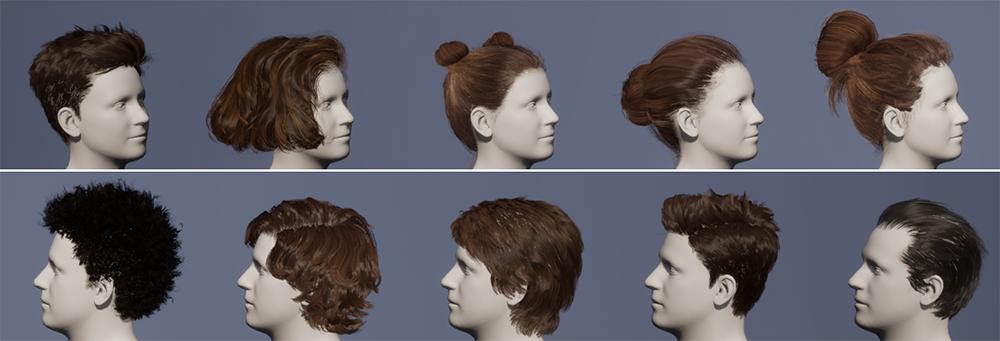}
 }
 \vspace{-0.1in}
\caption{10 examples of BEDLAM's \numWigs hairstyles. %Top row: female.  Bottom row male.
}
\vspace{-0.2in}
\label{fig:hair}
\end{figure}

\myparagraph{Hair.}
We use the Character Creator (CC) software from Reallusion \cite{reallusion}
and purchased hairstyles to generate \numWigs hairstyles (Fig.~\ref{fig:hair}).
We auto-align our \smplx female and male template mesh to the CC template mesh and then transfer the \smplx deformations to it. 
We then apply the hairstyles in the CC software to match our custom headshapes. 
We export the data to Blender to automatically process the hair mesh vertices so that their world vertex positions are relative to the head node positioned at the origin.
Note that vendor-provided plugins take care of the extensive shader setup needed for proper rendering of these hair-card-based meshes. 
Finally the ``virtual toupees" are imported into Unreal Engine where they are attached to the head nodes of the target \smplx animation sequences. 
The world-pose of each toupee is then automatically driven by the Unreal Engine animation system.

\myparagraph{Human motions.}
We sample human motions from the AMASS dataset~\cite{AMASS:ICCV:2019}. 
Due to the long-tail distribution of motions in the dataset, a naive random sampling leads to a strong bias towards a small number of frequent motions, resulting in low motion diversity. 
To avoid this, we make use of the motion labels provided by BABEL~\cite{BABEL:CVPR:2021}. 
Specifically, we sample different numbers of motion sequences for each motion category according to their motion diversity (see Sup.~Mat.~for details). 
This leads to \numMotions unique motions. 
Each motion sequence lasts from 4 to 8 seconds. 
Naively transferring these motions to new body shapes in the format of joint angle sequences may lead to self-interpenetration, especially for high-BMI bodies. To avoid this, we follow the approach in TUCH~\cite{mueller2021tuch} to resolve collisions among body parts for all the high-BMI bodies.
While the released dataset is rendered at 30fps,  we only use every $5^{th}$ frame for training and evaluation to reduce pose redundancy.
The full sequences will be useful for research on 3D human tracking, e.g.~\cite{rajasegaran2022tracking,yuan2022glamr,sun2023trace,SLAHMR}.

Unfortunately, most motion sequences in AMASS contain no hand motion. 
To increase realism, diversity, and enable research on hand pose estimation, we add hand motions sampled from the GRAB~\cite{GRAB:2020} dataset.
While these hand motions do not semantically ``match" the body motion, the rendered sequences still look realistic, and  are sufficient for training full-body and hand regressors.

\myparagraph{Scenes and lighting.}
We represent the environment either through \numHDRI panoramic HDRI images \cite{polyhaven} or through \numScenes 3D scenes. 
We manually select HDRI panoramas that enable 
the plausible placement of animated bodies on a flat ground plane up to a distance of 10m.
We randomize the viewpoint into the scenes and use the HDRI images for image-based lighting.
For the 3D scenes we focus on indoor environments since the HDRI images already cover  outdoor environments well. 
To light the 3D scenes, we either use Lightmass precalculated global illumination or the new Lumen real-time global illumination system introduced in UE5 \cite{unreal}.

\myparagraph{Multiple people in the scene.}
For each sequence we randomly select between 1 and 10 subjects. 
For each subject a random animation sequence is selected. 
We leverage binary ground occupancy maps and randomly place the moving people into the scene such that they do not collide with each other or scene objects.
See Sup.~Mat.~for details.

\myparagraph{Cameras.}
For BEDLAM, we focus on cameras that one naturally encounters in common computer vision datasets.
For most sequences we use a static camera with randomized camera extrinsics.
The extrinsics correspond to typical ground-level hand-held cameras in portrait and landscape mode. 
Some sequences use additional extrinsics augmentation by simulating a cinematic orbit camera shot. Camera intrinsics are either fixed at HFOV of 52 and 65 or  zoom in from 65 to 25 HFOV.

\myparagraph{Rendering.}
We render the image sequences using the UE5 game engine rasterizer with the cinematic camera model simulating a 16:9 DSLR camera with a 36x20.25mm sensor size. 
The built-in movie render subsystem (Movie Render Queue) is used for deterministic and high-quality image sequence generation. 
We simulate motion blur caused by the default camera shutter speed by generating 7 temporal image samples for each final output image. 
A single Windows 11 PC using one NVIDIA RTX3090 GPU was used to render all color images and store them as 1280x720 lossless compressed PNG files with motion blur at an average rate of more than 5 images/s.

\myparagraph{Depth maps and segmentation.}
While our focus is on HPS regression, BEDLAM can support other uses.
Since the data is synthetic, we also render out depth maps and segmentation masks with semantic labels (hair, clothing, skin).
These are all available as part of the dataset release.
 See Sup.~Mat.~for details.

\subsection{Dataset Statistics}

In summary, BEDLAM is generated from a combination of 
\numBodies bodies,
\numWigs hairstyles,
\numOutfits types of clothing,
with \numClothTextures clothing textures,
\numMotions human motions,
in \numHDRI HDRI scenes and \numScenes 3D scenes,
with on average \avgPeople person per scene,
and a variety of camera poses.
See Sup.~Mat.~for detailed statistics.
This results in \numClips motion clips, from which we use \numFrames RGB frames in total.
We compute the size of the dataset in terms of the number of unique bounding boxes containing individual people.
BEDLAM contains \numBoxes such bounding boxes, which we divide into sets of about \numTrain, \numValid, and \numTest examples for training, validation, and test, respectively.
See Sup.~Mat.~for a detailed comparison of BEDLAM's size and diversity relative to existing real and synthetic datasets.

\section{Experiments}
\subsection{Implementation Details}\label{implementation-details}

We train both HMR and CLIFF on the synthetic data (BEDLAM+AGORA) using an HRNet-W48\cite{sun2019deep} backbone and refer to these as BEDLAM-HMR and BEDLAM-CLIFF respectively. 
We conduct different experiments with the weights of the backbone initialized from scratch, using ImageNet \cite{imagenet_cvpr09}, or using a pose estimation network trained on COCO \cite{xiao2018simple}. 
We represent all ground truth bodies in a gender neutral shape space to supervise training; we do not use gender labels.
We remove the adversary from HMR and set the ground truth hand poses to neutral when training BEDLAM-HMR and BEDLAM-CLIFF. 
We apply a variety of data augmentations during training.
We experiment with a variety of losses; 
the final loss is a combination of MSE loss on model parameters, projected keypoints, 3D joints, and an L1 loss on 3D vertices.

We re-implement CLIFF (called CLIFF\textsuperscript{\textdagger}) and train it on only real image data using the same settings as BEDLAM-CLIFF.
Following \cite{li2022cliff}, we train CLIFF\textsuperscript{\textdagger} using Human3.6M\cite{ionescu2013human36m}, MPI-INF-3DHP\cite{Mehta17TDV}, and 2D datasets COCO\cite{Lin14ECCV} and MPII\cite{Andriluka14CVPR} with pseudo-GT provided by the CLIFF annotator.
Table \ref{tab:3dpw-rich-compare} shows that, when trained on real images, and fine-tuned on 3DPW training data, CLIFF\textsuperscript{\textdagger} matches the accuracy reported in \cite{li2022cliff} on 3DPW and is even more accurate on RICH.
Thus our implementation can be used as a reference.

We also train a full body network, BEDLAM-CLIFF-X, to regress body and hand poses.
To train the hand network, we create a dataset of hand crops from  BEDLAM training images using the ground truth hand keypoints. 
Since hands are occluded by the body in many images, MediaPipe\cite{mediapipe} is used to detect the hand in the crop. 
Only the crops where the hand is detected with a confidence greater than 0.8 are used in the training. 
For details see Sup.~Mat.

\begin{figure*}[t]
 \centerline{
 \includegraphics[trim=000mm 000mm 000mm 000mm, clip=true, width=1.0 \linewidth]{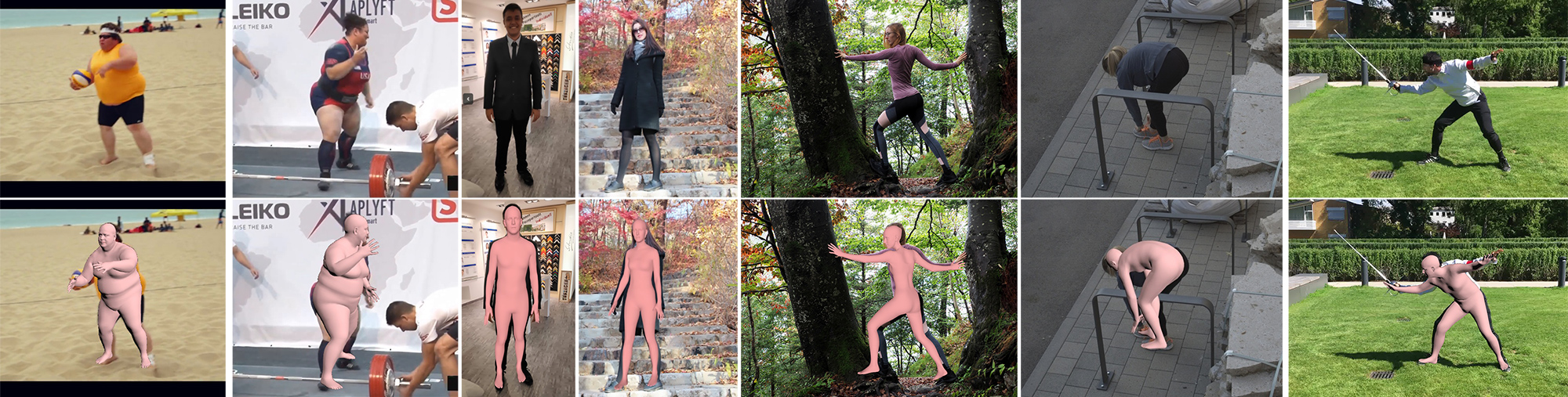}
 }
  \vspace{-0.1in}
 \caption{%{\bf Qualitative results.} 
Example BEDLAM-CLIFF results from all test datasets. Left to right: SSP-3D $\times$ 2, HBW $\times$ 3, RICH, 3DPW.}
	\label{fig:qualitative}
 	\vspace{-0.1in}
\end{figure*}

\subsection{Datasets and Evaluation Metrics}
{\bf Datasets.} For training we use around \numTrain crops from BEDLAM  and 85K crops from AGORA \cite{patel2020agora}.
We also finetune  BEDLAM-CLIFF and BEDLAM-HMR on 3DPW training data; these are called  BEDLAM-CLIFF* and BEDLAM-HMR*.
To do so, we convert the 3DPW\cite{vonMarcard2018} GT labels in SMPL-X format.
%AGORA is a synthetic dataset created with static scans rendered into 3D scenes and HDRI environment similar to BEDLAM. 
We use 3DPW for evaluation but, since it has limited camera variation, we also use RICH \cite{Huang:CVPR:2022} which has more varied camera angles. 
Both 3DPW and RICH have limited body shape variation, hence to evaluate body shape we use SSP-3D \cite{sengupta2020straps} and HBW \cite{Shapy:2022}.  
In Sup.~Mat.~we also evaluate on Human3.6M \cite{ionescu2013human36m}
and observe that, without fine-tuning on the dataset, training on BEDLAM produces more accurate results than training using real images; that is, BEDLAM generalizes better to the lab data.
To evaluate the output from BEDLAM-CLIFF-X, we use the AGORA and BEDLAM test sets.

\begin{table}[t]
% 	\scriptsize
%\small
	\resizebox{\columnwidth}{!}{
		\begin{tabular}{l|ccc|ccc}
			\toprule
			\multicolumn{1}{c|}{Method} & 
			\multicolumn{3}{c|}{3DPW (14) }&         
			\multicolumn{3}{c}{RICH (24)}\\
			\midrule
			& \multicolumn{1}{l}{\footnotesize PA-MPJPE} & \multicolumn{1}{l}{\footnotesize MPJPE} & 
			\multicolumn{1}{l|}{\footnotesize PVE} & 
			\multicolumn{1}{l}{\footnotesize PA-MPJPE} &
			\multicolumn{1}{l}{\footnotesize MPJPE} &
			\multicolumn{1}{l}{\footnotesize PVE} \\

			\midrule
			PARE*\cite{pare} &46.5	&74.5	&88.6 & 60.7 &109.2 &123.5\\
			METRO*\cite{metro} &47.9	& 77.1	& 88.2 & 64.8 & 114.3 & 128.9\\
			CLIFF*\cite{li2022cliff}	& 43.0	& 69.0 &81.2 & 56.6 & 102.6 & 115.0 \\
   			CLIFF\textsuperscript{\textdagger}*	&  43.6 & 68.8 & 82.1 & 55.7 & 91.6 & 104.4 \\
                BEDLAM-HMR*  &43.3 & 71.8 &83.6 & 50.9 & 88.2 & 101.8\\
                BEDLAM-CLIFF*   & \textbf{43.0} & \textbf{66.9} & \textbf{78.5} & \textbf{50.2} & \textbf{84.4} & \textbf{95.6}\\
	      %  HybrIK*\cite{li2021hybrik} &&&& 50.7 & 93.5 & 104.7 \\
			\midrule
			HMR\cite{Kanazawa2018_hmr} & 76.7 & 130 & N/A & 90.0 & 158.3 & 186.0 \\
			SPIN\cite{Kolotouros2019_spin} & 59.2	& 96.9 &116.4 &69.7& 122.9 & 144.2\\
			SPEC\cite{Kocabas_SPEC_2021} &	53.2 &	96.5 &	118.5 & 72.5 &127.5 & 146.5 \\
			PARE\cite{pare} & 50.9 & 82.0 & 97.9 & 64.9 & 104.0 & 119.7 \\
            HybrIK\cite{li2021hybrik} & 48.8 &	80 & 94.5 & 56.4 & 96.8 & 110.4 \\
            
			Pang et. al.\cite{pang2022benchmark}	&47.3	&81.9 &	96.5 & 63.7 & 117.6 & 136.5 \\
                CLIFF\textsuperscript{\textdagger} &  \textbf{46.4} &73.9 & 87.6 & 55.7 & 90.0 & 102.0 \\
			BEDLAM-HMR 	& 47.6 & 79.0 & 93.1 & 53.2 & 91.4 & 106.0 \\
			BEDLAM-CLIFF  & 46.6 & \textbf{72.0} & \textbf{85.0} & \textbf{51.2} & \textbf{84.5} & \textbf{96.6}\\

			\bottomrule						\end{tabular}
		}
		\vspace{-0.08in}
    \caption{Reconstruction error on 3DPW and RICH. *Trained with 3DPW training set.  \textsuperscript{\textdagger}Trained on real images with same setting as BEDLAM-CLIFF.  Parenthesis: (\#\/joints).}
	\label{tab:3dpw-rich-compare}
	\vspace{-0.1in}
\end{table}
{\bf Evaluation metrics.}
%We use standard PVE, MPJPE, PA-MPJPE metrics for evaluation of body pose and PVE-T-SC and $\text{P2P}_{\text{20k}}$ for evaluation on body shape. 
We use standard metrics to evaluate body pose and shape accuracy.
PVE and MPJPE represent the average error in vertices and joints positions, respectively, after aligning the pelvis. 
PA-MPJPE further aligns the rotation and scale before computing distance. 
PVE-T-SC is per-vertex error in a neutral pose (T-pose) after scale-correction \cite{sengupta2020straps}. $\text{P2P}_{\text{20k}}$ is per-vertex error in a neutral pose, computed by evenly sampling 20K points on SMPL-X’s surface \cite{Shapy:2022}. 
All errors are in mm.

%Explain conversion from SMPLX to SMPL for evaluation and RICH evaluation
For evaluation on 3DPW and SSP-3D, we convert our predicted \smplx meshes to SMPL format by using a vertex mapping $D\in\mathbb{R}^{10475 \times 6890}$~\cite{Pavlakos2019_smplifyx}.
The RICH dataset has ground truth in SMPL-X format but hand poses are less reliable than body pose due to noise in multi-view fitting. 
Hence, we use it only for evaluating body pose and shape. 
We convert the ground truth SMPL-X vertices to SMPL format using $D$ after setting the hand and face pose to neutral. 
To compute joint errors, we use 24 joints computed from these vertices using the SMPL joint regressor. 
For evaluation on AGORA-test and BEDLAM-test, we use a similar evaluation protocol as described in \cite{patel2020agora}. 

\subsection{Comparison with the State-of-the-Art}
Table~\ref{tab:3dpw-rich-compare} summarizes the key results.
(1) Pre-training on BEDLAM and fine-tuning with a mix of 3DPW and BEDLAM training data gives the most accurate results on 3DPW and RICH (i.e.~BEDLAM-CLIFF* is more accurate than CLIFF\textsuperscript{\textdagger}* or \cite{li2022cliff}).
(2) Using the same training, makes HMR (i.e.~BEDLAM-HMR*) nearly as accurate on 3DPW and more accurate than CLIFF\textsuperscript{\textdagger}* on RICH. 
This suggests that even simple methods can do well if trained on good data.
(3) BEDLAM-CLIFF, with no 3DPW fine-tuning, does nearly as well as the fine-tuned version and generalizes better to RICH than CLIFF with, or without, 3DPW fine-tuning.
(4) Both CLIFF and HMR trained only on synthetic data outperform the recent methods in the field. 
This suggests that more effort should be put into obtaining high-quality data. See Sup.~Mat.~for \smplx results.

Table~\ref{tab:ssp-hbw-compare} shows that BEDLAM-CLIFF has learned to estimate body body shape under clothing. 
While SHAPY \cite{shape_under_cloth:CVPR17} performs best on HBW and Sengputa et al.~\cite{sengupta2021hierarchicalICCV} performs best on SSP-3D, both of them perform poorly on the other dataset. 
Despite not seeing either of the training datasets, BEDLAM-CLIFF ranks 2nd on SSP-3D and HBW. 
BEDLAM-CLIFF has the best rank averaged across the datasets, showing its generalization ability.

Qualitative results on all these benchmarks are shown in Fig.~\ref{fig:qualitative}.
Note that, although we do not assign gender labels to any of the training data, we find that, on test data, methods trained on BEDLAM predict appropriately gendered body shapes.
That is, they have automatically learned the association between image features and gendered body shape.

\subsection{Ablation Studies}
Table~\ref{tab:ablation1} shows the effect of varying datasets, backbone weights and percentage of data; see Sup.~Mat.~for the full table with results for HMR.
We train with synthetic data only and measure the performance on 3DPW. 
Note that the backbones are pre-trained on image data, which is standard practice.
Training them from scratch on BEDLAM gives worse results. 
It is sufficient to train using simple 2D task for which there is plentiful data.
Similar to \cite{pang2022benchmark}, we find that training the backbone on a 2D pose estimation task (COCO) is important.
We also  vary the percentage of BEDLAM crops used in training. 
Interestingly, we find that  uniformly sampling just 5\%\/ of the crops from BEDLAM produces reasonable performance on 3DPW.
Performance monotonically improves as we add more training data.
Note that 5\% of BEDLAM, i.e.~38K crops, produces better results than 85K crops from AGORA, suggesting that BEDLAM is more diverse.
Still, these synthetic datasets are complementary, with our best results coming from a combination of the two.
We also found that realistic clothing simulation
 leads to significantly better results than training with textured bodies. 
 This effect is more pronounced when using a backbone pre-trained on ImageNet rather than COCO. 
 See Sup.~Mat.~for details.

\begin{table}[t]
	\centering
%\small
	\resizebox{\columnwidth}{!}{
		\begin{tabular}{l|c|c|c|c|c|c}
			\toprule
			\multicolumn{1}{c|}{Method} & 
			\multicolumn{1}{c|}{Model} & 
			\multicolumn{2}{c|}{SSP-3D} & 
    	\multicolumn{2}{c|}{HBW} &
    	\multicolumn{1}{c}{Average} \\ 
     & &
			\multicolumn{1}{c}{PVE-T-SC}&
			\multicolumn{1}{c|}{Rank} &
			\multicolumn{1}{c}{$\text{P2P}_{\text{20k}}$}&
			\multicolumn{1}{c|}{Rank} &
			\multicolumn{1}{c}{Rank}\\
			\midrule
            HMR \cite{Kanazawa2018_hmr} &  SMPL & 22.9 & 8 & - & - & -\\
            SPIN \cite{Kolotouros2019_spin} & SMPL & 22.2 & 7 & 29 & 4 & 5.5\\
            SHAPY \cite{Shapy:2022} & SMPL-X &  19.2 & 6 & 21 & 1 & 3.5 \\
            STRAPS \cite{sengupta2020straps} & SMPL & 15.9 & 4 & 47 & 6 & 5\\
            Sengupta et al.~\cite{sengupta2021probabilisticCVPR} & SMPL & 15.2 & 3 & -& - & -\\
            Sengupta et al.~\cite{sengupta2021hierarchicalICCV }& SMPL & 13.6 & 1 & 32 & 5 & 3\\
            CLIFF\textsuperscript{\textdagger} & SMPL & 18.4 & 5 & 27 & 3 & 4\\
            BEDLAM-CLIFF & SMPL-X & 14.2 & 2 & 22 & 2 & 2 \\
			\bottomrule						
			\end{tabular}
  }
		\vspace{-0.08in}
    \caption{Per-vertex 3D body shape error on the SSP-3D and HBW test set in T-pose (T). SC refers to scale correction.
    }
	\label{tab:ssp-hbw-compare}
	\vspace{-0.1in}
\end{table}

\begin{table}[t]
	\scriptsize
	\centering
%	\begin{center} 
	\resizebox{\columnwidth}{!}{
 \begin{tabular}{l|c|c|c|c|c|c}
	\toprule
	\multicolumn{1}{c|}{Method} & \multicolumn{1}{c|}{Dataset} & \multicolumn{1}{c|}{Backbone}  & \multicolumn{1}{c|}{Crops \%} & \multicolumn{1}{c|}{PA-MPJPE} & \multicolumn{1}{c|}{MPJPE} & \multicolumn{1}{c}{PVE} \\
	\midrule
	CLIFF & B+A & scratch & 100 & 61.8 & 97.8 & 115.9\\
	CLIFF &  B+A & ImageNet & 100 & 51.8 & 82.1 & 96.9\\
	CLIFF &  B+A & COCO & 100 & 47.4 & 73.0 & 86.6 \\

	\midrule
	CLIFF &  B & COCO & 5 &  54.0 & 80.8 & 96.8 \\
	CLIFF &  B & COCO & 10 &  53.8 & 79.9 & 95.7\\
	CLIFF &  B & COCO & 25 & 52.2 & 77.7 & 93.6\\
	CLIFF &  B & COCO & 50 & 51.0 & 76.3 & 91.1 \\

	\midrule
	CLIFF &  A & COCO & 100 & 54.0 & 88.0 & 101.8\\	
	CLIFF &  B & COCO & 100 & 50.5 & 76.1 & 90.6\\
\bottomrule
\end{tabular}
}
%	\end{center} 
\vspace{-0.08in}
	\caption{Ablation experiments on 3DPW. B denotes BEDLAM and A denotes AGORA. Crop \%'s only apply to BEDLAM.}
	\label{tab:ablation1}
\vspace{-0.1in}
\end{table}

\section{Limitations and Future Work}

Our work demonstrates that synthetic human data can stand in for real image data.
By providing tools to enable researchers to create their own data, we hope the community will create new and better synthetic datasets.
To support that effort, below we provide a rather lengthy discussion of limitations and steps for improvement; more in Sup.~Mat.

\textbf{Open source assets.}
There are many high-quality commercial assets that we did not use in this project because their licences restrict their use in neural network training.
This is a significant impediment to research progress.  More open-source assets are needed.

\textbf{Motion and scenes.}
The human motions we use are randomly sampled from AMASS. 
In real life, clothing and motions are correlated, as are scenes and motions.
Additionally, people interact with each other and with objects in the world.
Methods are needed to automatically synthesize such interactions realistically \cite{yi2022mime}. 
Also, the current dataset has relatively few sitting, lying, and complex sports poses, which are problematic for cloth simulation.

\textbf{Hair.}
BEDLAM lacks hair physics, long hairstyles, and hair color diversity.
Our solution, based on hair cards, is not fully realistic and suffers from artifacts under certain lighting conditions.
A strand-based hair groom solution would allow long flowing hair with hair-body interaction and proper rendering with diverse lighting.

\textbf{Body shape diversity.}
Our distribution of body shapes is not uniform (see Sup.~Mat.).
Future work should use a more even distribution and add 
children and people with diverse body types (scoliosis, amputees, etc.).
Note that draping high-BMI models in clothing is challenging because the mesh self-intersects, causing failures of the cloth simulation.
Retargeting AMASS motions to high-BMI subjects is also problematic.
We describe solutions in Sup.~Mat.

\textbf{More realistic body textures.}
Our skin textures are diverse but lack  details and realistic reflectance properties.
Finding high-quality textures with appropriate licences, however, is difficult.

\textbf{Shoes.}
BEDLAM bodies are barefoot.
Adding basic shoes is fairly straightforward but the general problem is actually complex because shoes, such as high heels, change body posture and gait.
Dealing with high heels requires retargeting, inverse kinematics, or new motion capture.

\textbf{Hands and Faces.}
There is very little mocap data with the full body and hands and even less with hands interacting with objects.
Here we ignored facial motion; there are currently no datasets that evaluate full body and facial motion. 

\section{Discussion and Conclusions}

Based on our experiments we can now try to answer the question ``Is synthetic data all you need?"
Our results suggest that BEDLAM is sufficiently realistic that methods trained on it generalize to real scenes that vary significantly (SSP-3D, HBW, 3DPW, and RICH).  
If BEDLAM does not well represent a particular real-image domain (e.g.~surveillance-camera footage), then one can re-purpose the data by changing camera views, imaging model, motions, etc.
Synthetic data will only get more realistic, closing the domain gap further.
Then, does architecture matter?
The fact that BEDLAM-HMR outperforms many recent, more sophisticated, methods argues that it may be less 
important than commonly thought.

There is one caveat to the above, however.
We find that HPS accuracy depends on backbone pre-training.
Pre-training the backbone for 2D pose estimation on COCO exposes it to all the variability of real images and seems to help it generalize.
We expect that pre-training will eventually be unnecessary as synthetic data improves in realism.

We believe that there is much more research that BEDLAM can support.
None of the methods tested here estimate humans in {\em world coordinates} \cite{sun2023trace,SLAHMR}. The best methods also do not exploit temporal information or action semantics.
BEDLAM can support new methods that push these directions.
BEDLAM can also be used to model 3D clothing and learn 3D avatars using implicit shape methods.

\smallskip
\textbf{Acknowledgments.} We thank STUDIO LUPAS GbR for creating the 3D clothing, Meshcapade GmbH for the skin textures, Lea M\"{u}ller for help removing self-intersections in high-BMI bodies and
Timo Bolkart for aligning SMPL-X to the CC template mesh. We thank T. Alexiadis, L. Sánchez, C. Mendoza, M. Ekinci and Y. Fincan for help with clothing texture generation.
%\textcolor{\hcolor}{We thank ??? for creating the clothing textures. }

\noindent 
\textbf{Disclosure:} \url{https://files.is.tue.mpg.de/black/CoI_CVPR_2023.txt}

%%%%%%%%% REFERENCES
{\balance\small
\bibliographystyle{ieee_fullname}
\bibliography{BEDLAM_main}
}
% Uncomment this to remove supmat as appendix
\clearpage
\appendix
{\noindent\Large\textbf{Supplementary Material}}

This document supplements the main text with 
(1) More details about the creation of the dataset.
(2) More statistics about the dataset's contents.
(3) More example images from the dataset.
(4) Experimental results referred to in the main text.
(5) Visual presentation of the qualitative results.

In addition to this document, please see the \textbf{Supplemental Video}, where the motions in the dataset are presented.
The video, data, and related materials can be found at \url{https://bedlam.is.tue.mpg.de/}

\paragraph{\modelname: Definition}
\begin{quote}
noun\\
\em A scene of uproar and confusion: there was bedlam in the courtroom.    
\end{quote}
The name of the dataset refers to the fact that the synthetic humans in the dataset are animated independently of each other and the scene. 
The resulting motions have a chaotic feel; please see the video for examples.

\section{Dataset creation}
\begin{figure}[h]
 \centerline{
 \includegraphics[trim=000mm 000mm 000mm 000mm, clip=true, width=1.00 \linewidth]{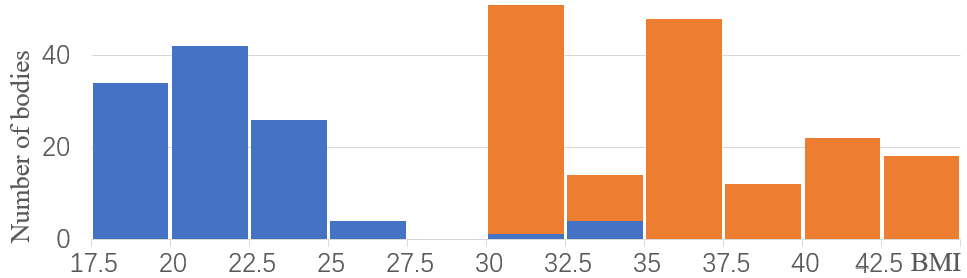}
 }
 \centerline{
 \includegraphics[trim=000mm 000mm 000mm 000mm, clip=true, width=1.00 \linewidth]{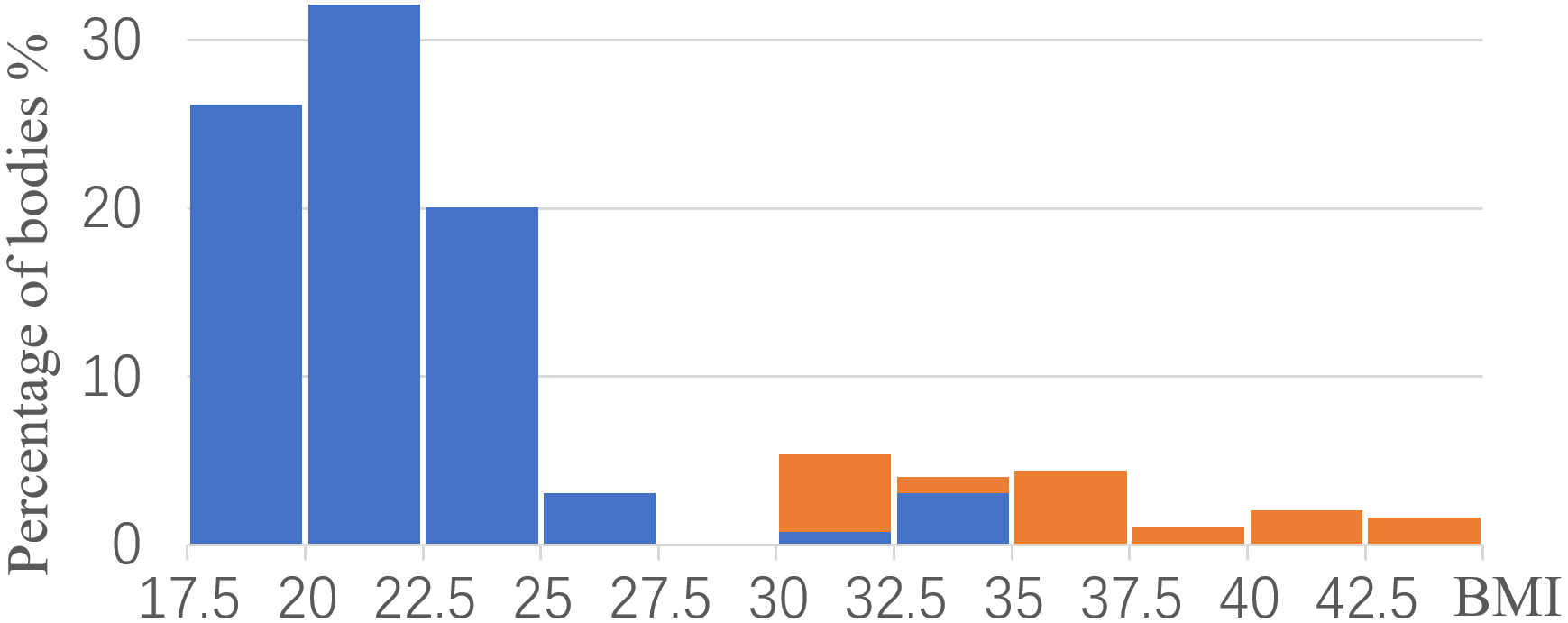}
 }
%  \caption{Body diversity; BMI distribution in BEDLAM. Blue bars represent bodies from AGORA dataset, while orange bars represents high-BMI bodies from CAESAR dataset. BEDLAM uses both to cover a wide range of BMIs; \numBodies in total.}
\caption{Body diversity in BEDLAM. Top: BMI distribution of the \numBodies different body shapes uses in BEDLAM. 
Bottom: BMI distribution in all rendered videos; 55009 in total. Blue bars represent bodies from the AGORA dataset, while orange bars represents high-BMI bodies from CAESAR dataset. BEDLAM uses both to cover a wide range of BMIs.}
	\label{fig:body_BMIs}
\end{figure}

\paragraph{Body shape diversity.} 

The AGORA \cite{patel2020agora} dataset has 111 adult bodies in \smplx format \cite{Pavlakos2019_smplifyx}.
These bodies mostly correspond to models with low BMI.
Why do we use the bodies from AGORA?
To create synthetic clothing we focused on creating synthetic versions of the clothed scans in AGORA.
That is, we create ``digital twins" of the AGORA scans.
Our hope is that having 3D scans paired with simulated digital clothing will be useful for research on 3D clothing.
Thus our 3D clothing is designed around AGORA bodies.
Note that we do not make use of this property in BEDLAM but did this to enable future use cases.
To  increase diversity beyond AGORA, we sample an additional 80 male and 80 female bodies with $\mathrm{BMI}>30$ from the CAESAR dataset~\cite{CAESAR}.

Note that the AGORA and CAESAR bodies are represented in gendered shape spaces using 10 shape components.
When we render the images, we use these gendered bodies.
For \modelname\/ we use a gender-neutral shape space, enabling networks to automatically learn the appropriate body shape within this space, effectively learning to recognize gender.
To make the ground truth shapes for BEDLAM in this gender-neutral space, we fit the gender-neutral model with 11 \smplx shape components to the gendered bodies.  
This is trivial since the meshes are in full correspondence.
We use 11 shape components because, in the gender neutral space, the first component roughly captures the differences between male and female body shapes.
Thus, adding one extra component means that the \smplx ground truth (GT) approximates the original gendered body shapes.
% Since we use a 10-D shape space for BEDLAM, 
There is some loss of fidelity but it is minimal;
the V2V error between the rendered bodies and the GT bodies in neutral pose is 2.4mm.

%\paragraph{Body shape diversity.} 

Ideally, we want a diversity of body shapes, from slim to obese.
Figure \ref{fig:body_BMIs} shows the distribution of body BMIs in the training set.
Specifically, we show the distribution of AGORA and CAESAR bodies, from which we sample.
We also show the final distribution of BMIs in the training images.

Notice that the AGORA bodies are almost all slim.  We add the CAESAR bodies to increase diversity and enable the network to predict high-BMI shapes.
There is a dip in the distribution between 25-30 BMI. 
This happens to be precisely where the peak of the real population lies.
Despite this lack of average BMIs, BEDLAM does a good job of predicting body shape, suggesting that it has learned to generalize.

Note that is it not clear what the right distribution for training is -- one could mimic the distribution of a specific population or uniformly sample across BMIs.
We plan to evaluate this and increase the diversity of the dataset; please check the project page for updates.
Future work should also expand the types of bodies used to include 
children and people with diverse body types (athletes, little people, scoliosis, amputees, etc.).
Note that draping high-BMI models in clothing is challenging because the mesh self-intersects, causing failures of the cloth simulation.
Future work could address this by automatically removing such intersections.
Additionally, there is little motion capture data of obese people.
So we need to retarget AMASS motions \cite{AMASS:ICCV:2019} to high-BMI subjects. 
But this is also problematic.
Naive retargeting of motion from low-BMI bodies to high-BMI bodies results in interpenetration.

Here we use a simple solution to this problem.
Given a motion sequence from AMASS, we first replace the original body shape with a high-BMI body.
Then, we optimize the pose for each frame to minimize the body-body intersection using the code provided by TUCH~\cite{mueller2021tuch}.
Although this resolves interpenetration between body parts, it can create jittery motion sequences.
As a remedy, we then smooth the jittery motion with a Gaussian kernel.
Although this simple solution does not guarantee a natural motion without body-body interpenetration, it is sufficient  to create a good amount of valid motion sequences for larger bodies.
Future work should address the capture or retargeting of motion for high-BMI body shapes.

\paragraph{Skin tone diversity.}

Our skin tones were provided by Meshcapade GmbH and are categorized into several ethnic backgrounds, with skin-tone variety within each category.
To generate \modelname subjects, we sample uniformly from the Meshcapade skins.
This means the final renders are sampled with the following representations
\begin{itemize}
    \item African 20\%,
\item Asian 24\%,
\item Hispanic 6\%,
\item Indian 20\%,
\item Mideast 6\%,
\item South East Asian 10\%,
\item White 14\%.
\end{itemize}
The same proportions hold in the training, validation and test sets.

\paragraph{Motion sampling.}
Due to the imbalanced distribution of motions in AMASS, we use the motion labels from BABEL \cite{BABEL:CVPR:2021} to sample the motions for a wide and even coverage of the motion space.
After visualizing the motions in each labelled category, we manually assign the number of motions sampled from each category.
Specifically, we sample 64 sequences for motions such as ``turn", ``cartwheel", ``bend", ``sit ", ``touch ground", etc.
We sample 4 sequences from motion labels containing less pose variation, such as ``draw", ``smell", ``lick", ``listen ", ``look", etc.
We do not sample any sequences from labels indicating static poses, for example, ``stand", ``a pose", and ``t pose''.
For the remaining motion labels, we sample 16 random sequences from each.
Each sampled motion sequence lasts from 4 to 8 seconds.

\begin{figure}
    \centering
    \includegraphics[trim=000mm 000mm 000mm 000mm, clip=true, width=1.00 \linewidth]{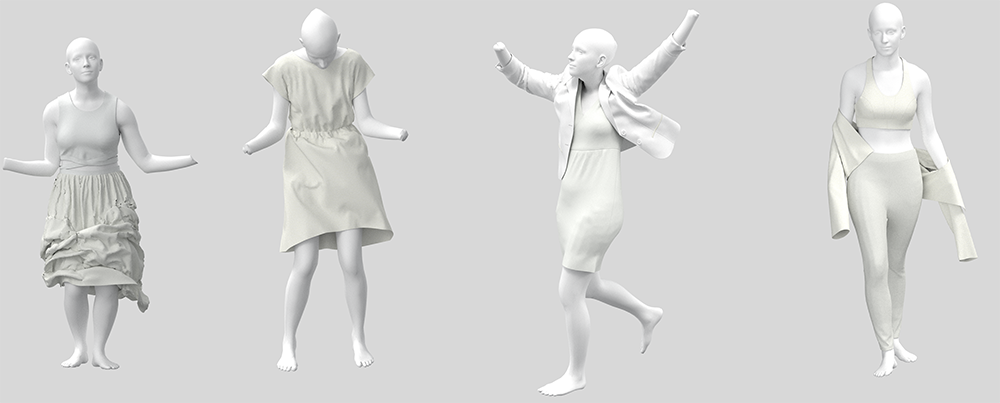}
    \caption{Clothing deformation is well modeled by physics-based simulation.}
    \label{fig:clothing_simulation}
\end{figure}

\paragraph{Clothing.}
Our outfits are designed to reflect  real-world clothing complexity. 
We have layered garments and detailed structures such as pleats and pockets.
We also have open jackets and many wide skirts, which usually have large deformation under different body motion.
These deformations can only be well modeled with a physics-based simulation. See ~\cref{fig:clothing_simulation} for examples.

\paragraph{Putting multiple people in the scene.}

\begin{figure}[t]
 \centerline{
 \includegraphics[trim=000mm 000mm 000mm 000mm, clip=true, width=1.00 \linewidth]{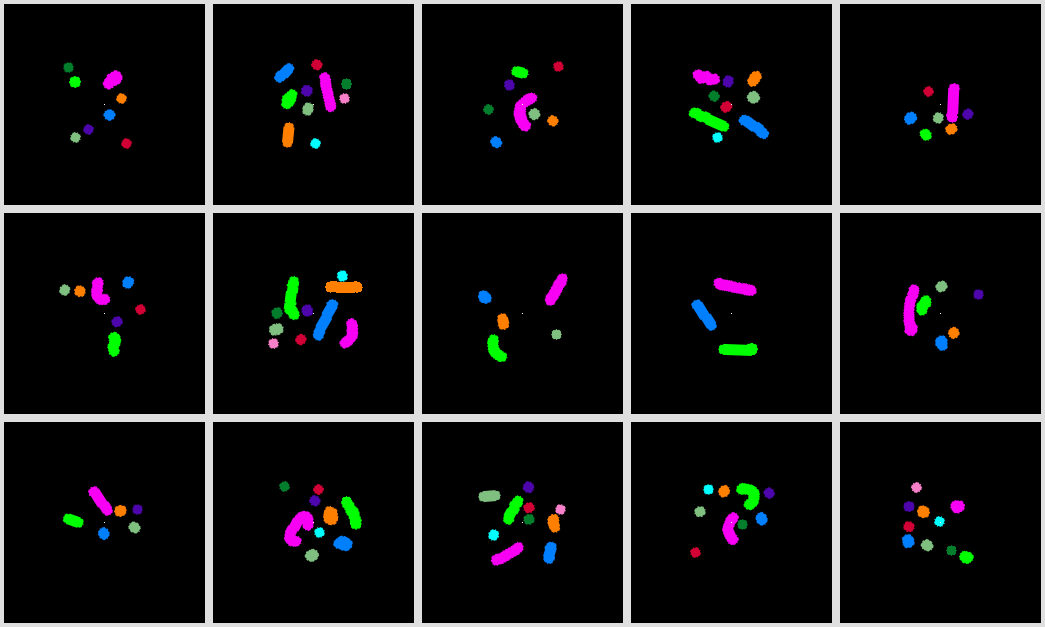}
 }
\caption{Examples of animation ground trajectories. Top-view pelvis trajectories, color coded by subject. These trajectories are automatically placed so that the bodies do not collide. Here, 15 sample sequences are shown with varying numbers of subjects.}
	\label{fig:ground_trajectories}
\end{figure}

For each sequence we randomly select between 1 and 10 subjects. 
For each subject a random animation sequence is selected. 
The shortest animation sequence determines the image sequence length to ensure that there are no ``frozen" body poses. % or animation looping mismatches in a sequence. 
We then pick a random sub-motion of the desired sequence length from each body motion in the sequence. Next the body motions are placed in a desired target area of the scene at a randomized position with a randomized camera yaw. 
To avoid overlapping body motions and collisions with the 3D environment,  we use 2D binary ground plane occupancy masks of the pelvis location for each randomly placed motion. 
The order of motion placement is determined by the ground plane pelvis coverage bounding box. 
This ensures that walking motions, which are challenging to place in a limited space, have the maximum free ground space available before more constrained motions fill the remaining space; cf.~\cite{bazavan2021hspace}. 
Generated root trajectories can be seen in Fig.~\ref{fig:ground_trajectories}.
This is a simple strategy (cf.~\cite{bazavan2021hspace}) and future work should explore the generation or placement of motions that make more sense together and with respect to the scene.
One direction would use MIME \cite{yi2022mime} to take human motions and produce 3D scenes that are consistent with them.

\paragraph{Additional limitations: Hair and shadows.}
Designing high-quality hair assets requires experienced artists. 
Here we used a commercial hair solution based on ``hair cards"; these are simpler than strand-based methods.
The downside is that they require the use of temporal accumulation buffers in the deferred rendering system. 
This can introduce ghosting artefacts when rendering fast motions at low frame rates.
We also observed hair shader illumination issues under certain conditions. When used with the new real-time global illumination system (Lumen) in Unreal Engine 5 (UE5), some hairstyles exhibit a strong hue shift. 
Also, the number of hair colors that we have is limited.
When used in the HDRI environments, with ray traced HDRI shadows enabled, most hairstyles turn black. 
For this reason we do not use ray traced HDRI shadows in the HDRI environment renders, though the 3D scenes do have cast shadows. 
Adding ground contact shadows to the HDRI scenes would require the 
%If ground contact shadows are required then a potential approach is to 
use of a separate ground shadow caster render pass to composite the shadow into the image.
We have not pursued this because we plan to upgrade the hair assets to remove these issues for future releases of the dataset.

\paragraph{Other body models.}
BEDLAM is designed around \smplx but many methods in the field use \smpl \cite{SMPL:2015}.  
In particular, most, if not all, current methods that process video sequences are based on \smpl and not \smplx.
We will provide the ground truth in \smpl format as well for backward compatibility.
We also plan to support other body models like GHUM \cite{xu2020ghum} or SUPR \cite{SUPR} in the future.

\begin{figure*}
    \includegraphics[trim=000mm 000mm 000mm 000mm, clip=true, width=1.0\linewidth]{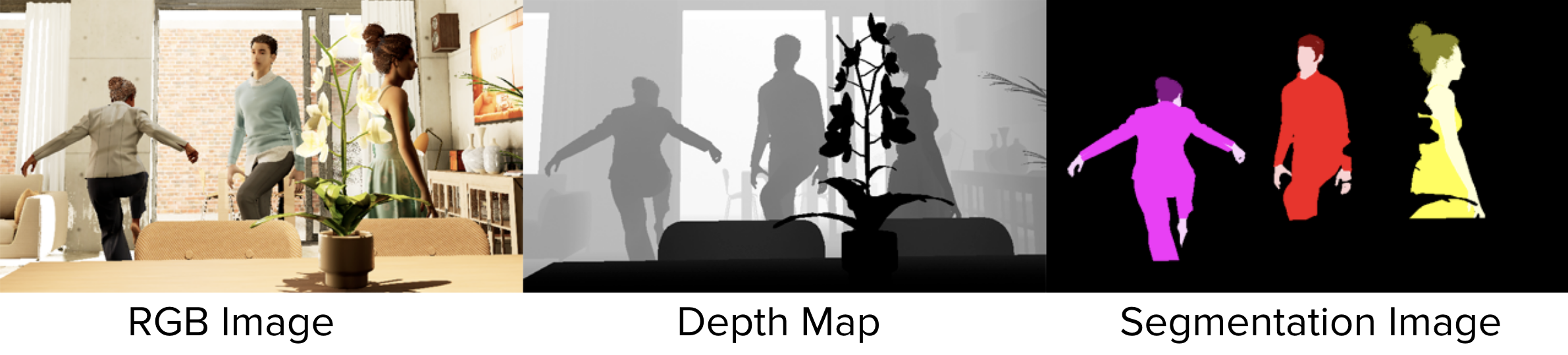}
     \vspace{-1.7 em}
    \caption{
               Additional ground truth: Depth maps and semantic segmentation masks. 
               The segmentation maps are color coded for each individual and each material type (hair, clothing, skin). 
    }
  %  \vspace{-1.5 em}
    \label{fig:reb_seg_mask}
\end{figure*}

\paragraph{Additional ground truth data: Depth maps and semantic segmenation.}
Since BEDLAM is rendered with UE5, we can render out more than RGB images.
In particular, we render depth maps and segmentation masks as illustrated in \cref{fig:reb_seg_mask}.
The segmentation information includes semantic labels for hair, clothing and skin.
With these additional forms of ground truth, BEDLAM can be used to train and evaluate methods that regress depth from images, fit bodies to RGB-D data, perform semantic segmentation, etc.

\paragraph{Assets.}
We will make available the rendered images and the SMPL-X ground truth.
We also release the 3D clothing and clothing textures as well as the skin textures.
We also will make available the process to create more data. 
All assets used are  described in Table \ref{tab:used-assets}.
The table provides a ``shopping list" to recreate BEDLAM.
The only asset that presents a problem for recreating BEDLAM is the hair since new licenses of the the hair assets prohibit training of neural networks (we acquired the data under an older license).
This motivates us to develop new hair assets with an unrestricted license.
More information about how to create new data is provided on the project website.

\begin{table*}[t]
	\centering
	\resizebox{\linewidth}{!}{%
	\begin{tabular}{l|l|l}
        \toprule
		Asset Type & Name & Source \\
		\midrule
		Body Texture & Various & Meshcapade GmbH, https://meshcapade.com\\
            Clothing Texture & Various & WowPatterns, https://www.wowpatterns.com/\\
		Hair & Prime Hairstyles & Reallusion, https://www.reallusion.com/ContentStore/Character-Creator/Pack/Prime-hairstyles/\\
		Hair & Trendy Hairstyles for Men Vol. 1 & Reallusion, https://www.reallusion.com/ContentStore/Pack/universal-hairstyles-vol-1\\
		Hair & Trendy Hairstyles for Men Vol. 2 & Reallusion, https://www.reallusion.com/ContentStore/Pack/universal-hairstyles-vol-2\\
		Environment - HDRI & Various free HDRIs & Poly Haven, CC0 1.0 Universal Public Domain Dedication, https://polyhaven.com/hdris\\		
		Environment - 3D & ArchViz User Interface 3 &  https://www.unrealengine.com/marketplace/en-US/product/archviz-user-interface-3\\	
		Environment - 3D & Big Office & https://www.unrealengine.com/marketplace/en-US/product/big-office\\		
		Environment - 3D & High School Basketball Gym & https://www.unrealengine.com/marketplace/en-US/product/high-school-basketball-gym-day-night-afternoon-midnight-lighting\\		
		Environment - 3D & Sports Stadium &  https://www.unrealengine.com/marketplace/en-US/product/sports-stadium\\
		Environment - 3D & Suburb Neighborhood House Pack &  https://www.unrealengine.com/marketplace/en-US/product/suburb-neighborhood-house-pack-modular\\
		\bottomrule
	\end{tabular}}
    \vspace{-0.1in}
    	\caption{Third-party assets used for rendering BEDLAM. All 3D environments are from the Unreal Marketplace.}
    	\label{tab:used-assets}
\end{table*}

\begin{table*}[t]
	\centering
	\resizebox{\textwidth}{!}{
		\begin{tabular}{l|r|c|c|l|l|r|r}
			\toprule 
			Dataset & \#Sub & \#Frames & Image & Subj/image & Clothing & Motion & Ground truth\\
			\midrule
			3D HUMANS-Train~\cite{moulding} & 19 & 50K & composite & 1 & captured & $>$15 & SMPL \\
			SURREAL~\cite{varol17_surreal} 	& 145 & $\approx$6.5M & composite & 1  & texture &  $>$ 2000 & SMPL \\
 		Human3.6M~\cite{ionescu2013human36m} & few  & 7.5K & mixed reality & 1   & rigged & unk. & 3D joints\\
			MPI-INF-3DHP-Train~\cite{Mehta17TDV} 	& 8 & $>$1.3M  & mixed/composite & 1 & real &  8+ & 3D joints\\
			MuCo-3DHP~\cite{ds:mucomupots} 	& 8 & $\approx$400K  & mixed/composite & 1-4 & real & 8 & 3D joints \\
   Daněček et al.~\cite{10.1111/cgf.13125} & 10 & unk. & rendered (simple) & 1 &  
 physics   & 20 min & unk. \\
   Liang and Lin \cite{liang2019shape} & 100 & 128K & composite & 1 & physics &  5 seqs & SMPL\\
   BCNet (a) \cite{DBLP:conf/eccv/JiangZHLLB20} & 285 & 13K & composite & 1 & rigged & unk. & SMPL\\
  BCNet (b) \cite{DBLP:conf/eccv/JiangZHLLB20} & 3048 & 17K & composite  & 1 & static physics & 55 & SMPL\\
   Liu et al.~\cite{liu2019temporally} & unk. & 3M & composite & 1 & physics &5k & SMPL\\
			Ultrapose~\cite{yan2021ultrapose} & $>$1000 & $\approx$500K &composite & 1  &  physics & n/a & dense points\\
			3DPeople~\cite{pumarola20193dpeople}& 80 & $\approx$2.5M  & composite & 1  & rigged  &  70 & 3D joints\\
   HSPACE \cite{bazavan2021hspace} & 100 & 1M &  rendered  & 5 avg. & rigged (100) & 100 & GHUM\\
 			GTA-Human~\cite{cai2021playing} & $>$600 & $\approx$ 1.4M & game & 1  & rigged &  20K & SMPL \\
 			{AGORA \cite{patel2020agora}}	& $>$350  & $\approx$18K  & rendered & 5-15 & scans &  n/a & SMPL-X, SMPL\\
\midrule
 			{\modelname (ours)}	& 217 & 380K  & rendered  & 1-10 &  physics (110) & 2311& SMPL-X\\
			\bottomrule
		\end{tabular}
}
\vspace{-0.1in}
	\caption{ Comparison of synthetic human datasets that provide images with 3D human pose annotations. See text.
		}
	\label{tab:dataset-comparison}
\end{table*}

\section{Comparison to other datasets}

Table \ref{tab:dataset-comparison} compares synthetic datasets mentioned in the related work section of the main paper.
Here we only survey methods that provide images with 3D ground truth; this excludes datasets focused solely on 3D clothing modeling.
Some of the listed datasets are not public but we include them anyway and some information is not provided in the publications (``unk."~in the table).

Methods vary in terms of the number of subjects, from a handful of bodies to over 1000 in the case of Ultrapose. 
Ultrapose, however, is not guaranteed to have realistic bodies and the dataset is biased towards mostly thin Asian bodies.  The released dataset also has blurred faces.
The number of frames also varies significantly among datasets.
To get a sense of the diversity of images, one must multiply the number of frames by the average number of subjects per image (Sub/image).

The methods vary in how images are generated.
The majority composite a rendered 3D body onto an image background.
This has limited realism.
Human3.6M has mixed reality data in which simple graphics characters are inserted into real scenes using structure from motion.
Mixed/composite methods capture images of real people with a green screen in a multi-camera setup. They can then get pseudo-ground tuth and composite the original images on new backgrounds.
In the table, ``rendered" means that the synthetic body is rendered in a scene (HDRI panorama or 3D model) with reasonable lighting. These are the most realistic methods.

Clothing in previous datasets takes several forms.
The simplest is a texture map on the SMPL body surface (like in SURREAL \cite{varol17_surreal}).
Some methods capture real clothing or use scans of real clothing.
Another class of methods uses commercial ``rigged" models with rigged clothing.
This type of clothing lacks the realism of physics simulation.
Most methods that do physics simulation use a very limited number of garments (often as few as 2) due to the complexity and cost.

It is hard to get good, comparable, data about motion diversity in these datasets.
Here we list numbers of motions gleaned from the papers but these are quite approximate.
Some of the low numbers describe classes of motions that may be repeated with some unknown number of variations.
At the same time, some of the larger numbers may lack divesity.
With BEDLAM, we are careful to sample a diverse set of motions.

For comparison with real-image datasets, 3DPW contains 60 sequences captured with a moving camera, with roughly 51K frames, and 7 subjects in a total of 18 clothing styles.
With roughly 2 subjects per frame, this gives around 100K unique bounding boxes.
Human3.6M training data has 1,464,216 frames captured by 4 static cameras at 50 fps, which means there are 366K unique articulated poses. If one reduces the frame rate to 30 fps, that gives roughly 220K bounding boxes of 5 subjects performing 15 different types of motions.  
We observe that the total number of frames is less important than the diversity of those frames in terms of scene, body, pose, lighting, and clothing.

\section{Implementation Details}

\paragraph{BEDLAM-CLIFF-X.}
Since most HPS methods output \smpl bodies, we focus on that in the main paper and describe the \smplx methods here.
Specifically, we use BEDLAM hand poses to train a full body network called BEDLAM-CLIFF-X. 
For this, we train a separate hand network on hand crops from BEDLAM with an HMR architecture but replace SMPL with the MANO hand \cite{mano}, which is compatible with \smplx. 
We merge the body pose output $\theta_b \in \mathbb{R}^{22 \times 3}$ from  BEDLAM-CLIFF (see Sec.~4.1 of the main paper) and hand pose output $\theta_h \in \mathbb{R}^{16 \times 3}$ from the hand network to get the full body pose with articulated hands $\theta_{fb} \in \mathbb{R}^{55 \times 3}$. 
The face parameters, $\theta_{jaw}$, $\theta_{leye}$ and $\theta_{reye}$ are kept as neutral.
Since both BEDLAM-CLIFF and the hand network output different wrist poses, we cannot merge them directly. 
Hence, we train a small regressor  $R_{fb}$  to combine them.

Specifically, we define the body pose  $\theta_b$ = \{$\hat{\mathbf{\theta}}_b$, $\theta_{elbow}$, $\theta_{wrist}^b$ \} and and hand pose $\theta_h$ = \{$\theta_{wrist}^h$\, $\theta_{fingers}$\}, where $\hat{\mathbf{\theta}}_b \in \mathbb{R}^{20 \times 3}$ represents the first 20 pose parameters of \smplx.
$R_{fb}$ takes global average pooled features as well as  $\theta_b$ and $\theta_h$ from the BEDLAM-CLIFF and hand networks, and outputs $\theta_{fb}$ = \{$\hat{\mathbf{\theta}}_b$, $\theta_{elbow}$ + $\Delta_{elbow}$, $\theta_{wrist}^b$+$\Delta_{wrist}$, $\theta_{fingers}$ \}. 
Basically, $R_{fb}$ learns an update  of the elbow and wrist pose from the body network using information from both the body and hand network.
Since we learn only an update on the wrist pose generated by the body network, this prevents the unnatural bending of the wrists.
% unlike the methods directly regressing the wrist pose. pose\cite{Choutas2020_expose,feng2021pixie}.
Similar to BEDLAM-CLIFF, to train BEDLAM-CLIFF-X, we use a combination of MSE loss on model parameters, projected keypoints, 3D joints, and an L1 loss on 3D vertices. All other details can be found the code (see project page).

\paragraph{Data augmentation.}

A lot of data augmentation is included during training, including random crops, scale, different kinds of blur and image compression, brightness and contrast modification, noise addition, gamma, hue and saturation modification, conversion to grayscale, and downscaling using \cite{info11020125}.

\begin{table}[t]
	\scriptsize
	\centering
%	\begin{center} 
	\resizebox{\columnwidth}{!}{
 \begin{tabular}{l|c|c|c|c|c|c}
	\toprule
	\multicolumn{1}{c|}{Method} & \multicolumn{1}{c|}{Dataset} & \multicolumn{1}{c|}{Backbone}  & \multicolumn{1}{c|}{Crops \%} & \multicolumn{1}{c|}{PA-MPJPE} & \multicolumn{1}{c|}{MPJPE} & \multicolumn{1}{c}{PVE} \\
	\midrule
	HMR & B+A & scratch & 100 & 67.9 & 108.8 & 129.0\\
	HMR & B+A & ImageNet & 100  & 57.3 & 91.7 & 108.8 \\
	HMR &  B+A & COCO & 100 & 47.6 & 79.0 & 93.1\\	
	\midrule
	CLIFF & B+A & scratch & 100 & 61.7 & 96.5 & 115.0\\
	CLIFF &  B+A & ImageNet & 100 & 51.8 & 82.1 & 96.9\\
	CLIFF &  B+A & COCO & 100 & 47.4 & 73.0 & 86.6 \\
	\midrule
	HMR &  B & COCO & 5 &  55.8 & 86.9 & 104.3\\
	HMR &  B & COCO & 10 & 55.5 & 85.7 & 102.9\\
	HMR &  B & COCO & 25 &  53.9 & 83.9 & 100.4\\
	HMR &  B & COCO & 50 &  53.8 & 81.1 & 97.3 \\
	HMR &  B+A & COCO & 100 & 47.6 & 79.0 & 93.1\\	
	\midrule
	CLIFF &  B & COCO & 5 &  54.0 & 80.8 & 96.8 \\
	CLIFF &  B & COCO & 10 &  53.8 & 79.9 & 95.7\\
	CLIFF &  B & COCO & 25 & 52.2 & 77.7 & 93.6\\
	CLIFF &  B & COCO & 50 & 51.0 & 76.3 & 91.1 \\
	CLIFF &  B+A & COCO & 100 & 47.4 & 73.0 & 86.6 \\
	\midrule
	HMR &  A & COCO & 100 &  58.3 & 94.9 & 109.0\\	
	HMR &  B & COCO & 100 & 51.2 & 80.6 & 96.1\\
	HMR &  B+A & COCO & 100 & 47.6 & 79.0 & 93.1\\	
	\midrule
	CLIFF &  A & COCO & 100 & 54.0 & 88.0 & 101.8\\	
	CLIFF &  B & COCO & 100 & 50.5 & 76.1 & 90.6\\
	CLIFF &  B+A & COCO & 100 & 47.4 & 73.0 & 86.6 \\
\bottomrule
\end{tabular}
}
%	\end{center} 
\vspace{-0.08in}
	\caption{Ablation experiments on 3DPW. B denotes BEDLAM and A denotes AGORA. Crops \% only applies to BEDLAM.}
	\label{tab:ablation1supmat}
\vspace{-0.1in}
\end{table}

\section{Supplemental experiments}

\subsection{Ablation of training data and backbones}
Table \ref{tab:ablation1supmat} expands on Table 3 from the main paper, providing the full set of dataset ablation experiments.
The key takeaways are: (1) training with a backbone pretrained on the 2D pose-estimation task on COCO produces the best results, 
(2) training from scratch on BEDLAM does not work as well as either pre-training on ImageNet or COCO,
(3) training only on BEDLAM is better than training only on AGORA,
(4) training on BEDLAM+AGORA is consistently better than using either alone (note that both are synthetic),
(5) one can get by with using a fraction of BEDLAM (50\%\/ or even 25\%\/ gives good performance), but training error continues to decrease up to 100\%\/.
All of this suggest that there is still room for improvement in the synthetic data in terms of variety.

\subsection{Ablation on losses}
To understand which loss terms are important,
we perform an ablation study on standard losses used in training HPS methods including $L_{\text{SMPL}}$,  $L_{{j3d}}$, $L_{{j2d}}$,  $L_{{v3d}}$,  $L_{{v2d}}$. Individual losses are described here and the ablation on them is reported in Table \ref{tab:loss-ablation}.
\begin{equation}
 L_{\text{SMPL}} = \lVert \hat{\theta} - \theta \rVert + \lVert \hat{\beta} - \beta \rVert \nonumber \\
 \end{equation}
 \begin{equation}
  L_{{j3d}} = \lVert \hat{\mathcal{J}} - \mathcal{J} \rVert \nonumber\\
  \end{equation}
  \begin{equation}
  L_{{j2d}} = \lVert \hat{j} - j \rVert \nonumber\\
 \end{equation}
 \begin{equation}
 L_{{v3d}} = \lVert \hat{\mathcal{V}} - \mathcal{V} \rVert \nonumber\\
 \end{equation}
 \begin{equation}
 L_{{v2d}} = \lVert \hat{v} - v \rVert \nonumber\\
 \end{equation}
$\hat{x}$  denotes the ground truth for the corresponding
variable $x$ and $\lVert \cdot \rVert$ is the type of loss that can be L1 or L2. For shape we always use L1 norm. $\mathcal{J}$, $\mathcal{V}$,  $\beta$ and $\theta$ denote the 3D joints, 3D vertices, shape and pose parameters of \smplx model respectively. $j$ and $v$ denote the 2D  joints and vertices projected into the full image using the predicted camera parameters similar to \cite{li2022cliff}. $\theta$ is predicted in a 6D rotation representation form \cite{zhou2019continuity} and converted to a 3D axis-angle representation when passed to SMPL-X model. Since we set the hand poses to neutral in BEDLAM-CLIFF, we use only the first 22 pose parameters  in the training loss. We use a subset of BEDLAM training data for this ablation study.
Note that, to compute $L_{{v2d}}$ we use a downsampled mesh with 437 vertices, computed using the downsampling method in \cite{ranjan2018generating}.
We find this optimal for training speed and performance. Since the downsampling module samples more vertices in regions with high curvature, it helps preserve the body shape and we can store the sampled vertices directly in memory without the need to load them during training. We include a 2D joints loss in all cases as it is necessary to obtain proper alignment with the image.

As shown in Table \ref{tab:loss-ablation}, $L_{j3d}$ or $L_{v3d}$ alone do not provide enough supervision for training. Similar to \cite{pang2022benchmark} we find that $L_{\text{SMPL}}$ provides stronger supervision reducing the loss by a large margin when used in combination with $L_{v3d}$ and $L_{j3d}$. Surprisingly, we find that including $L_{v2d}$ makes the performance slightly worse. 
A plausible reason for this could be that using  $L_{v2d}$ provides high weight on aligning the predicted body to the image but the mismatch between the ground truth and estimated camera used for projection during inference makes the 3D pose worse, thus resulting in higher 3D error. We suspect that $L_{v2d}$ could provide strong supervision in the presence of a better camera estimation model; this is future work.

We also experiment with two different types of losses, L1 and MSE and find that L1 loss yields lower error on the 3DPW dataset as shown in Table~\ref{tab:loss-ablation}. However, Table~\ref{tab:l1vsl2} shows that the model using L1 loss performs worse when estimating body shape on the SSP and HBW datasets compared to the model using MSE loss. This discrepancy may be attributed to the L1 loss treating extreme body shapes as outliers, thereby learning only average body shapes. Since the 3DPW dataset does not have extreme body shapes, it benefits from the L1 loss. Consequently, we opted to use the MSE loss for our final model and all results reported in the main paper. 
Note that $L_{j3d}$ or $L_{v3d}$ alone is worse with L1 loss compared to MSE loss.

\begin{table}[t]
\centering
	\resizebox{\columnwidth}{!}{
\begin{tabular}{l|c|ccc}
\toprule
\multicolumn{1}{c|}{Losses} & \multicolumn{1}{c|}{Type} & \multicolumn{1}{c}{PAMPJPE} & \multicolumn{1}{c}{MPJPE}  & \multicolumn{1}{c}{MVE} \\
\midrule
$L_{{j3d}}$ & MSE &59.1 & 86.1 & 105.1\\
$L_{{v3d}}$ & MSE &56.2 & 83.4 & 96.7 \\
 $L_{\text{SMPL}}$ &MSE  & 51.3 & 83.8 & 96.7\\
$L_{\text{SMPL}}$ + $L_{{j3d}}$ & MSE &48.5 & 76.0 & 89.6\\
$L_{\text{SMPL}}$ + $L_{{v3d}}$ & MSE &48.2 & 74.7 & 87.9\\
$L_{\text{SMPL}}$ + $L_{{v3d}}$ + $L_{{j3d}}$ & MSE & \textbf{47.6} & \textbf{74.2} & \textbf{87.2}\\
$L_{\text{SMPL}}$ + $L_{{v3d}}$ + $L_{{j3d}}$ + $L_{{v2d}}$ & MSE & 48.7 & 74.4 & 87.6\\
 \midrule
  $L_{{j3d}}$ & L1 &59.4 & 85.7 & 114.6\\
 $L_{{v3d}}$ & L1 & 72.5 & 97.4 & 111.6\\
 $L_{\text{SMPL}}$ & L1  & 50.6 & 83.6 & 96.0\\
 $L_{\text{SMPL}}$ + $L_{{j3d}}$ & L1 &46.9 & 74.7 & 87.6\\
 $L_{\text{SMPL}}$ + $L_{{v3d}}$ & L1 &48.8 & 76.2 & 88.8\\
 $L_{\text{SMPL}}$ + $L_{{v3d}}$ + $L_{{j3d}}$ & L1 & \textbf{46.9} & \textbf{73.0} & \textbf{86.0}\\
 $L_{\text{SMPL}}$ + $L_{{v3d}}$ + $L_{{j3d}}$ + $L_{{v2d}}$ & L1 & 47.4 & 73.5 & 86.8\\
 \bottomrule
\end{tabular}}
	\caption{{\bf Ablation of different losses.} Error on 3DPW in mm.}
	\label{tab:loss-ablation}
\end{table}

\begin{table}[t]
\centering
	\resizebox{\columnwidth}{!}{
\begin{tabular}{l|c|ccccc}
\toprule
    \multicolumn{1}{c|}{\footnotesize{Loss type}} & \multicolumn{1}{c|}{\footnotesize{SSP-3D}} & \multicolumn{5}{c}{\footnotesize{HBW}} \\
    \midrule
       & \multicolumn{1}{c|}{\footnotesize{PVE-T-SC}} & \multicolumn{1}{c}{\footnotesize{Height}} &\multicolumn{1}{c}{\footnotesize{Chest}} &\multicolumn{1}{c}{\footnotesize{Waist}}&\multicolumn{1}{c}{\footnotesize{Hips}} &\multicolumn{1}{c}{\footnotesize{$\text{P2P}_{\text{20k}}$}}  \\
       \midrule
       L1 &  15.1 & {51} & 73 & 97 & 64 & 22 \\
       MSE & {14.2} & {51} & {69} & {88} & {62} & 22\\

\bottomrule
\end{tabular}}
	\caption{{\bf Losses.} The use of L2 or L1 losses are explored for shape estimation accuracy  using BEDLAM-CLIFF: error on HBW \cite{mueller2021tuch} and SSP-3D \cite{sengupta2020straps} in mm.}
	\label{tab:l1vsl2}
\end{table}

\subsection{Ablation of dataset attributes}
We also perform an ablation study by varying different dataset attributes. 
We generated 3 different sets of around 180K images by varying the use of different assets. Keeping the scenes and the motion sequences exactly the same, we experiment by ablating hair and then further replacing the cloth simulation with simple cloth textures. 
We  use a backbone pretrained with either COCO \cite{lin2014coco} or ImageNet and study the performance on 3DPW \cite{vonMarcard18ECCV}. 
When using the ImageNet backbone, we find that training with clothing simulation leads to better accuracy than training with clothing texture mapped onto the body.
Adding hair gives a modest improvement in MPJPE and MVE.
Surprisingly, with the COCO backbone, the difference in the training data makes less difference.
Still, clothing simulation is consistently better than just using clothing textures.
It is likely that the backbone pretrained on a 2D pose estimation task using COCO is already robust to clothing and hair. 
As mentioned above, however, our hair models are not ideal and not as diverse as we would like. 
Future work, should explore whether more diverse and complex hair has an impact.
\begin{table}[t]
\centering
	\resizebox{\columnwidth}{!}{
\begin{tabular}{l|c|ccc}
\toprule
\multicolumn{1}{c|}{Dataset attribute} & \multicolumn{1}{c|}{Backbone} &  \multicolumn{1}{c}{PAMPJPE} & \multicolumn{1}{c}{MPJPE}  & \multicolumn{1}{c}{MVE} \\
\midrule
Simulation + Hair & ImageNet & {65.6} &  {101.8} & {120.8}\\
Simulation & ImageNet & 66.3 & 104.5 & 124.5\\
Texture & ImageNet & 72.2 & 116.1 & 136.7 \\
\midrule
Simulation + Hair & COCO & {51.6} & {77.8} & {92.4} \\
Simulation & COCO & {51.6} & 78.7 & 93.0 \\
Texture & COCO & 54.3 & 80.8 & 96.0 \\

\bottomrule
\end{tabular}}
	\caption{{\bf Ablation of different dataset attributes.} Error on 3DPW in mm. See text.}
	\label{tab:attribute-ablation}
\end{table}

\begin{table}[t]
% 	\scriptsize
%\small
	\resizebox{\columnwidth}{!}{
		\begin{tabular}{l|cc|ccc}
			\toprule
			\multicolumn{1}{c|}{Method} & 
			\multicolumn{2}{c|}{H3.6M}&         
			\multicolumn{3}{c}{3DPW}\\
			\midrule
			& \multicolumn{1}{l}{\footnotesize PA-MPJPE} & \multicolumn{1}{l}{\footnotesize MPJPE} & 
			\multicolumn{1}{l}{\footnotesize PA-MPJPE} &
			\multicolumn{1}{l}{\footnotesize MPJPE} &
   			\multicolumn{1}{l}{\footnotesize PVE}\\
                \midrule
                CLIFF\cite{li2022cliff} & 32.7 &47.1 & - & - & - \\
                CLIFF\textsuperscript{\textdagger}* & 39.4 & 62.9 & 43.6 & 68.8 & 82.1\\
                CLIFF\textsuperscript{\textdagger}* w/o H3.6M & 56.1 & 89.6 & 44.4 & 68.9 & 82.3\\
                BEDLAM-HMR & 51.7 & 81.6 & 47.6 & 79.0 & 93.1 \\
                BEDLAM-CLIFF & 50.9 & 70.9 & 46.6 & 72.0 & 85.0\\
			\bottomrule						
   \end{tabular}
		}
    \caption{{\bf Impact of training without Human3.6M on Human3.6M and 3DPW.} CLIFF\textsuperscript{\textdagger}* is the same model as Table 1 in main paper.}
	\label{tab:h3.6m-compare}
\end{table}

\begin{table*}[t]
%	\scriptsize
	\centering
%	\begin{center} 
	\begin{tabular}{l|cccc|cccc}
	\toprule
	\multicolumn{1}{c|}{Method} & \multicolumn{4}{c|}{MVE}  & \multicolumn{4}{c}{MPJPE} \\
	\midrule
	& \multicolumn{1}{c}{FB} & \multicolumn{1}{c}{B} & \multicolumn{1}{c}{F} & \multicolumn{1}{c|}{LH/RH} 	& \multicolumn{1}{c}{FB} & \multicolumn{1}{c}{B} & \multicolumn{1}{c}{F} & \multicolumn{1}{c}{LH/RH} \\
	\midrule

    SMPLify-X\cite{Pavlakos2019_smplifyx} & 236.5&	187.0&	48.9&	48.3/51.4 &	231.8&	182.1&	52.9&	46.5/49.6\\
    ExPose\cite{Choutas2020_expose} & 217.3 & 151.5	&51.1&	74.9/71.3&	215.9&	150.4&	55.2&	72.5/68.8 \\
    Frankmocap\cite{rong2021frankmocap} && 168.3	&&	54.7/55.7&&		165.2	&&	52.3/53.1 \\
    PIXIE\cite{feng2021pixie} & 191.8 & 142.2&	50.2&	49.5/49.0&	189.3&	140.3&	54.5&	46.4/46.0 \\
    BEDLAM-CLIFF-X & \textbf{131.0} & \textbf{96.5} & \textbf{25.8} & \textbf{38.8/39.0} & \textbf{129.6} & \textbf{95.9} & \textbf{27.8} & \textbf{36.6/36.7} \\
    \midrule
    Hand4Whole+ \cite{Moon_2022_CVPRW_Hand4Whole} & 135.5 & 90.2&	41.6&	46.3/48.1&	132.6&	87.1&	46.1&	44.3/46.2 \\
    PyMAF+ \cite{pymafx2022} & 125.7 & 84.0	& 35.0 & 44.6/45.6 & 124.6 & 83.2 &	37.9 &	42.5/43.7\\
  %  Ours* (neutral hands) & 112.4 & 82.1 &24.1 & 37.6/38.2 & 111.4 & 81.9 &25.9 & 32.9/33.7 \\
    BEDLAM-CLIFF-X+ & \textbf{103.8} & \textbf{74.5} & \textbf{23.1} & \textbf{31.7/33.2} & \textbf{102.9} & \textbf{74.3} & \textbf{24.7} & \textbf{29.9/31.3} \\

\bottomrule
\end{tabular}
%	\end{center} 
\vspace{-0.07in}
	\caption{{\bf \smplx methods on the AGORA test set.} + denotes methods include AGROA training set. FB is full-body, B is body only, F is face, and LH/RH are the left and right hands respectively.
 }
	\label{tab:agora-evaluation}
\end{table*}

\begin{table*}[t]
	\small
	\centering
%	\begin{center} 
	\begin{tabular}{l|cc|cc|cccc|cccc}
	\toprule
	\multicolumn{1}{c|}{Method}  & \multicolumn{2}{c|}{NMVE}   & \multicolumn{2}{c|}{NMJE} & \multicolumn{4}{c|}{MVE}  & \multicolumn{4}{c}{MPJPE} \\
	\midrule
	& \multicolumn{1}{c}{FB} & \multicolumn{1}{c|}{B} & \multicolumn{1}{c}{FB} & \multicolumn{1}{c|}{B} & \multicolumn{1}{c}{FB} & \multicolumn{1}{c}{B} & \multicolumn{1}{c}{F} & \multicolumn{1}{c|}{LH/RH} 	& \multicolumn{1}{c}{FB} & \multicolumn{1}{c}{B} & \multicolumn{1}{c}{F} & \multicolumn{1}{c}{LH/RH} \\
	\midrule

    PyMAF-X\cite{pymafx2022} & 172.1 & 123.6 & 167.2 & 120.1 & 161.8 & 117.4 & 50.3 & 40.5/42.6 & 157.2 & 114.1 & 51.6 & 38.2/39.7 \\
    Hand4Whole\cite{Moon_2022_CVPRW_Hand4Whole} &178.8 & 119.1 & 176.2 & 117.6 & 168.1 & 112.0 &	59.7 &	52.8/55.8 &	165.7 &	110.5 &	63.7&	50.0/52.0 \\
    PIXIE\cite{feng2021pixie} & 160.0 & 107.2 & 154.8 & 103.5 & 150.4 & 100.8 & 51.4 & 47.2/50.2 & 145.6 & 97.3 & 55.4 & 43.6/46.0\\

    BEDLAM-CLIFF-X & 101.7 & 65.6 & 99.0 & 64.7 & 95.6 & 61.7 & 29.9 & 35.7/36.2 &  93.1 & 60.8 & 30.5 & 33.2/33.3 \\
    
    BEDLAM-CLIFF-X+ & \textbf{93.4} & \textbf{61.2} & \textbf{92.5} & \textbf{60.4} & \textbf{87.8} & \textbf{56.8} & \textbf{27.3} & \textbf{31.9/33.9} &  \textbf{87.0} & \textbf{57.5} & \textbf{28.0} & \textbf{29.5/31.1} \\

\bottomrule
\end{tabular}
%	\end{center} 
\vspace{-0.07in}
	\caption{{\bf \smplx methods on the BEDLAM test set.} Comparison of SOTA methods on the BEDLAM test set. + denotes methods include AGROA training set.
}
	\label{tab:bedlam-evaluation}
\end{table*}

\subsection{Experiment on Human3.6M}

We also evaluate our method on the Human3.6M dataset \cite{ionescu2013human36m} by calculating MPJPE and PA-MPJPE on 17 joints obtained using the Human3.6M regressor on vertices. Previous methods have used Human3.6M training images when evaluating on the test set. 
Specifically, CLIFF \cite{li2022cliff} and our re-implementation, CLIFF\textsuperscript{\textdagger}*, both use Human3.6M data for training and, consequently get low errors on Human3.6M test data.
Note that our implementation does not get as low an error as reported in \cite{li2022cliff} despite the fact that we match their performance on 3DPW and RICH (see main paper).

To ensure a fair comparison and to measure the generalization of the methods, we trained a version of CLIFF (CLIFF\textsuperscript{\textdagger}* w/o H3.6M)  using 3D datasets MPI-INF-3DHP, 3DPW and 2D datasets COCO and MPII but excluding Human3.6M, following the same settings as BEDLAM-CLIFF. 
The results in \cref{tab:h3.6m-compare} demonstrate that BEDLAM-CLIFF outperforms CLIFF when Human3.6M is not included in training.
This is another confirmation of the results in the main paper showing that BEDLAM-CLIFF has better generalization ability than CLIFF. Without using Human3.6M in training, BEDLAM-HMR is also better than CLIFF on Human3.6M.

Note that this experiment illustrates how training on Human3.6M is crucial to getting low errors on that dataset. The training and test sets are similar (same backgrounds and similar conditions) meaning that methods trained on the dataset can effectively over-fit to it.
This can be seen by comparing CLIFF\textsuperscript{\textdagger}* with CLIFF\textsuperscript{\textdagger}* w/o H3.6M.
Training on Human3.6M significantly reduces error on Human3.6M without reducing error on 3DPW. 

\subsection{\smplx experiments on the AGORA dataset}
AGORA is interesting because it is one of the few datasets with \smplx ground truth.
Table~\ref{tab:agora-evaluation} evaluates methods that estimate \smplx bodies on the AGORA dataset. The results are taken from the AGORA leaderboard. 
BEDLAM-CLIFF-X does particularly well on the face and hands. 
Since the BEDLAM training set contains body shapes sampled from AGORA, it gives BEDLAM-CLIFF-X an  advantage over  methods that are not fine-tuned on the AGORA training set (bottom section of \cref{tab:agora-evaluation}).
Consequently, we also compare a version of BEDLAM-CLIFF-X that is trained only on the BEDLAM training set. This still outperforms all the methods that were not trained using AGORA (top section of \cref{tab:agora-evaluation}).
Please see Figure~\ref{fig:bedlam-agora-qualitative} for qualitative results.

\subsection{\smplx experiments on BEDLAM}
For completeness, \cref{tab:bedlam-evaluation} shows that BEDLAM-CLIFF-X outperforms recent SOTA methods that estimate \smplx on the BEDLAM test set. Not surpisingly, our method is more accurate by a large margin. 
Note, however, that the prior methods are not trained on the BEDLAM training data. 
We follow a similar evaluation protocol as \cite{patel2020agora}. 
Since the hands are occluded in a large number of frames, we use MediaPipe \cite{mediapipe} to detect the hands and evaluate hand accuracy only if they are visible. To detect individuals within an image during evaluation, we use the detector that is included in the respective method's demo code. In cases where the detector is not provided, we use \cite{redmon2018yolov3}, the same detector use by BEDLAM-CLIFF-X. 
Please see \cref{fig:bedlam-agora-qualitative} for qualitative results.
															
\section{Qualitative Comparison}

\begin{figure*}
    \centering
%ORIG IMAGE
    \includegraphics[trim=000mm 000mm 033mm 000mm, clip=true, height=0.17\linewidth]{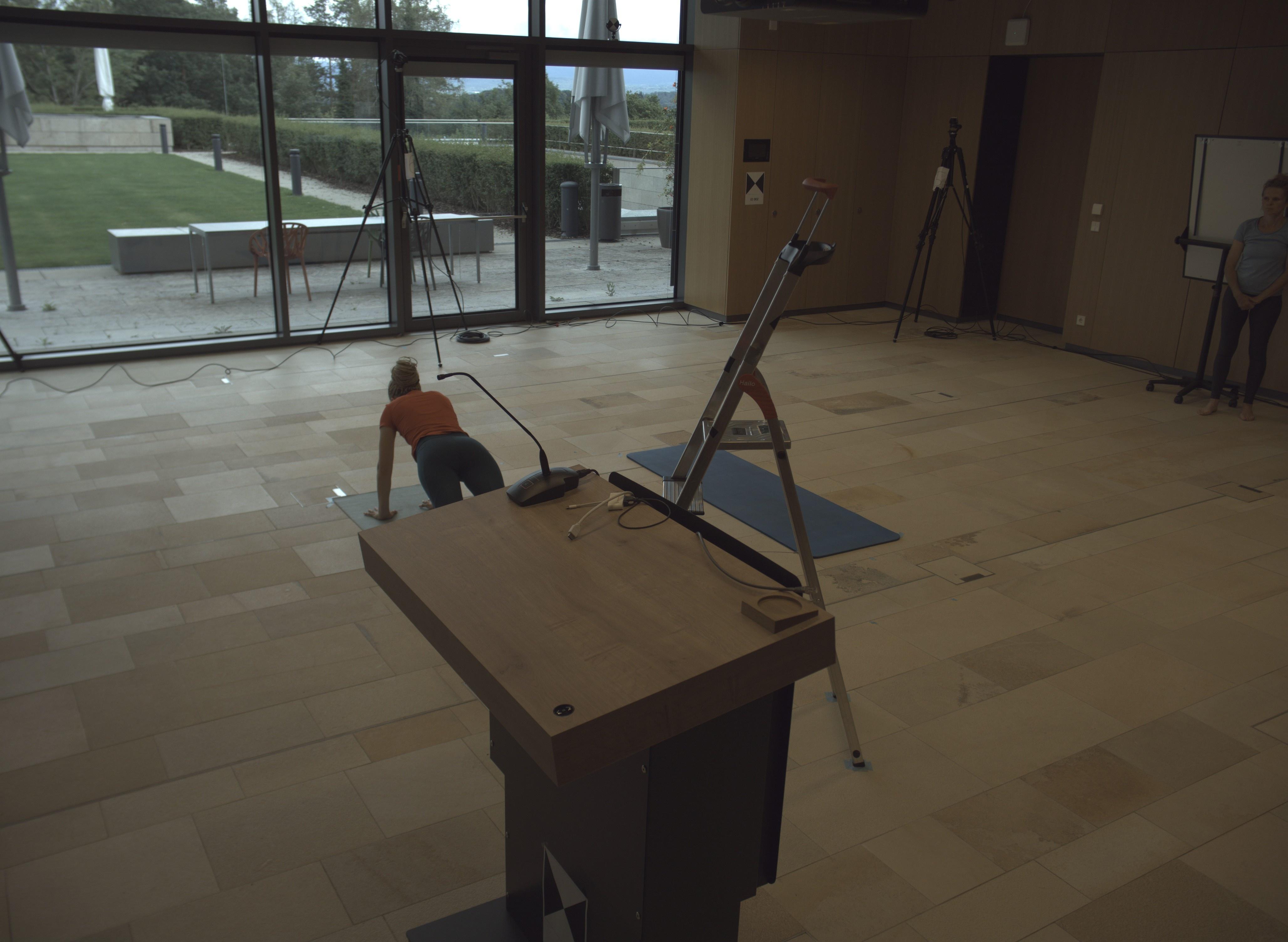}
    \includegraphics[trim=000mm 000mm 033mm 000mm, clip=true, height=0.17\linewidth]{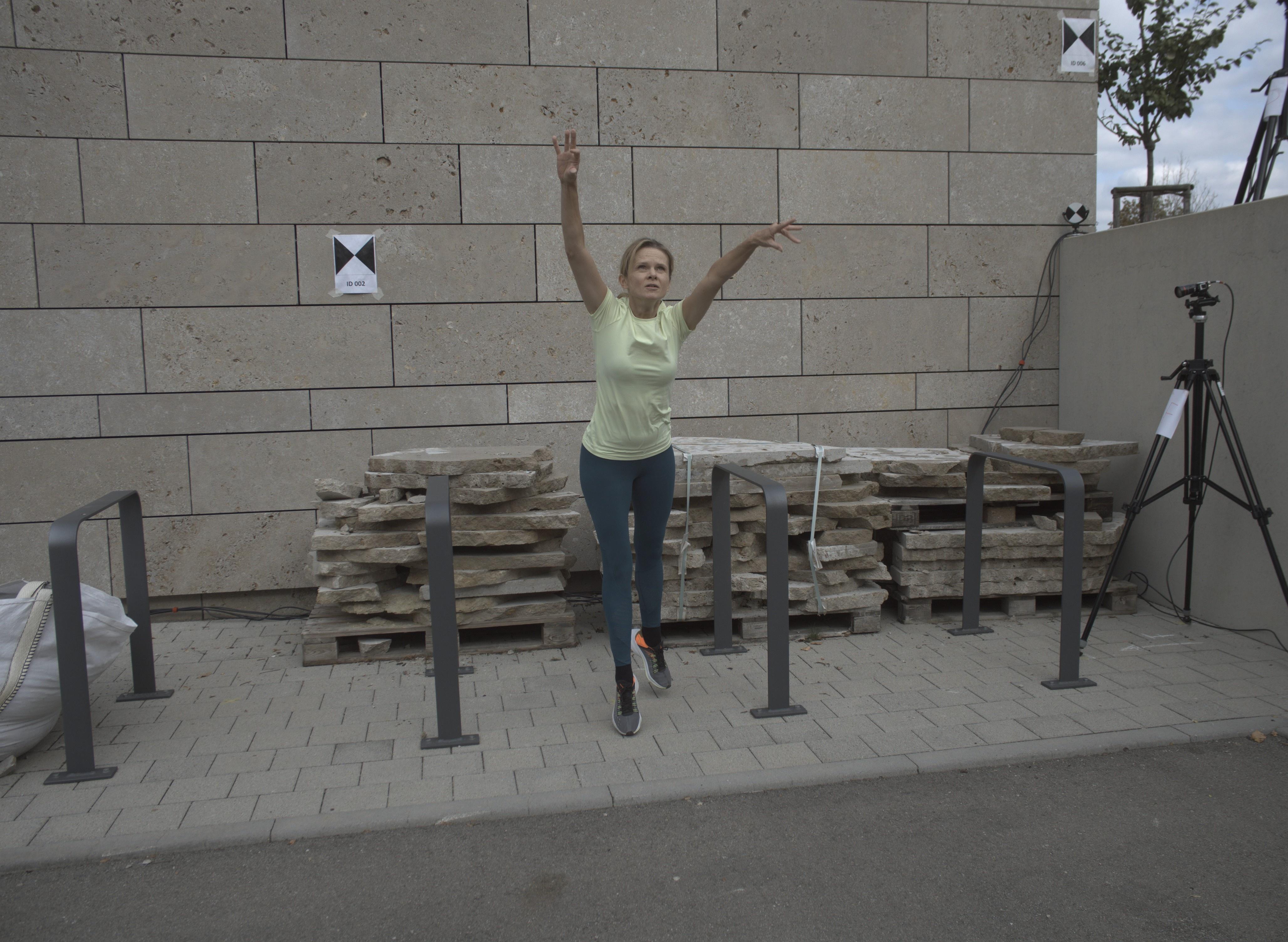}
    \includegraphics[trim=000mm 000mm 169mm 000mm, clip=true, height=0.17\linewidth]{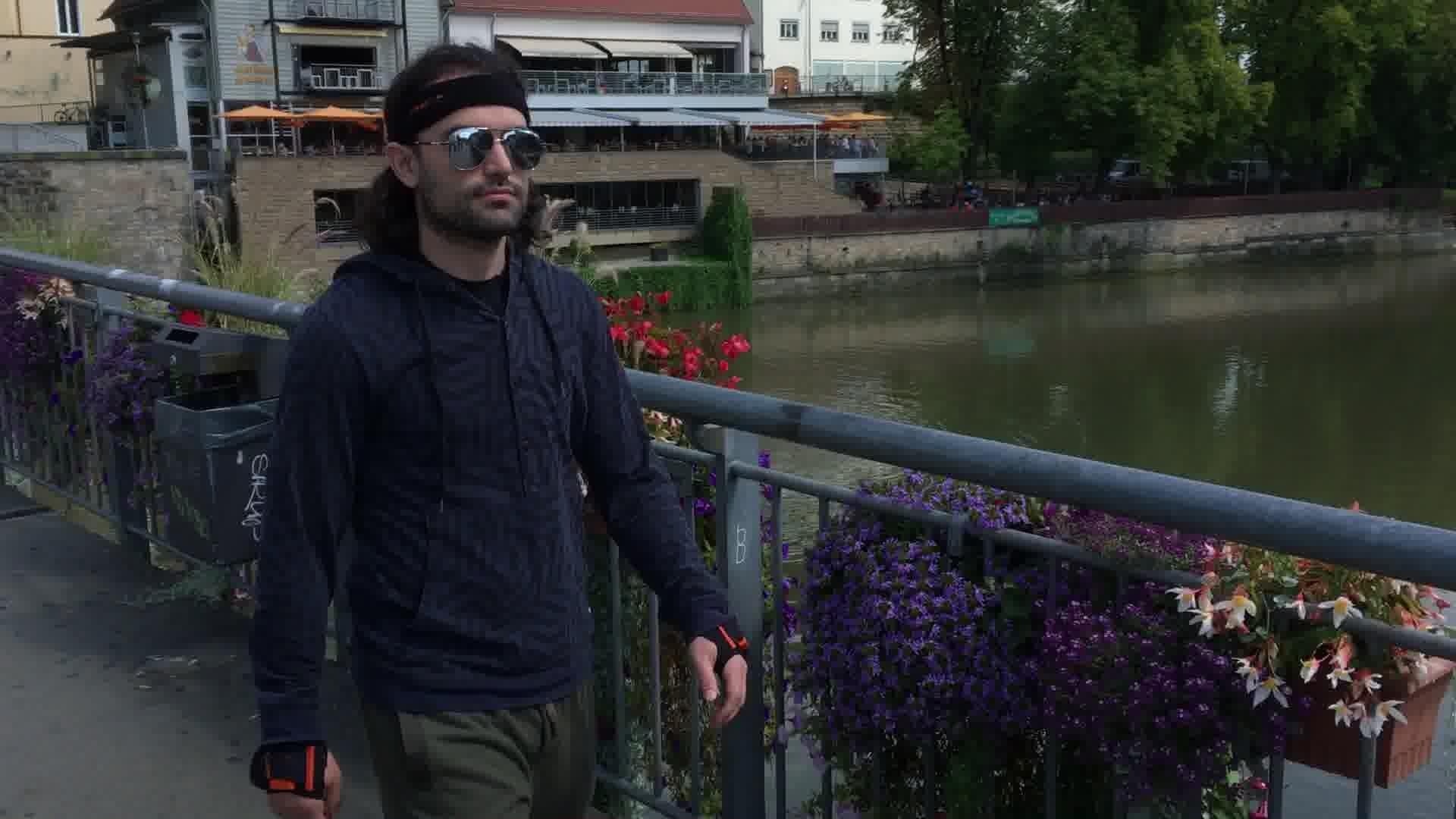}
    \includegraphics[trim=050mm 000mm 119mm 000mm, clip=true, height=0.17\linewidth]{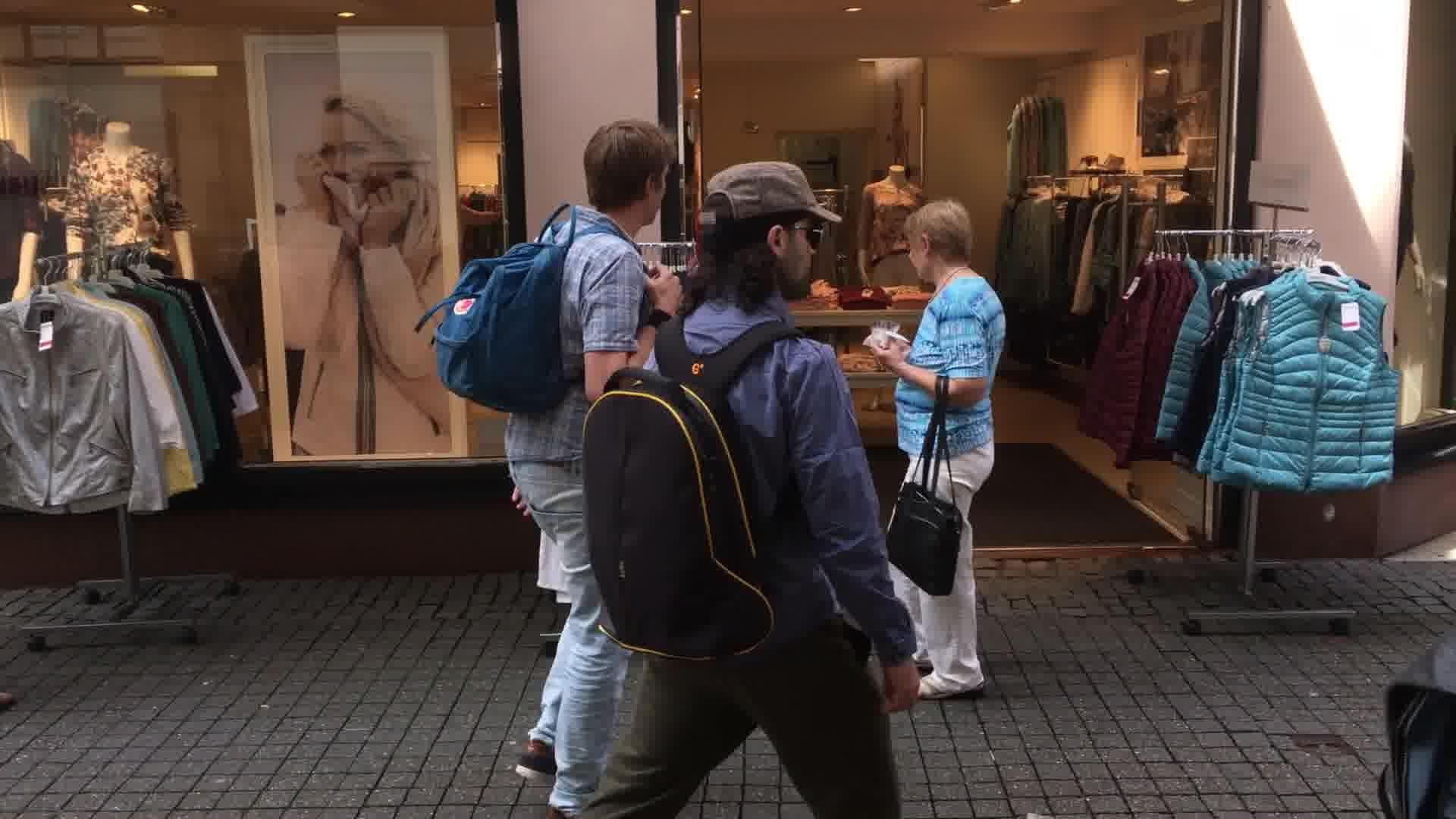}
    \\
%ORIG VIEW

     \includegraphics[trim=000mm 000mm 033mm 000mm, clip=true, height=0.17\linewidth]{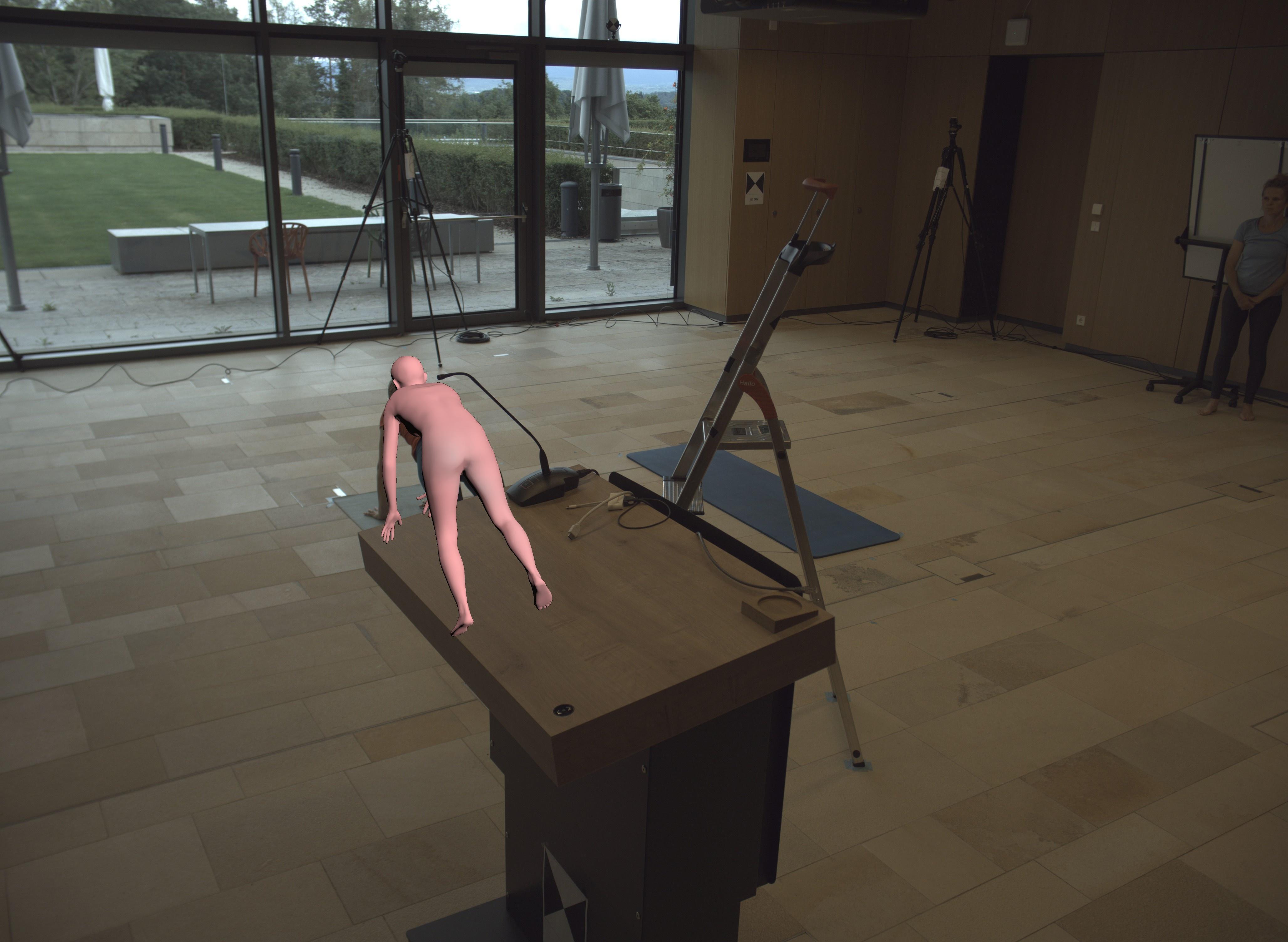}
    \includegraphics[trim=000mm 000mm 033mm 000mm, clip=true, height=0.17\linewidth]{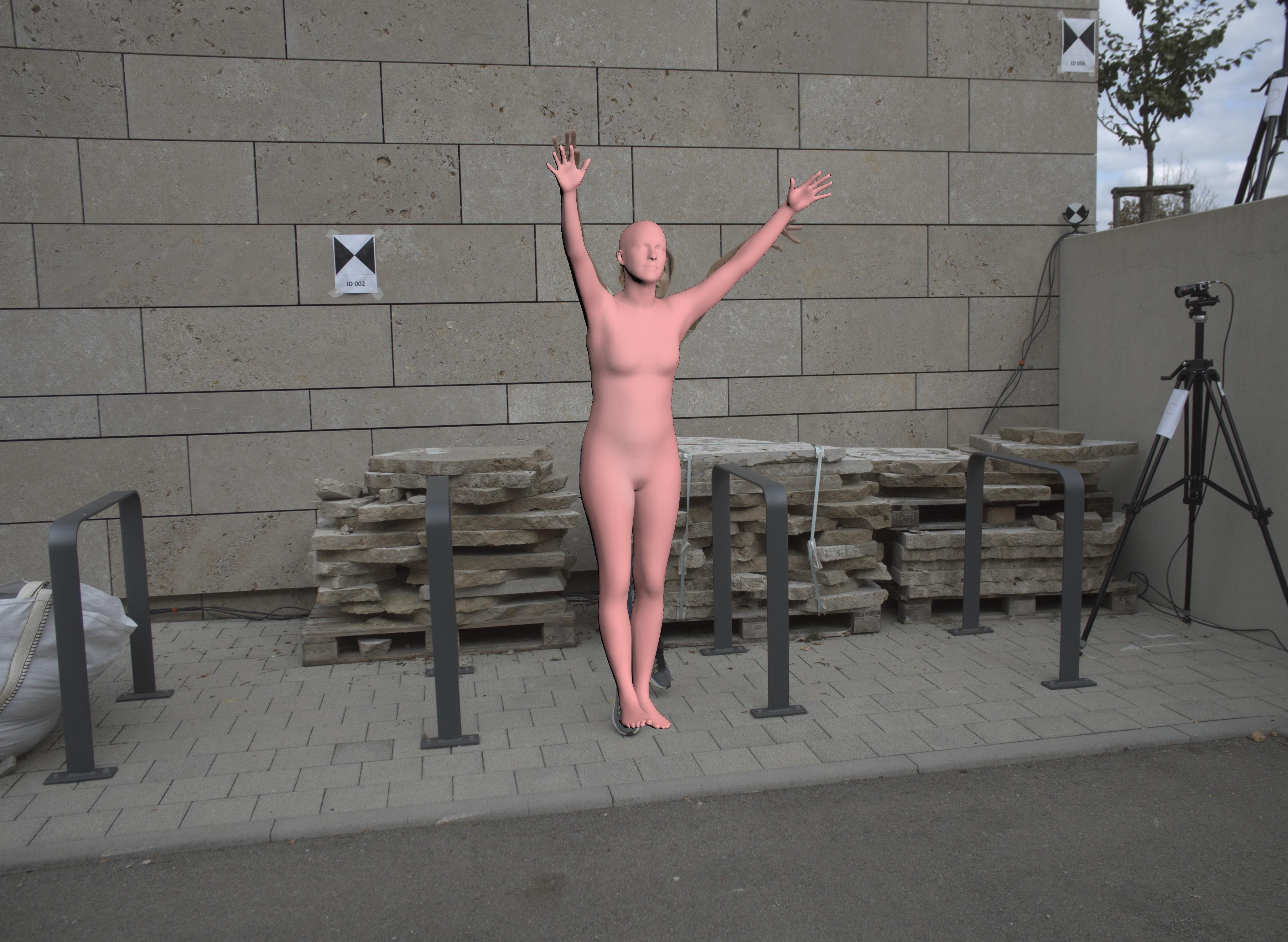}
    \includegraphics[trim=000mm 000mm 169mm 000mm, clip=true, height=0.17\linewidth]{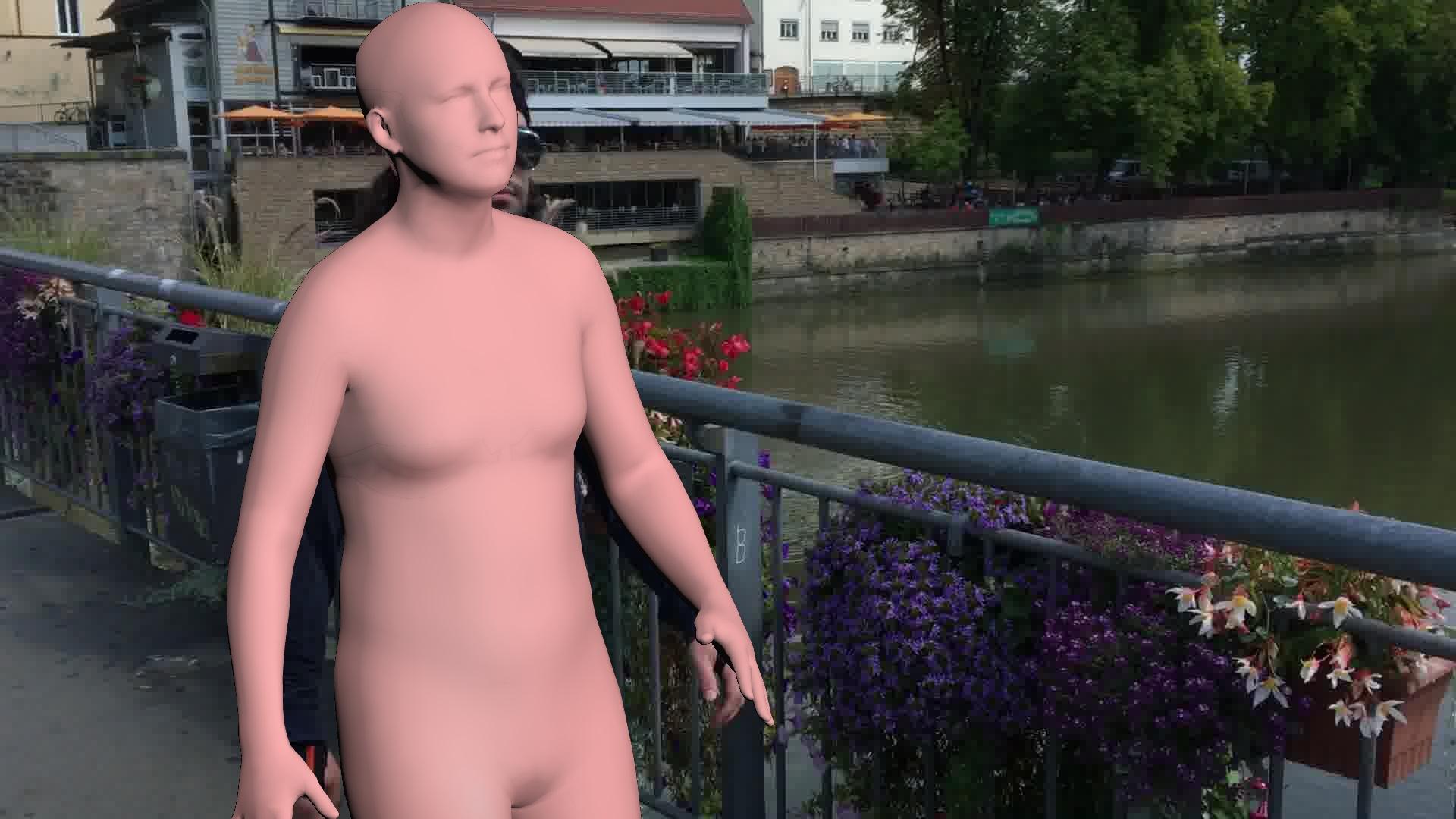}
    \includegraphics[trim=050mm 000mm 119mm 000mm, clip=true, height=0.17\linewidth]{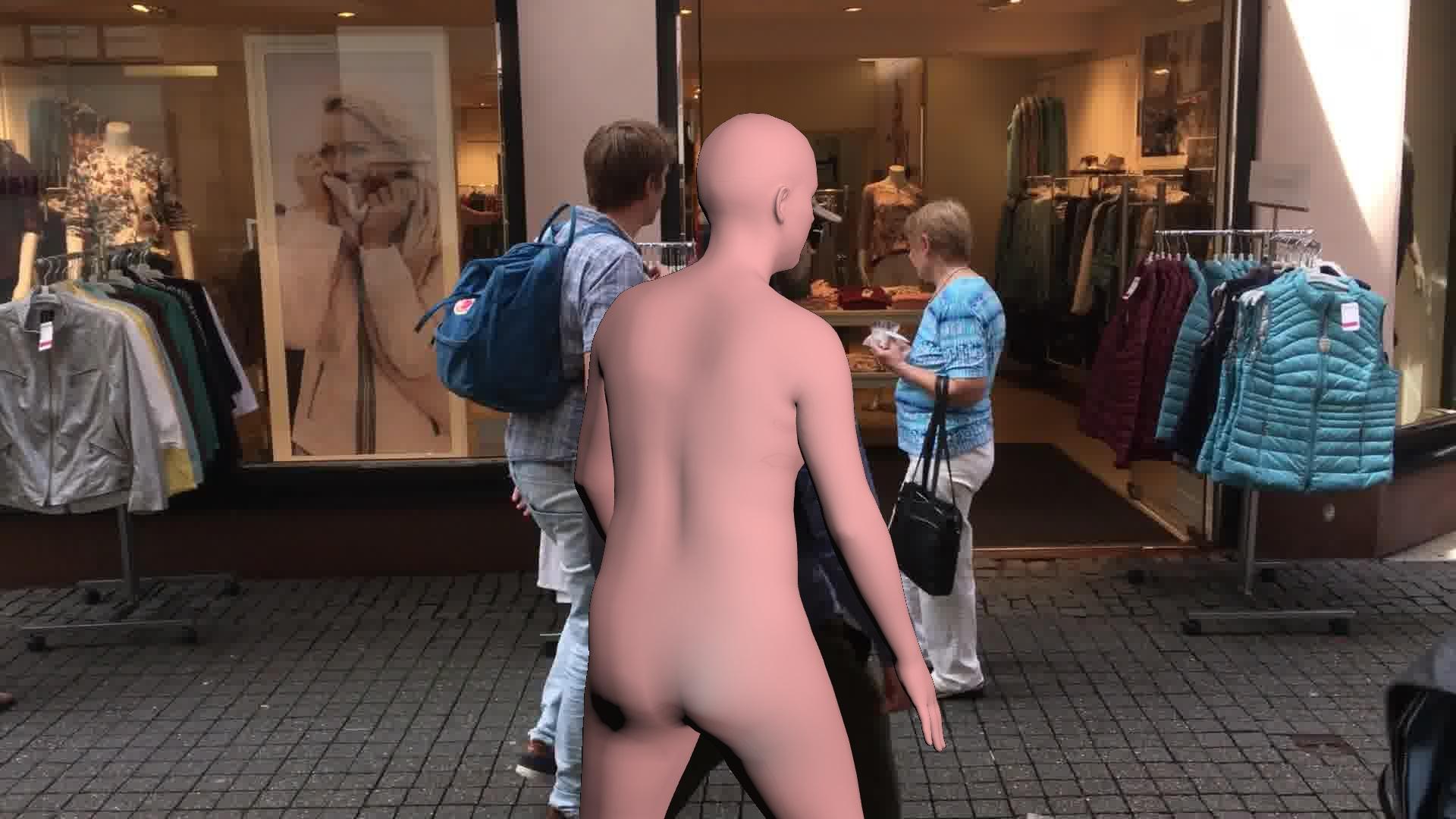}   
    \\
%SIDE VIEW
     \includegraphics[trim=000mm 000mm 000mm 000mm, clip=true, height=0.17\linewidth]{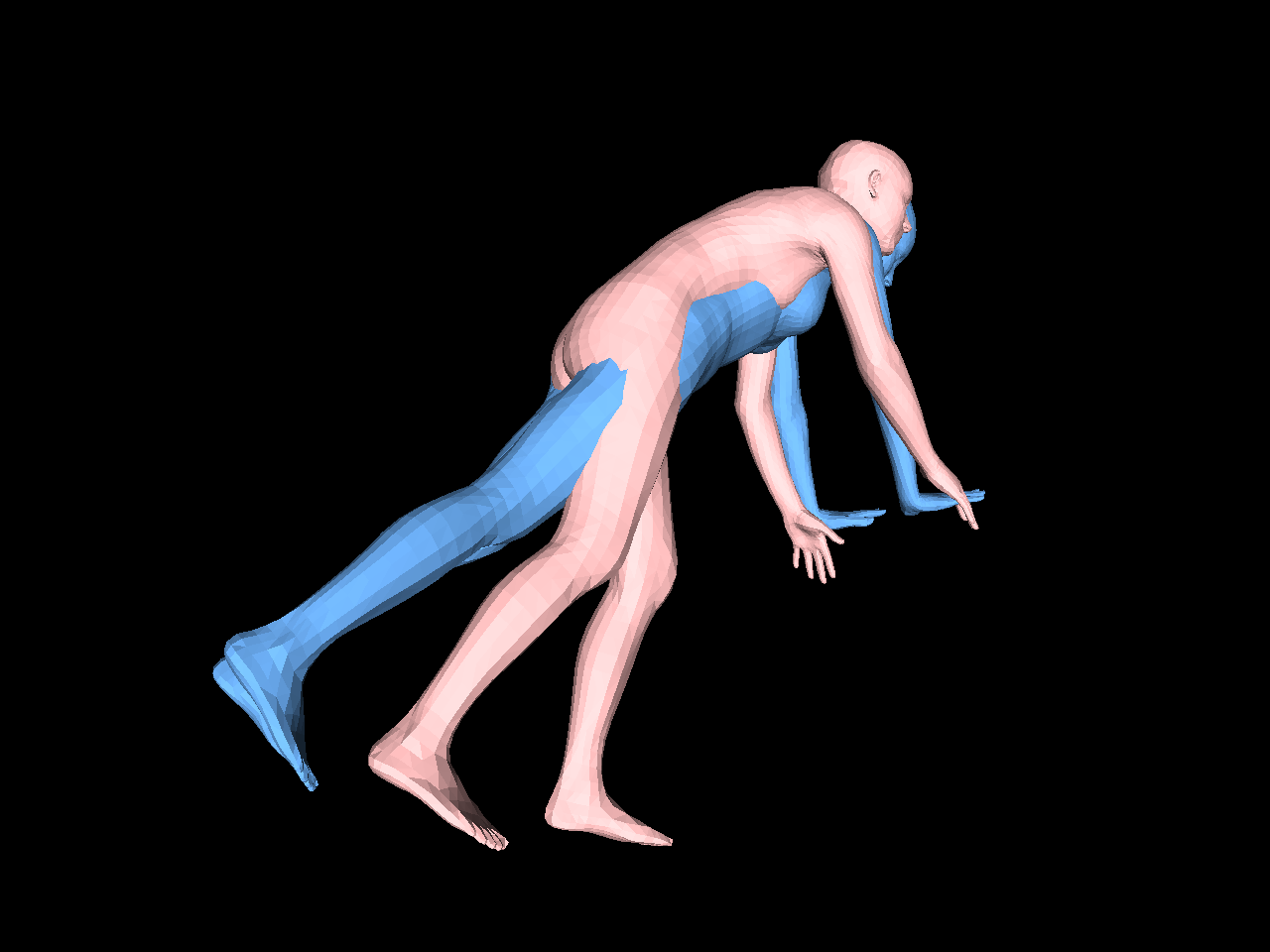}
    \includegraphics[trim=000mm 000mm 000mm 000mm, clip=true, height=0.17\linewidth]{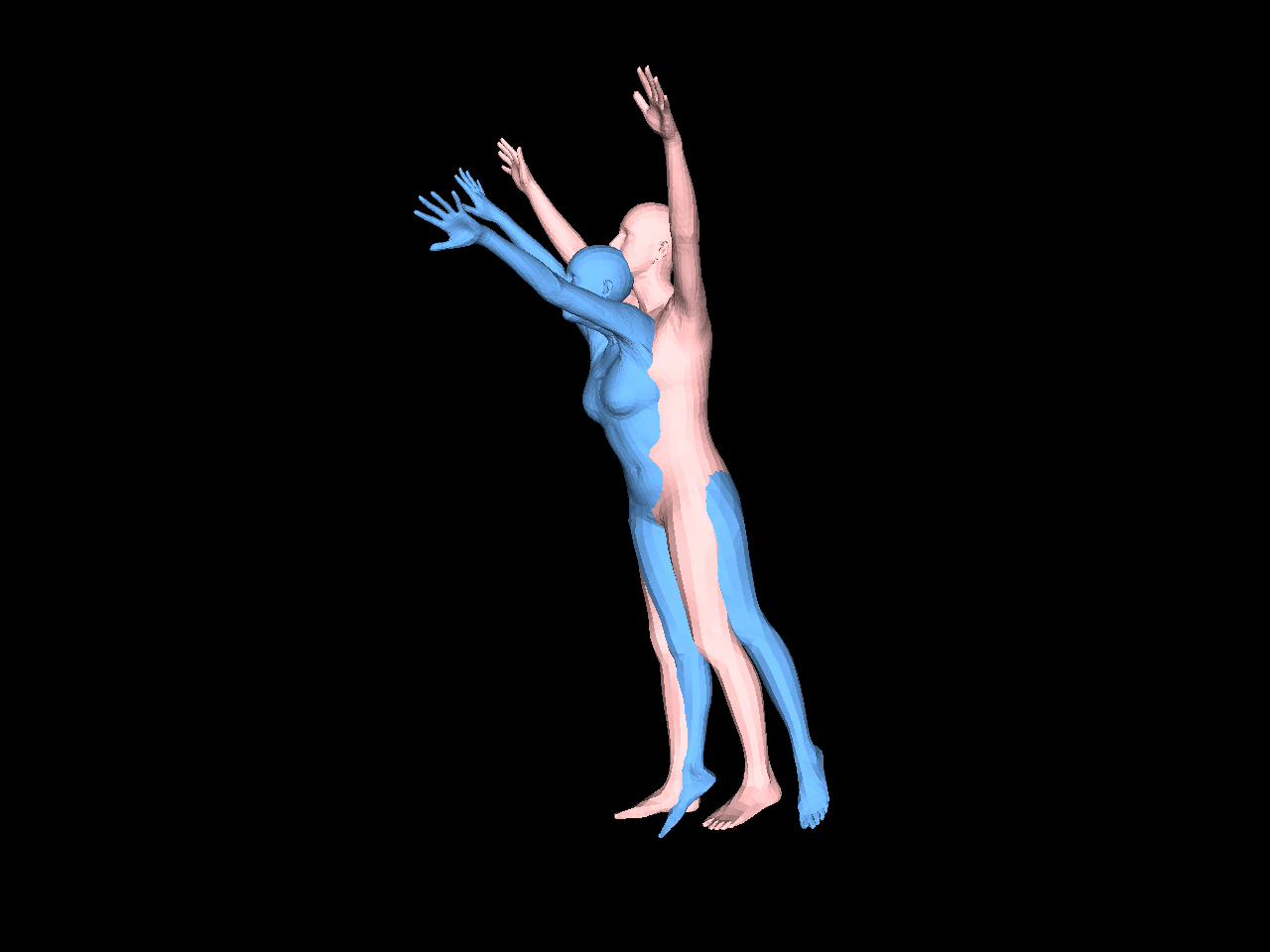}
    \includegraphics[trim=000mm 000mm 000mm 000mm, clip=true, height=0.17\linewidth]{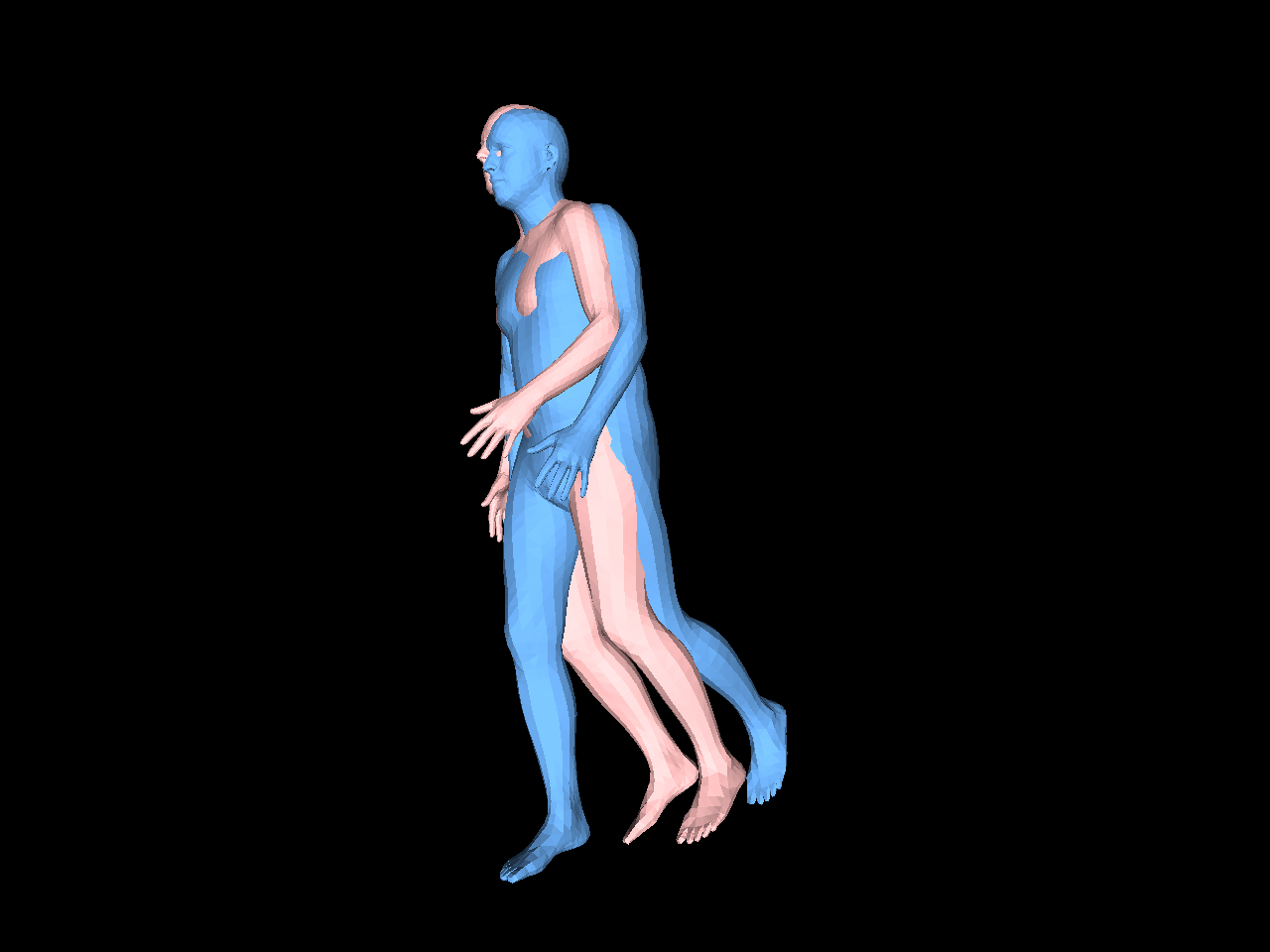}
    \includegraphics[trim=000mm 000mm 000mm 000mm, clip=true, height=0.17\linewidth]{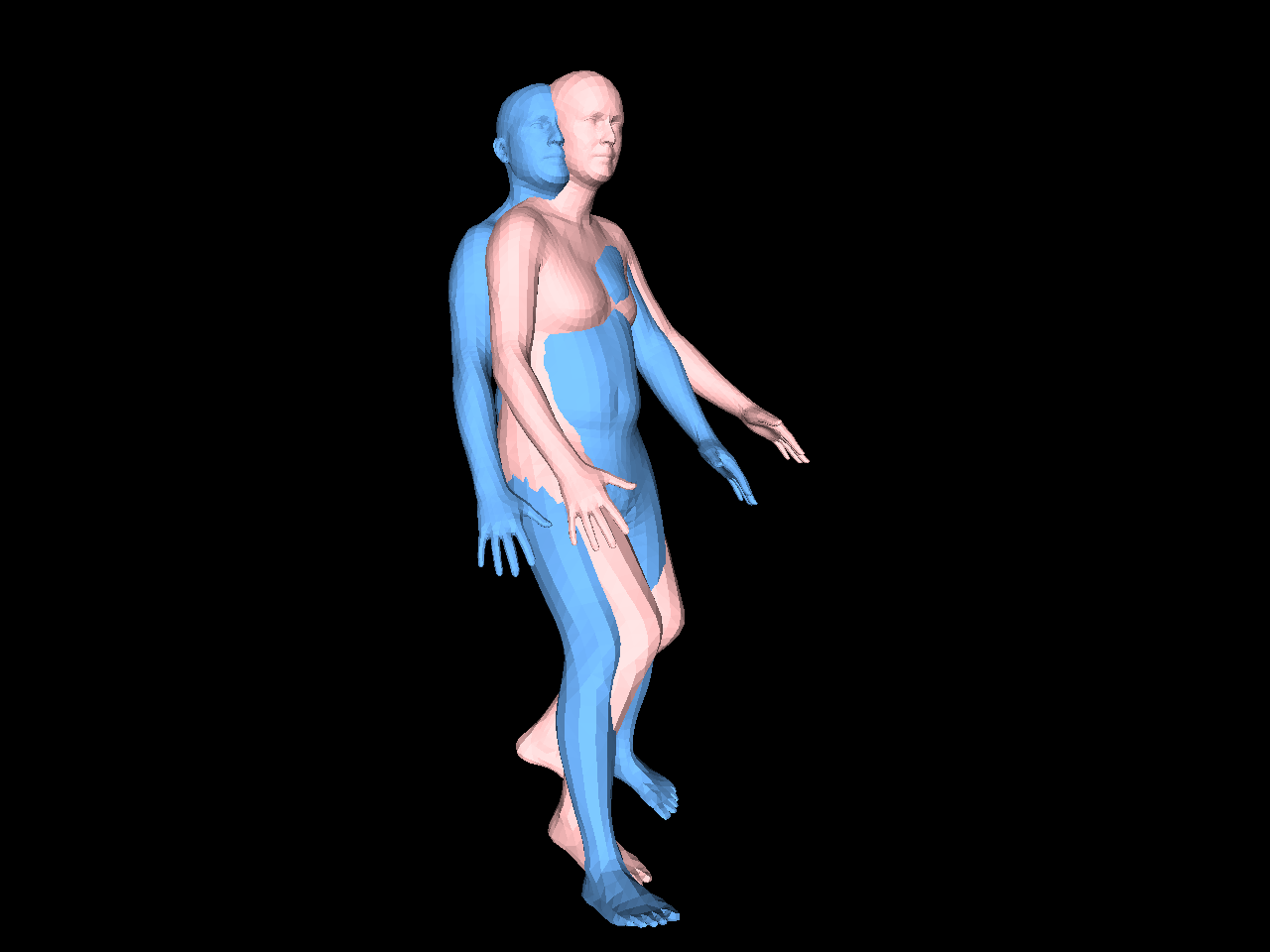}   
    \\
%ORIG VIEW
     \includegraphics[trim=000mm 000mm 033mm 000mm, clip=true, height=0.17\linewidth]{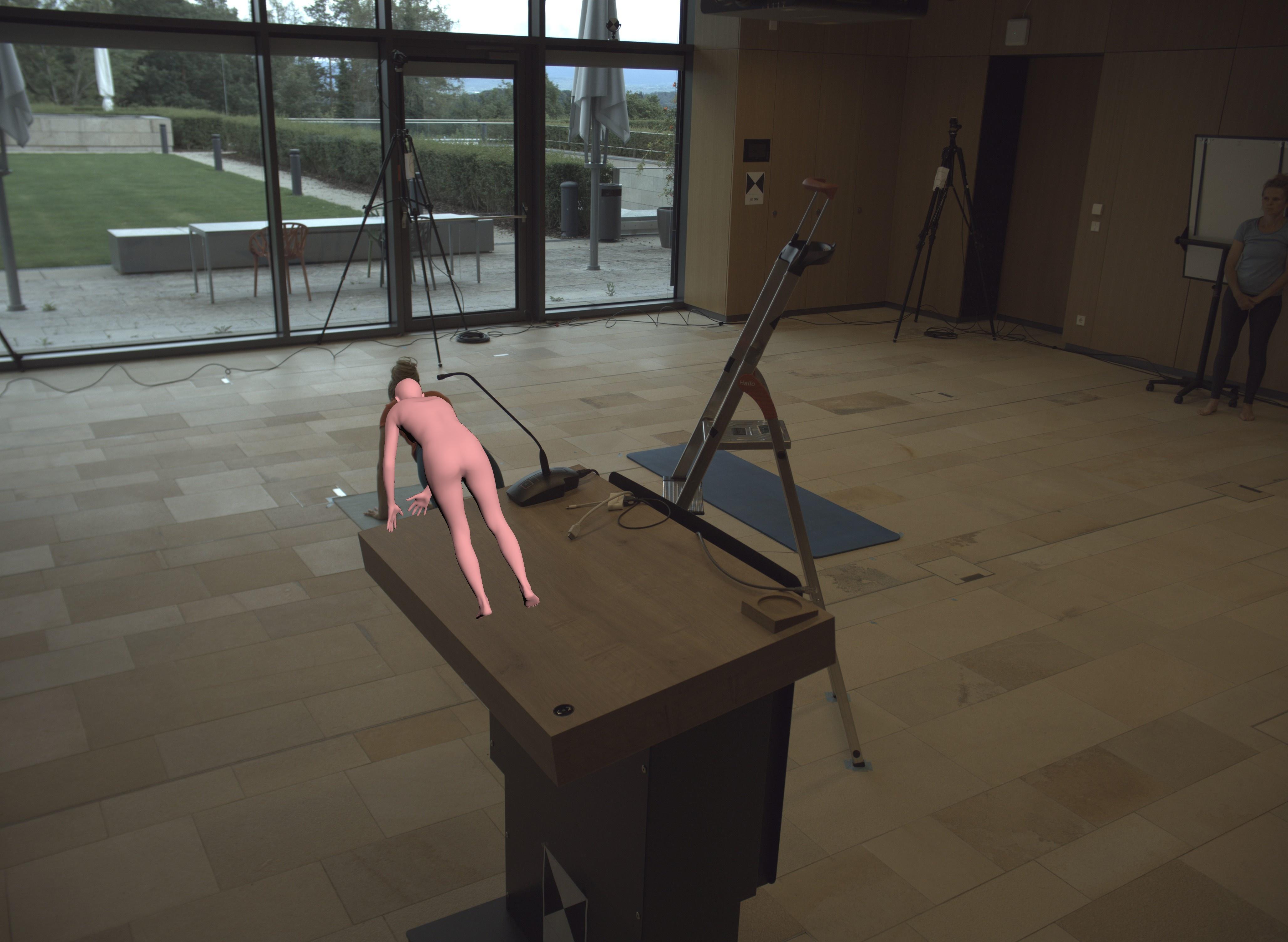}
    \includegraphics[trim=000mm 000mm 033mm 000mm, clip=true, height=0.17\linewidth]{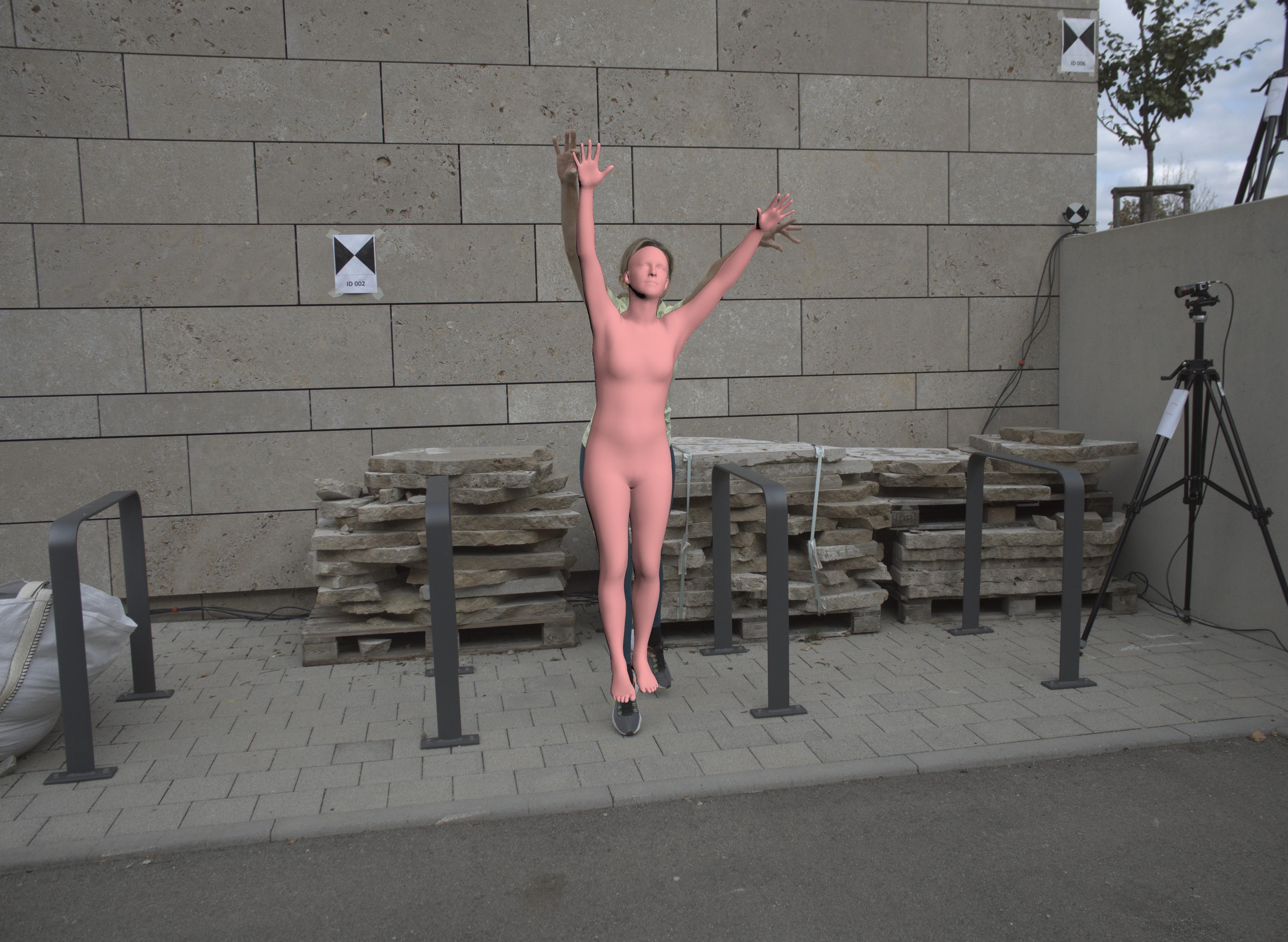}
    \includegraphics[trim=000mm 000mm 169mm 000mm, clip=true, height=0.17\linewidth]{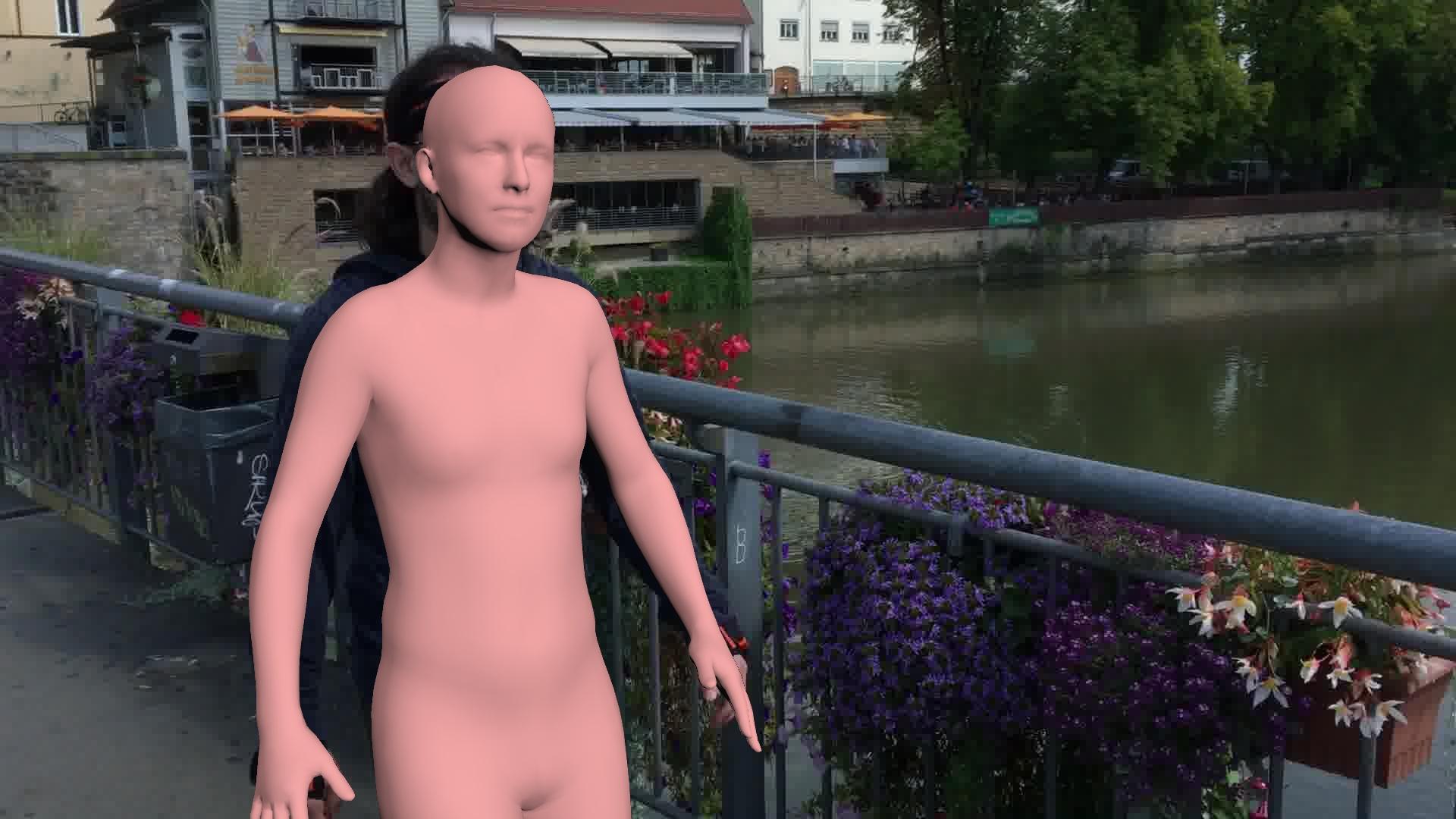}
    \includegraphics[trim=050mm 000mm 119mm 000mm, clip=true, height=0.17\linewidth]{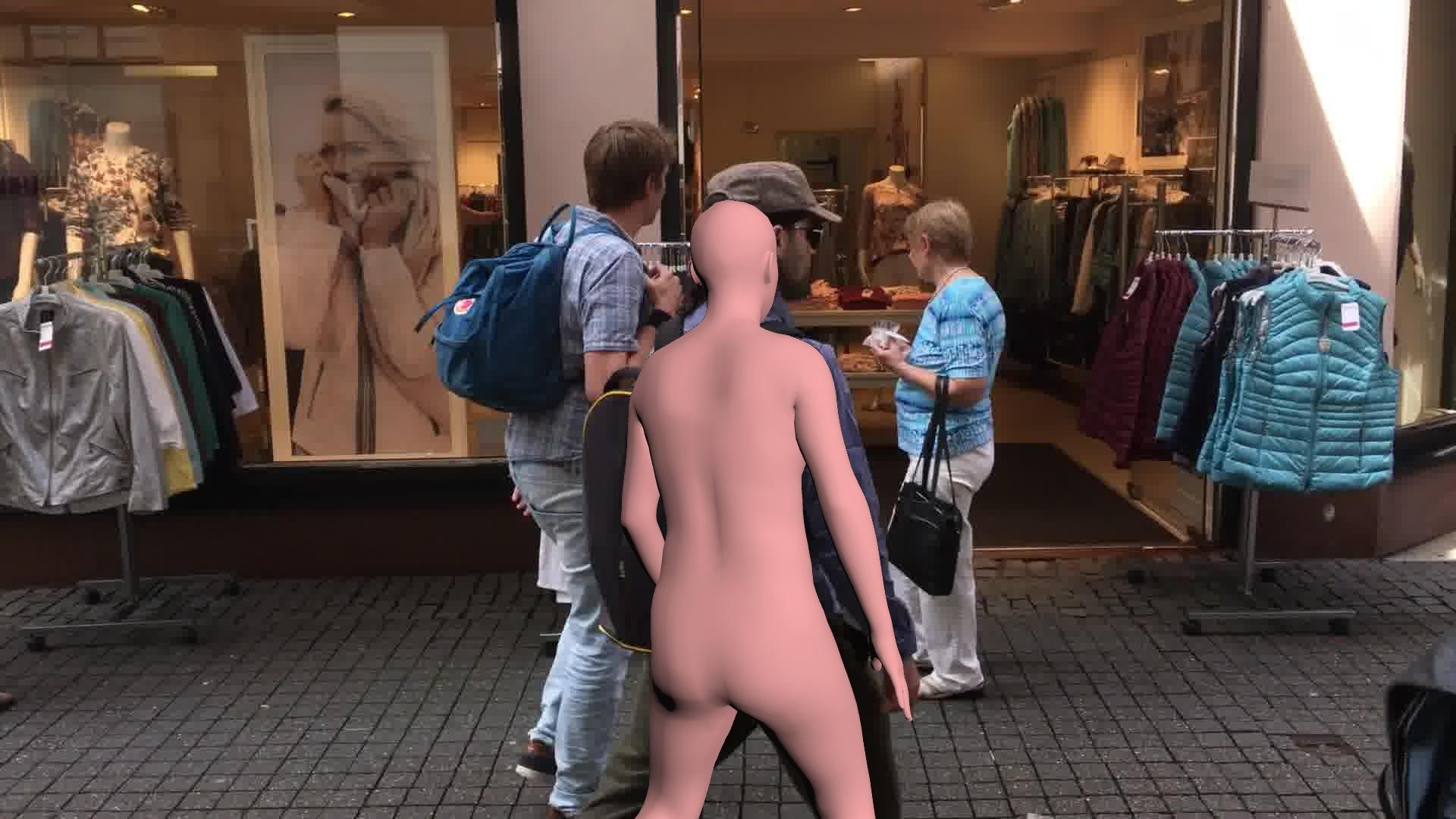}   
    \\
%SIDE VIEW

     \includegraphics[trim=000mm 000mm 000mm 000mm, clip=true, height=0.17\linewidth]{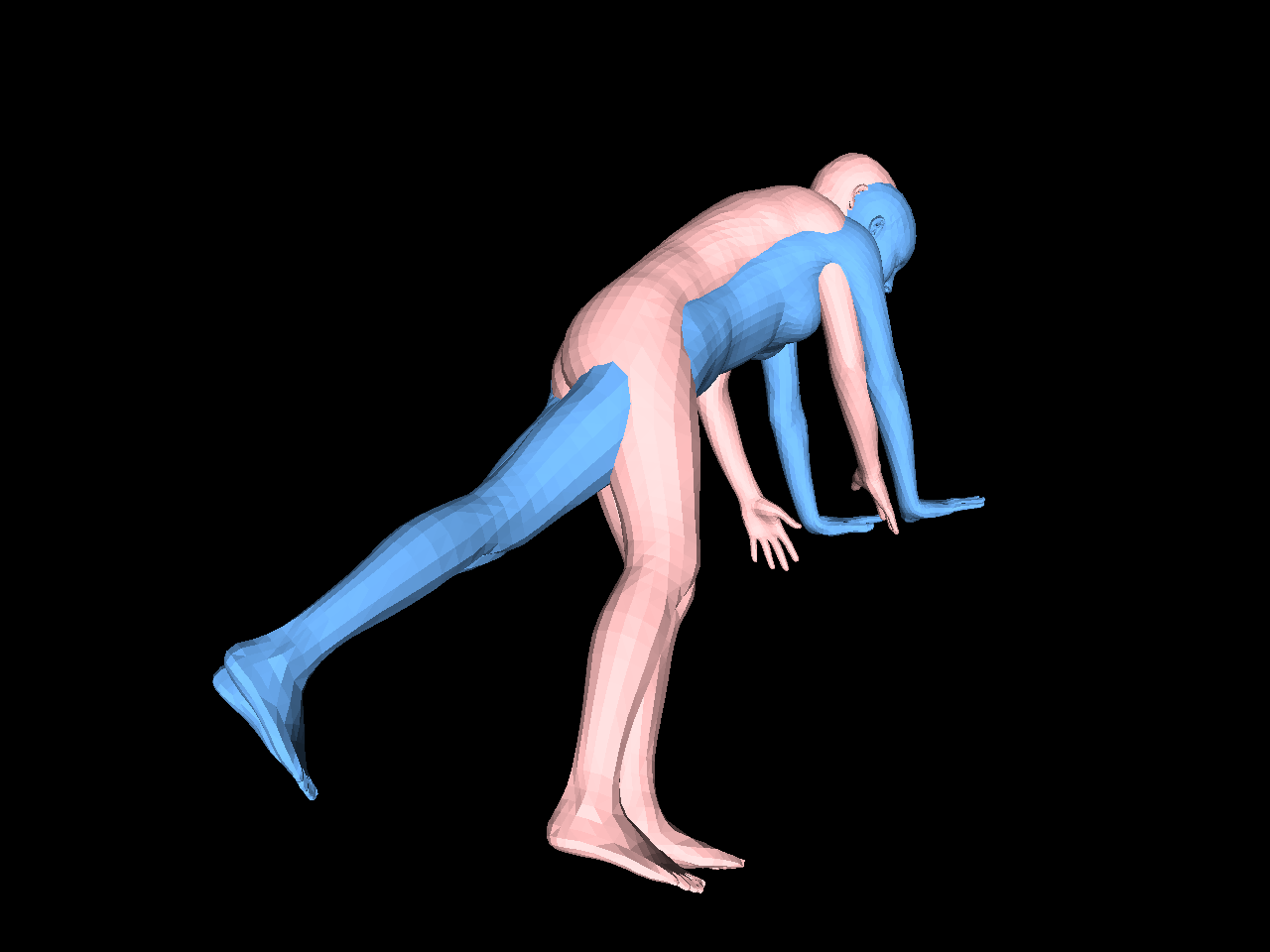}
    \includegraphics[trim=000mm 000mm 000mm 000mm, clip=true, height=0.17\linewidth]{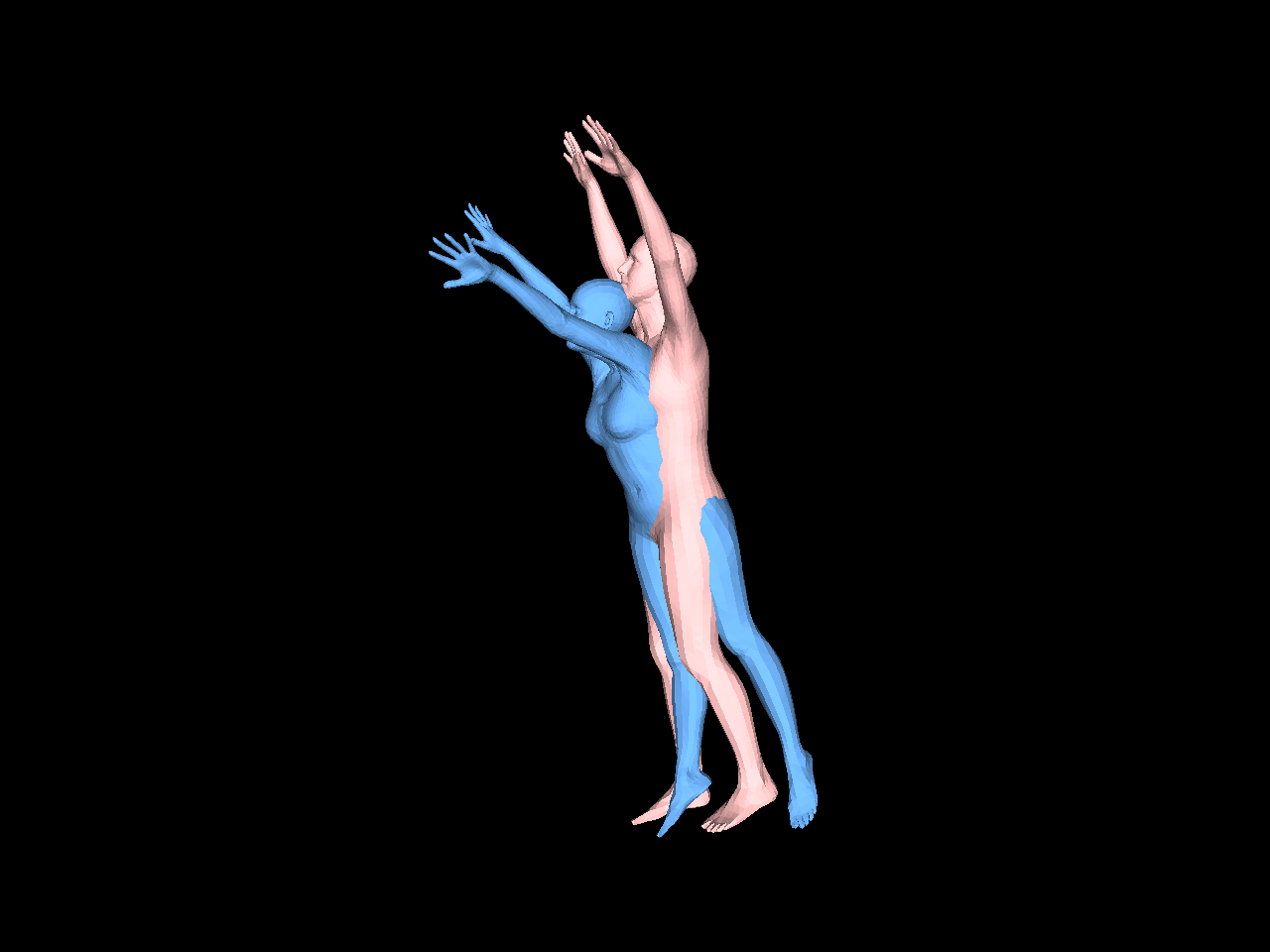}
    \includegraphics[trim=000mm 000mm 000mm 000mm, clip=true, height=0.17\linewidth]{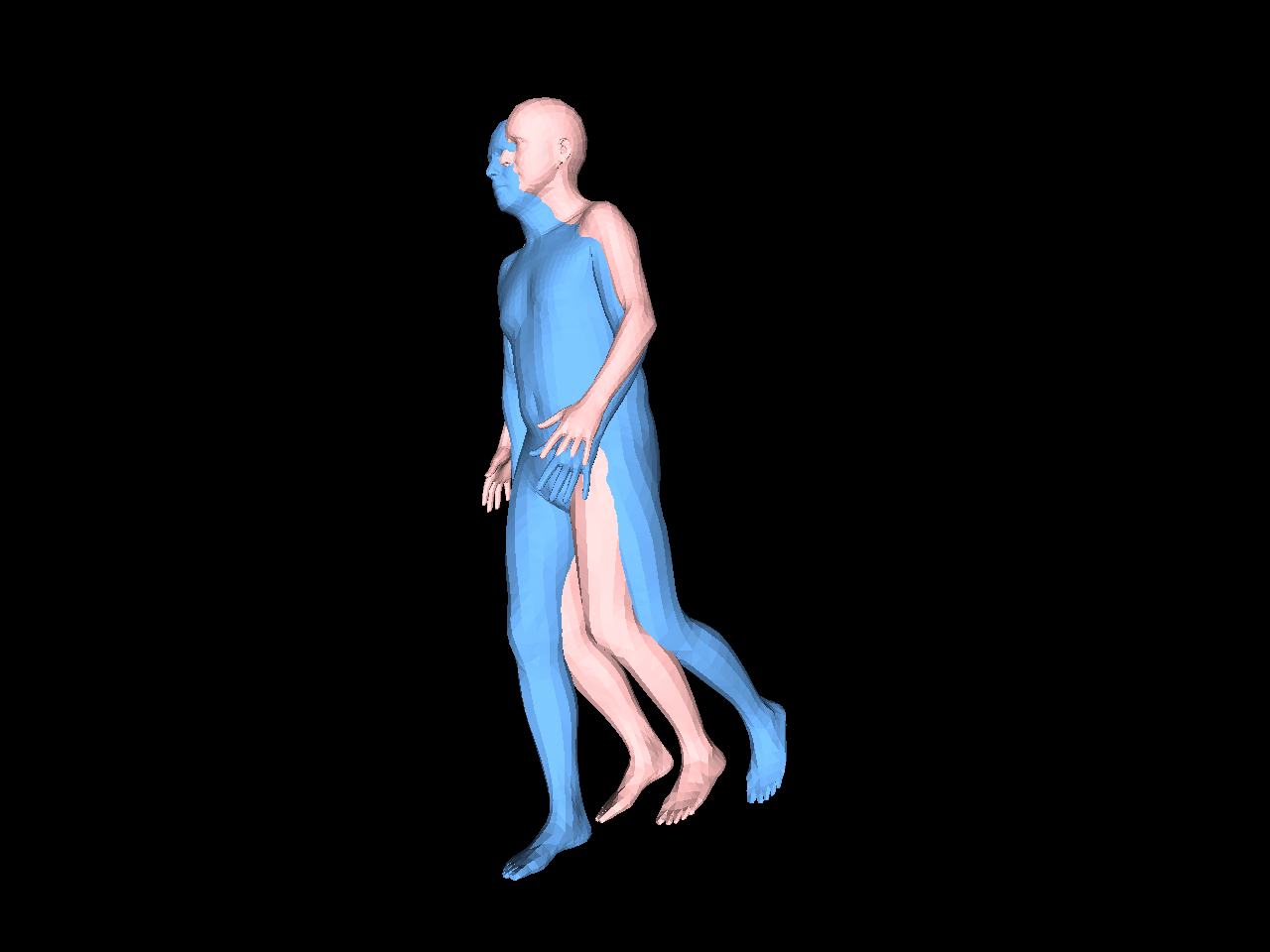}
    \includegraphics[trim=000mm 000mm 000mm 000mm, clip=true, height=0.17\linewidth]{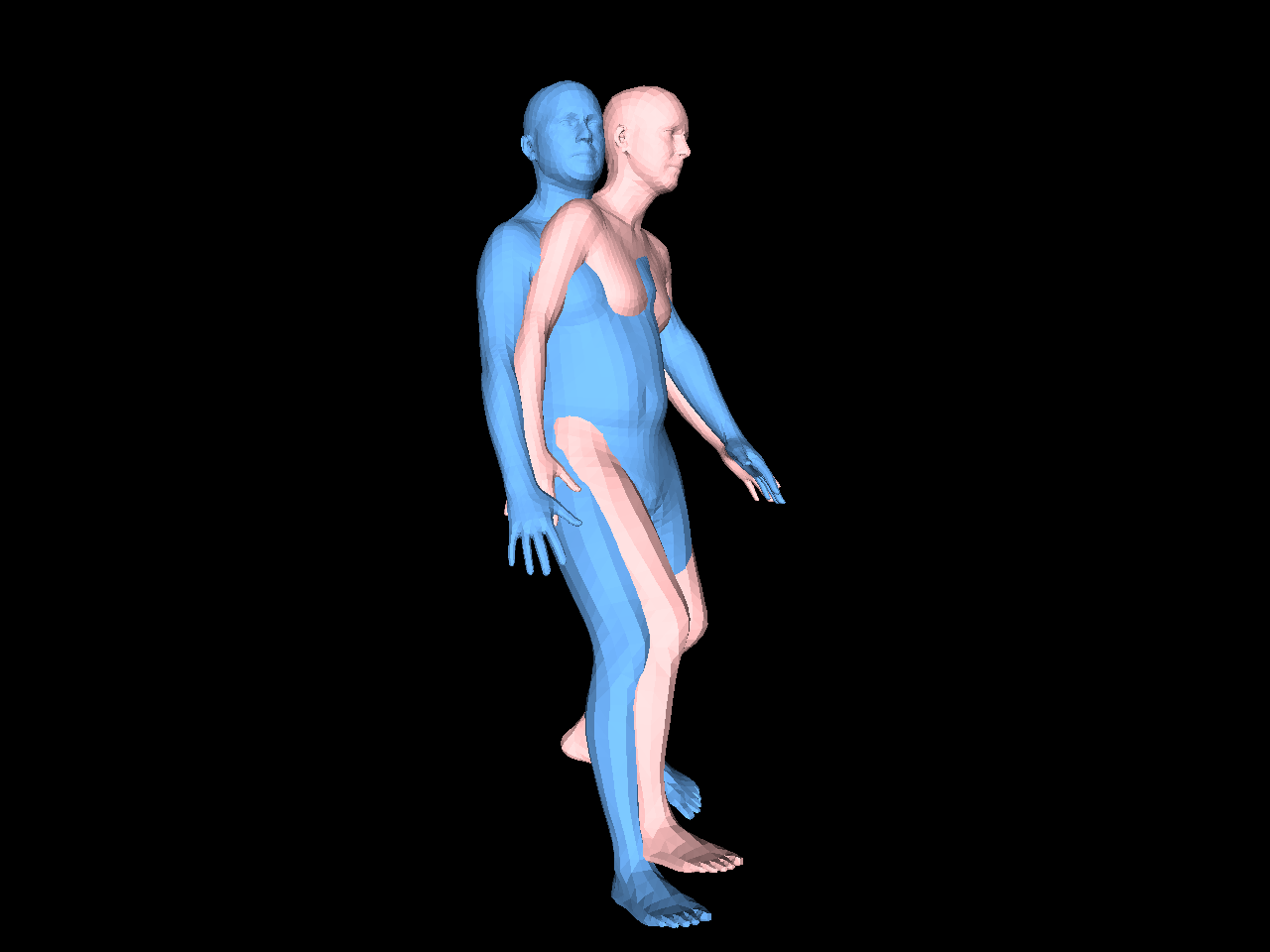}
    \\
    
     \includegraphics[trim=000mm 000mm 033mm 000mm, clip=true, height=0.17\linewidth]{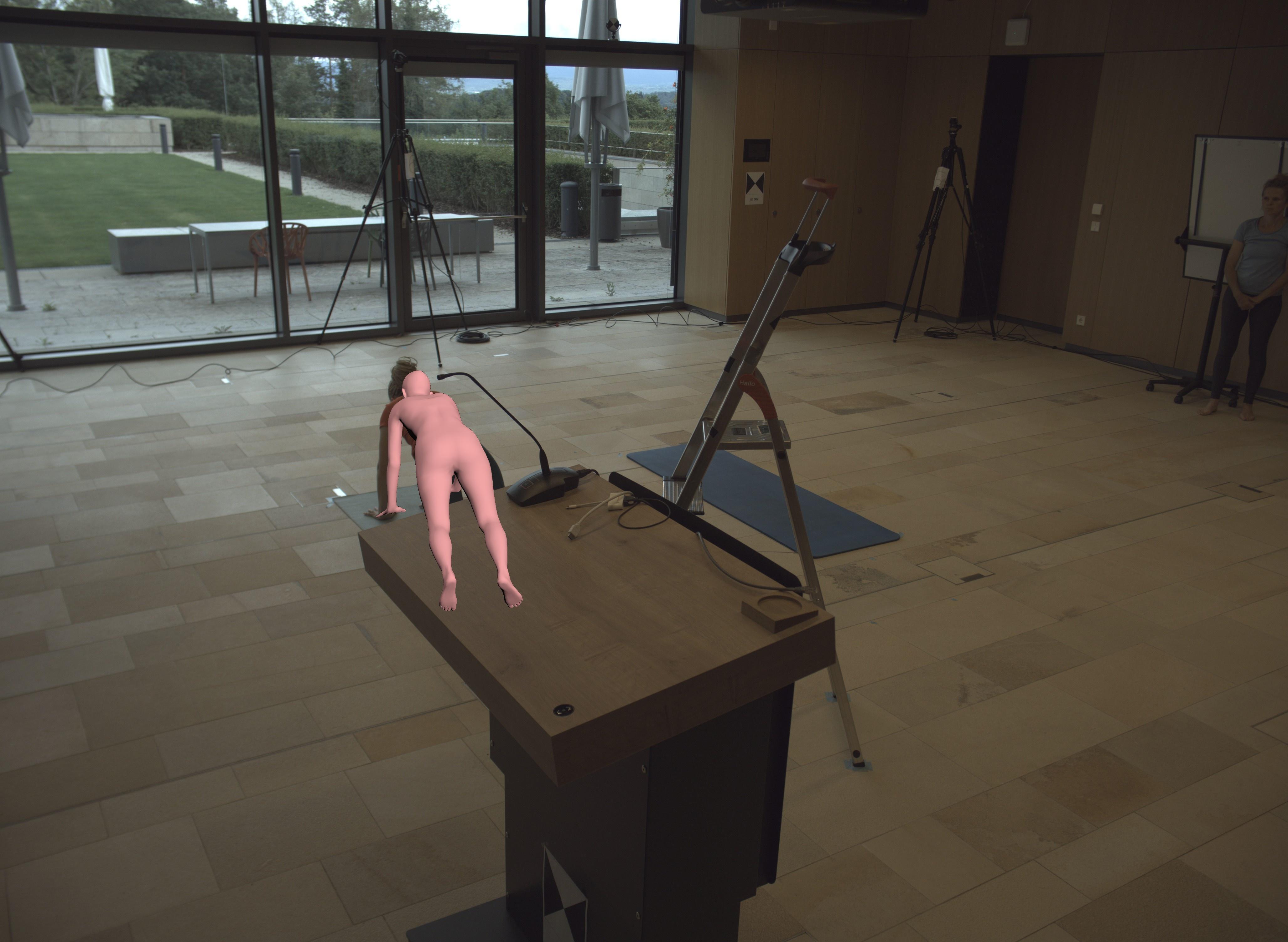}
    \includegraphics[trim=000mm 000mm 033mm 000mm, clip=true, height=0.17\linewidth]{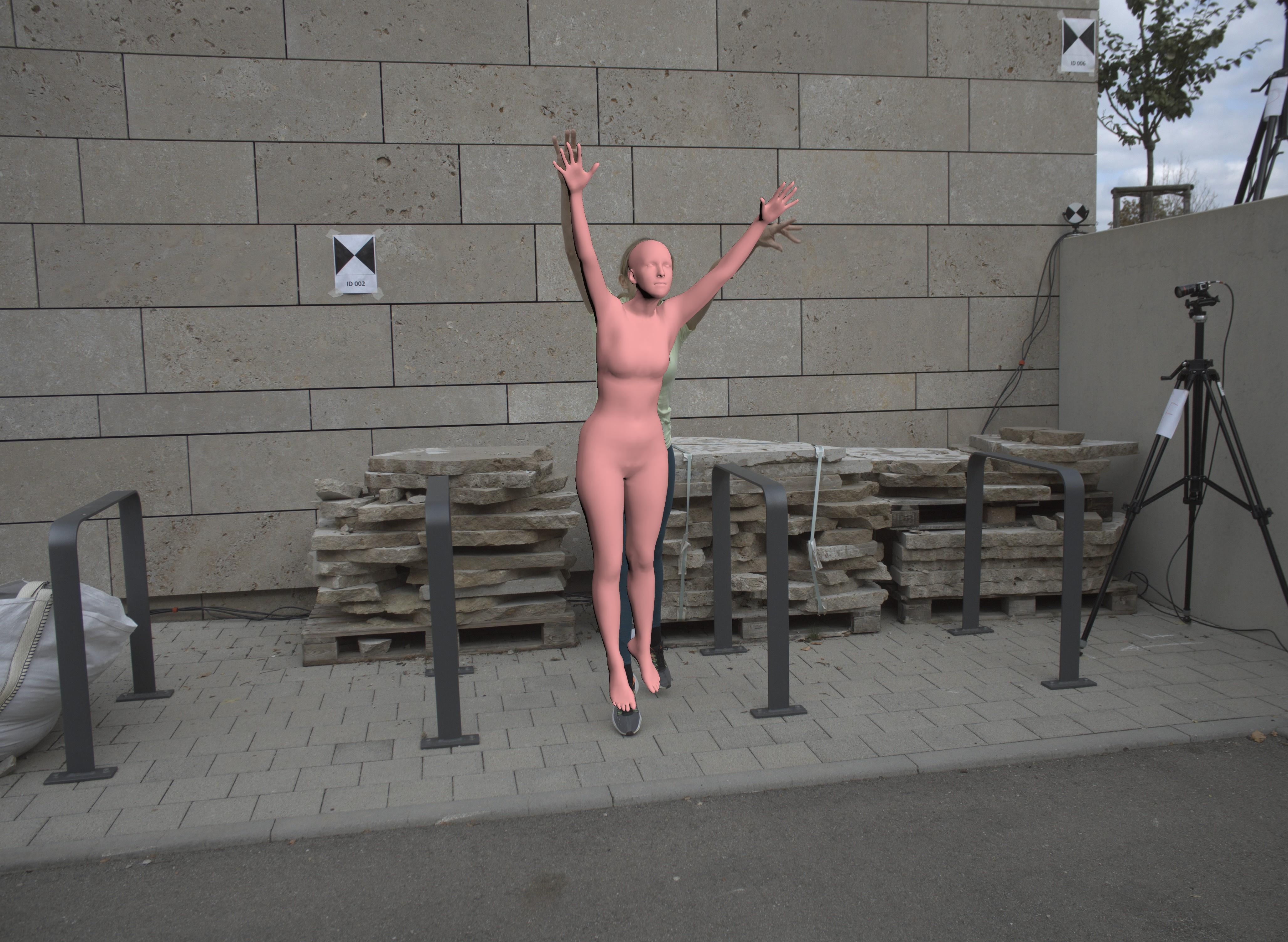}
    \includegraphics[trim=000mm 000mm 169mm 000mm, clip=true, height=0.17\linewidth]{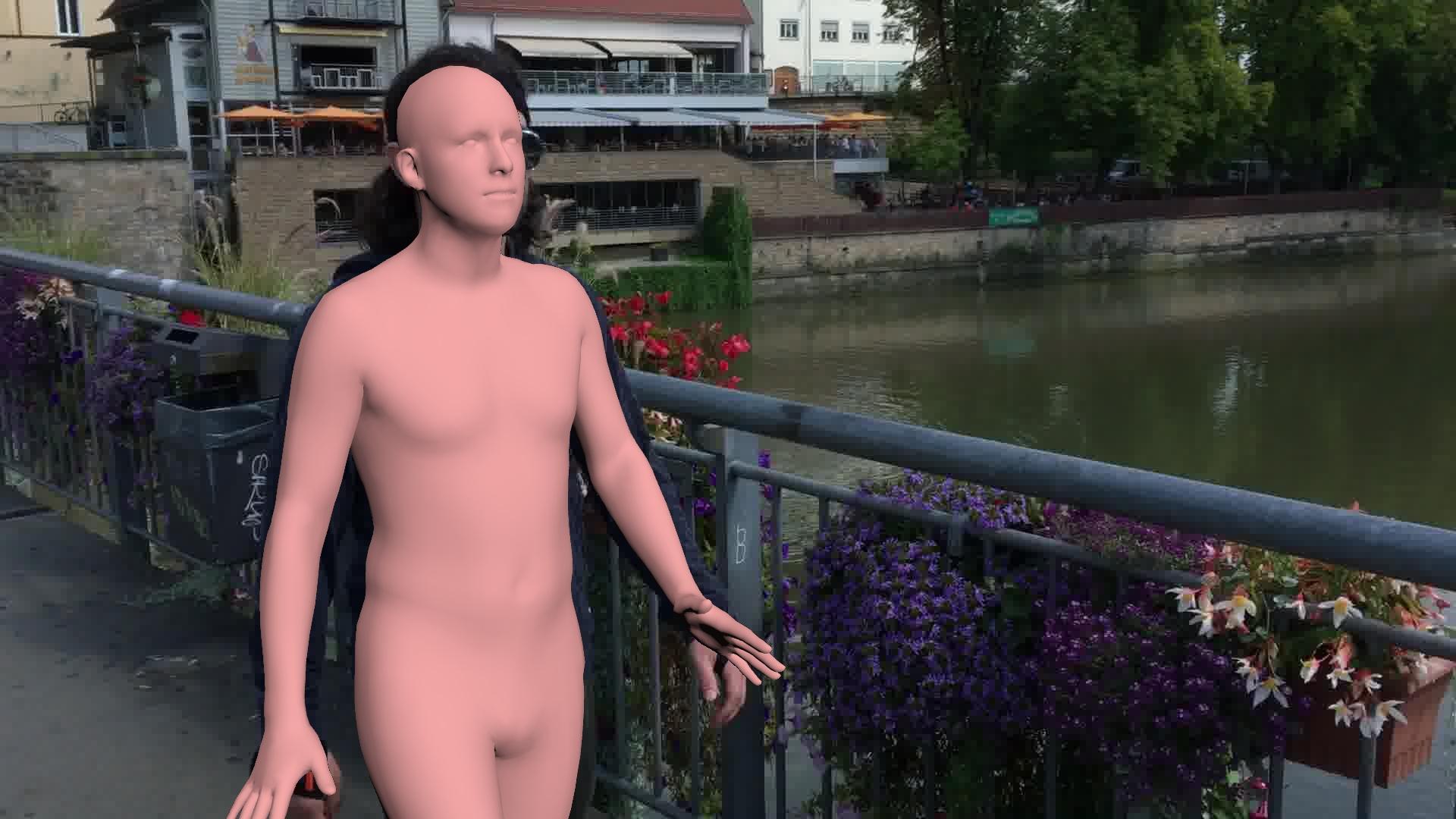}
    \includegraphics[trim=050mm 000mm 119mm 000mm, clip=true, height=0.17\linewidth]{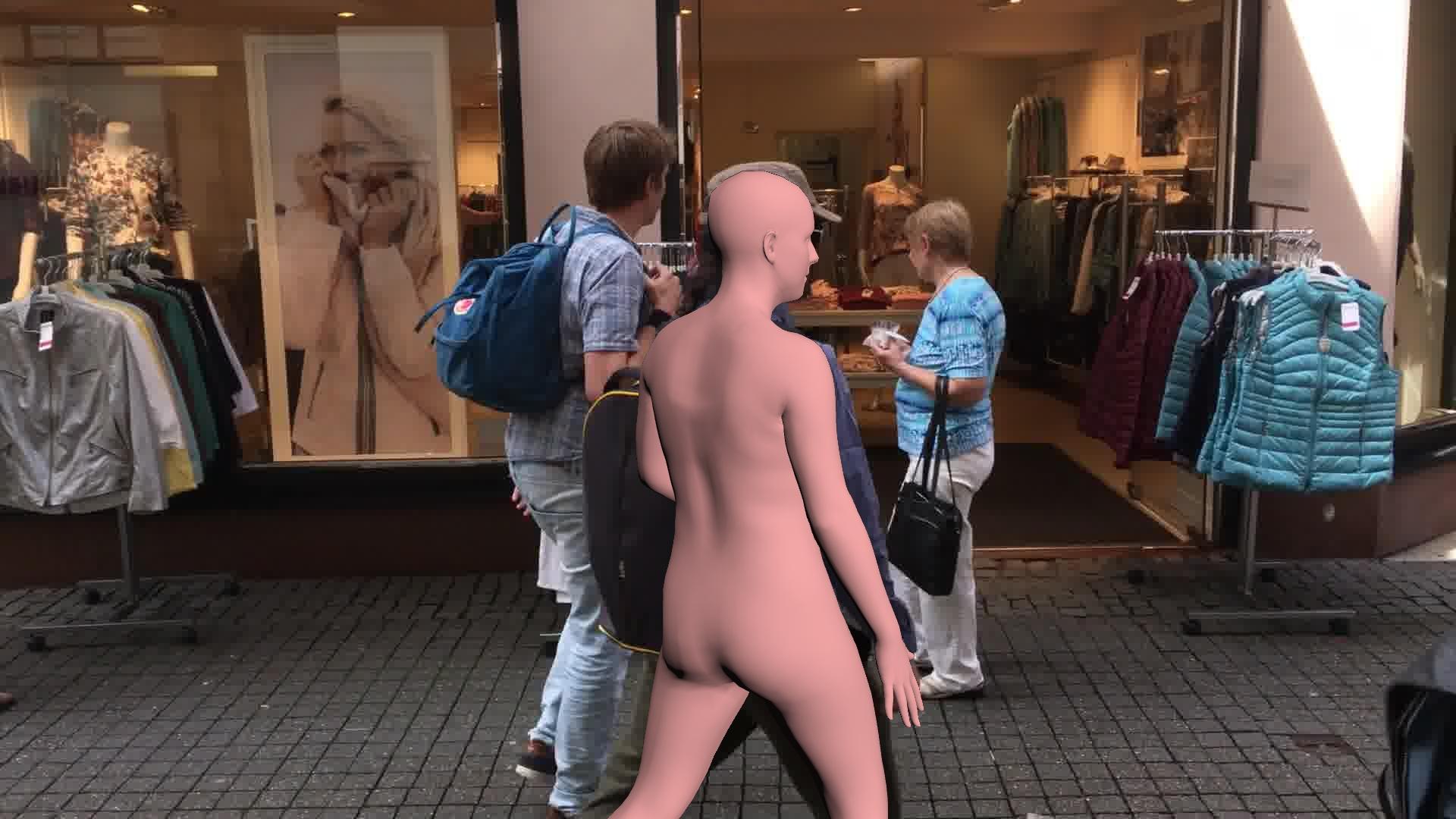}   
    \\
    %SIDE VIEW
    
     \includegraphics[trim=000mm 000mm 000mm 000mm, clip=true, height=0.17\linewidth]{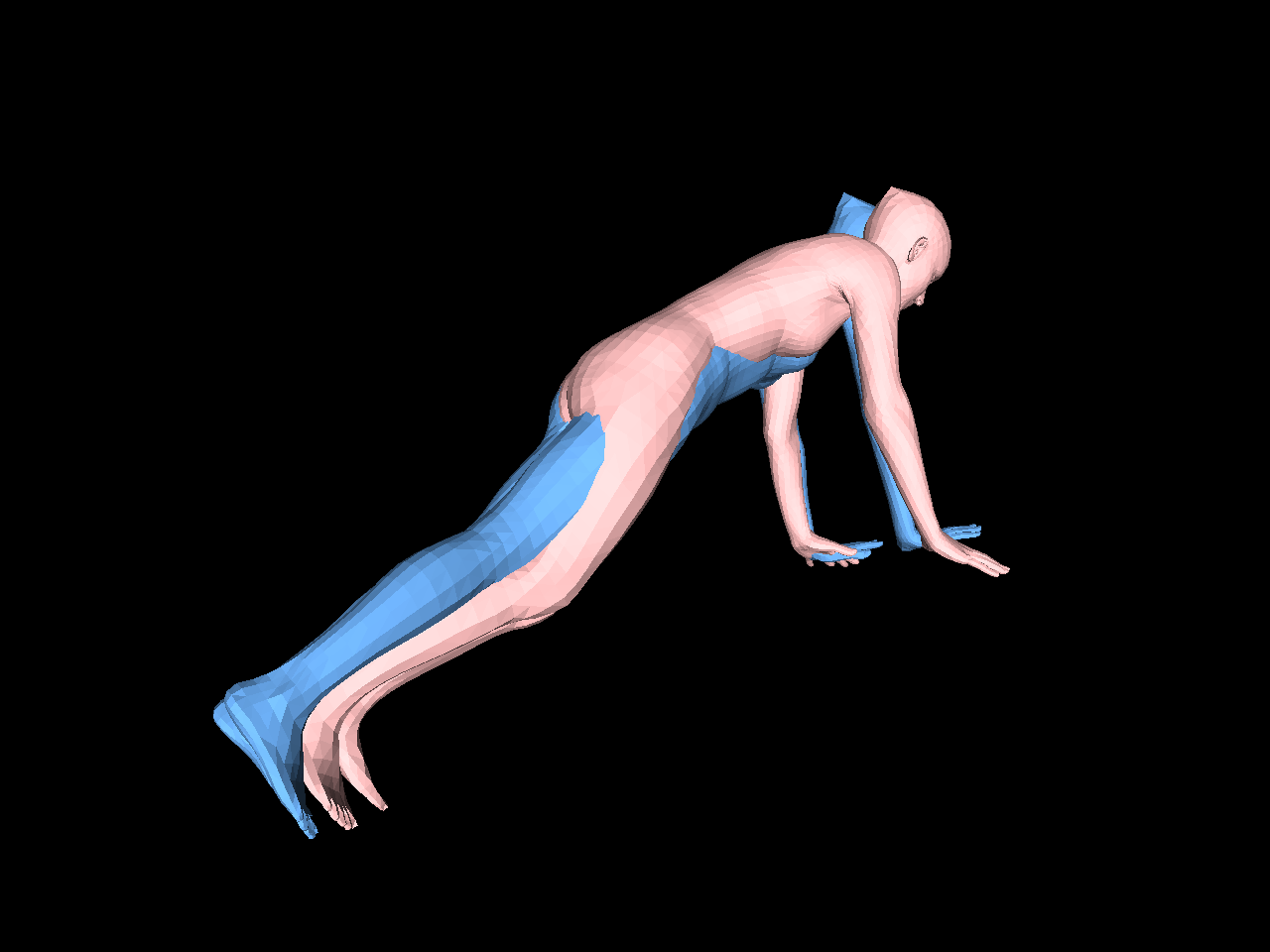}
    \includegraphics[trim=000mm 000mm 000mm 000mm, clip=true, height=0.17\linewidth]{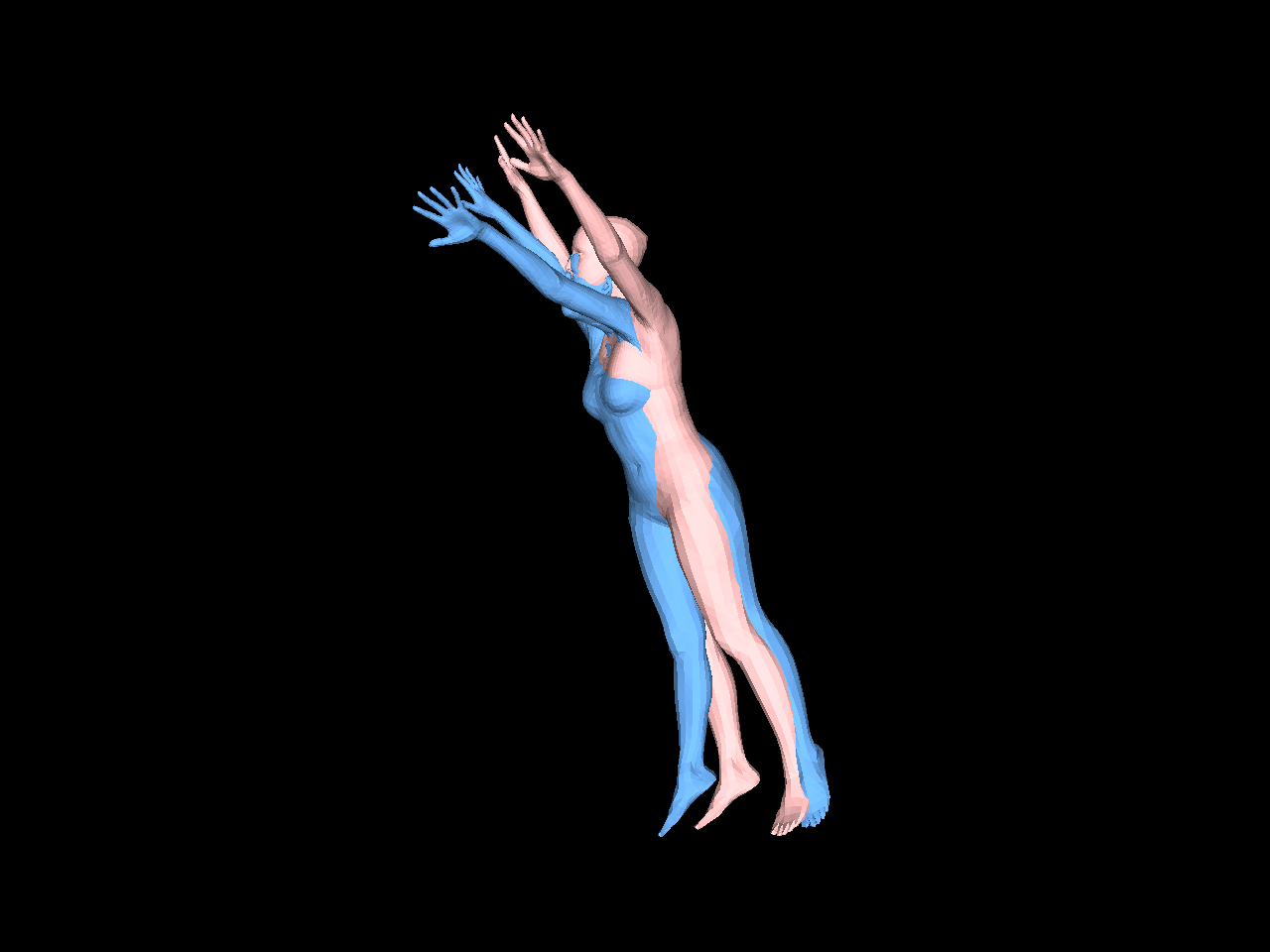}
    \includegraphics[trim=000mm 000mm 000mm 000mm, clip=true, height=0.17\linewidth]{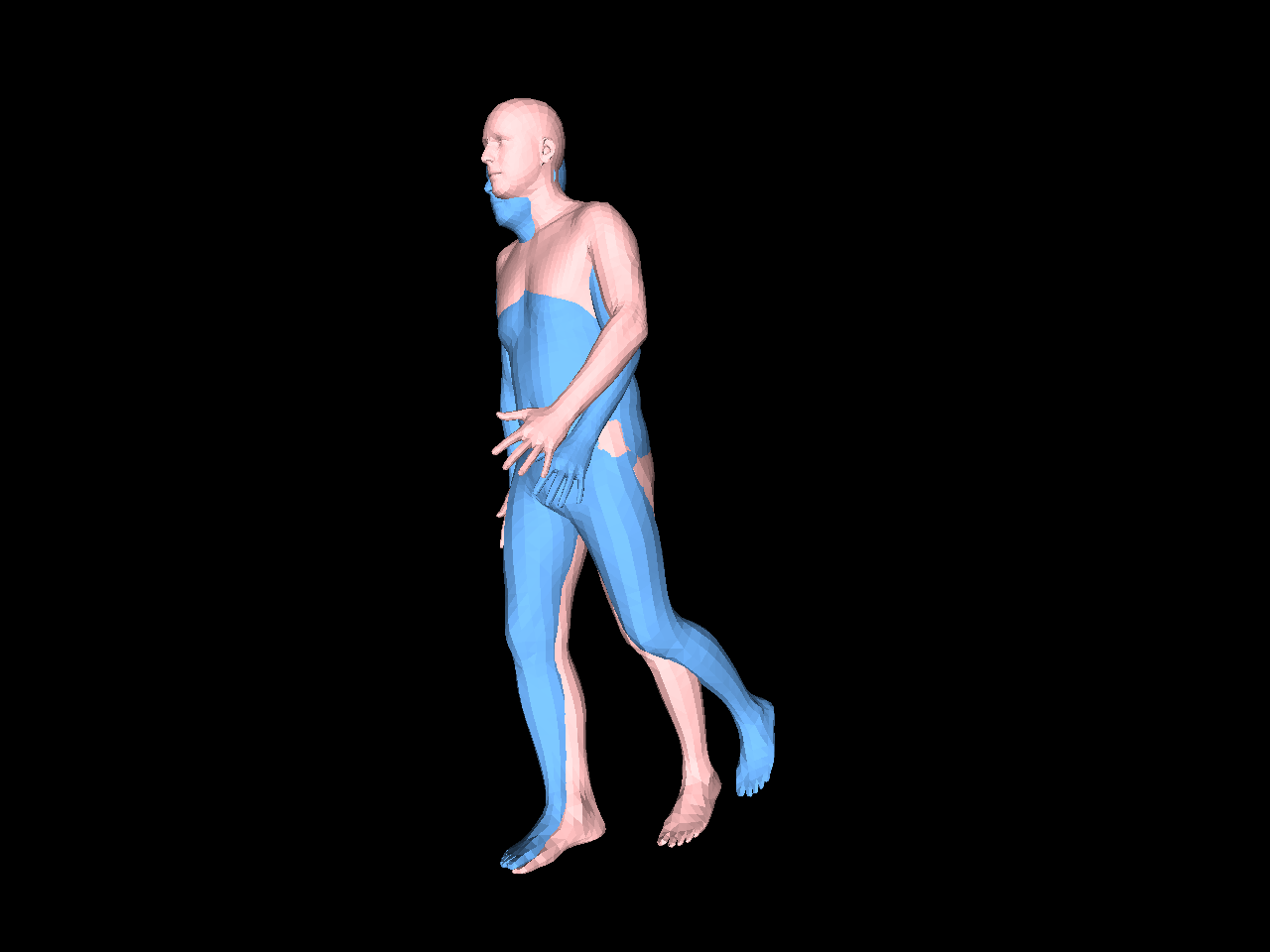}
    \includegraphics[trim=000mm 000mm 000mm 000mm, clip=true, height=0.17\linewidth]{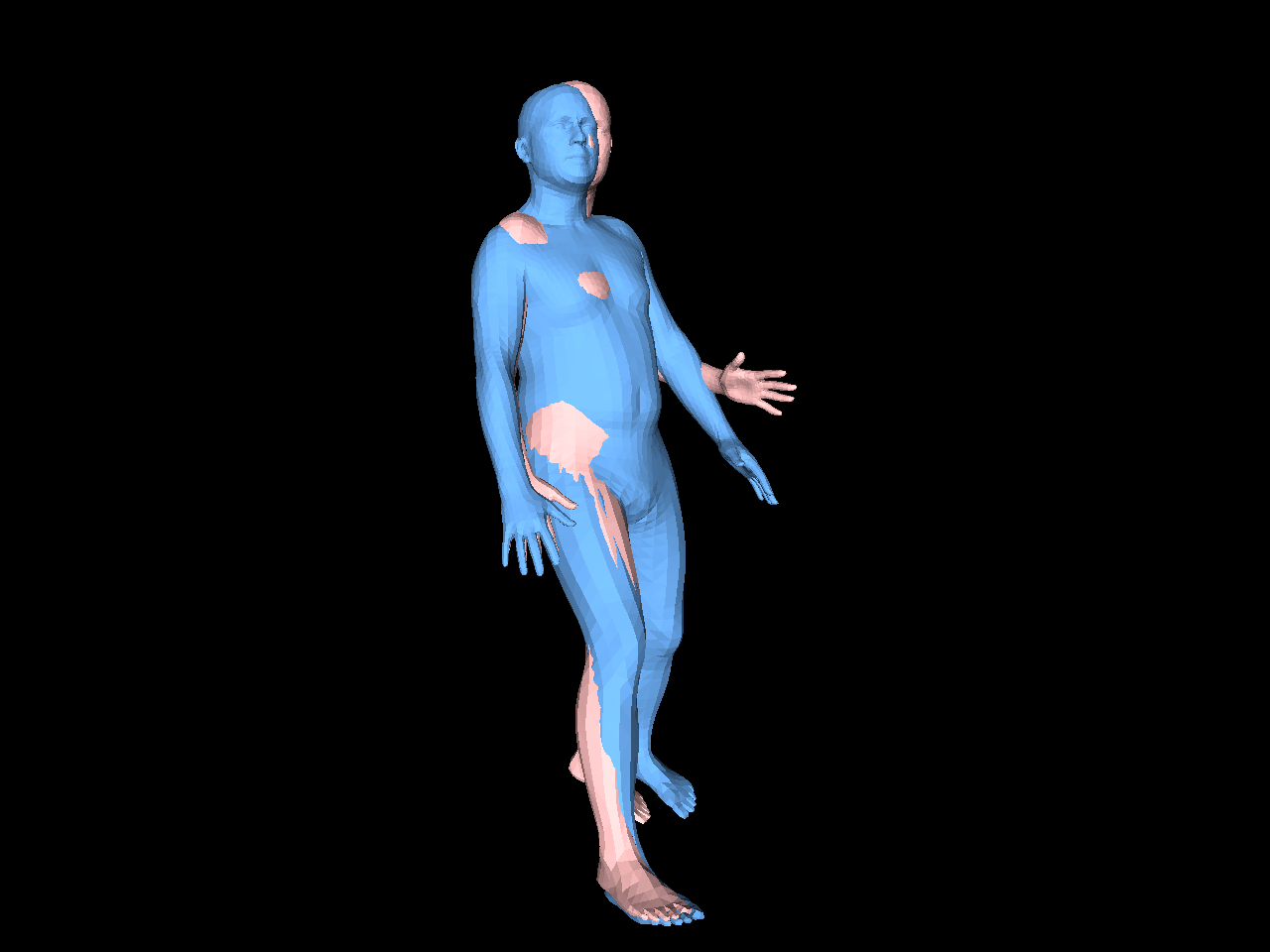}  
    \caption{Qualitative results on RICH (left two columns) and 3DPW (right two columns). RGB images (row 1), PARE front (row 2), PARE side (row 3), CLIFF front (row 4), CLIFF side (row 5), BEDLAM-CLIFF front (row 6), BEDLAM-CLIFF side (row 7). Ground truth body is in blue and predicted body is in pink. The BEDLAM-CLIFF predicted 3D body is better aligned with ground truth in both front and side views despite wide camera variation or frame occlusion. }
    \label{fig:qualitative-comparison}
\end{figure*}

Figure~\ref{fig:qualitative-comparison} provides a qualitative comparison between PARE\cite{pare}, CLIFF\cite{li2022cliff} (includes 3DPW training) and BEDLAM-CLIFF (only synthetic data). 
We show results on both RICH (left two) and 3DPW (right two).
We render predicted bodies overlaid on the image and in a side view.
In the side view, the pelvis of the predicted body is aligned (translation only) with the ground truth body. 
Note that, when projected into the image, all methods look reasonable and relatively well aligned with the image features.
The side view, however, reveals that BEDLAM-CLIFF (bottom row) predicts a better aligned body pose with the ground truth body in 3D despite variation in the cameras, camera angle, and frame occlusion. 
Also, please notice that BEDLAM-CLIFF produces more natural leg poses in the case of  occlusion compared to the other methods as shown in columns 1, 3 and 4 of \cref{fig:qualitative-comparison} 

\begin{figure*}
\centering

\includegraphics[width=0.45\linewidth]{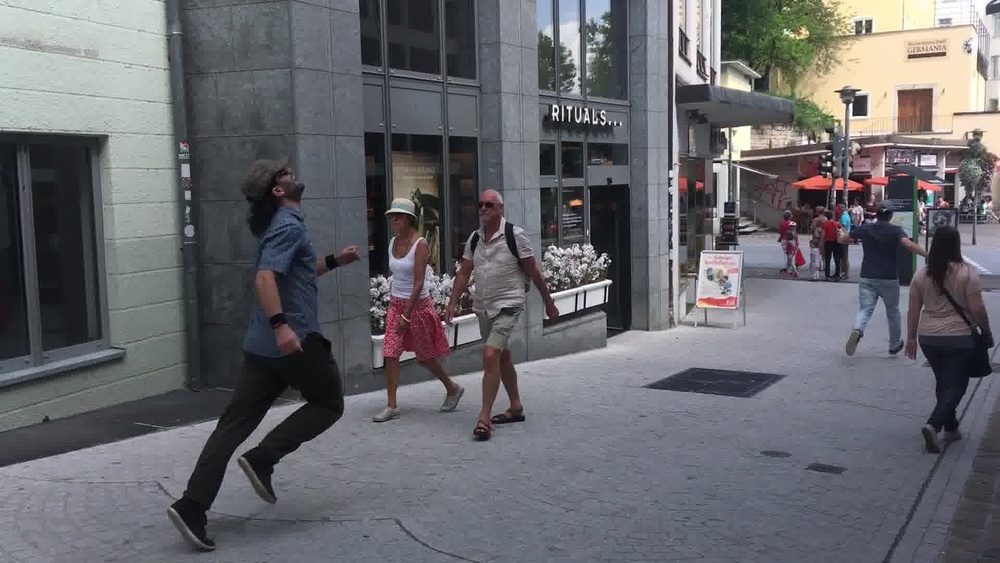}
\includegraphics[width=0.45\linewidth]{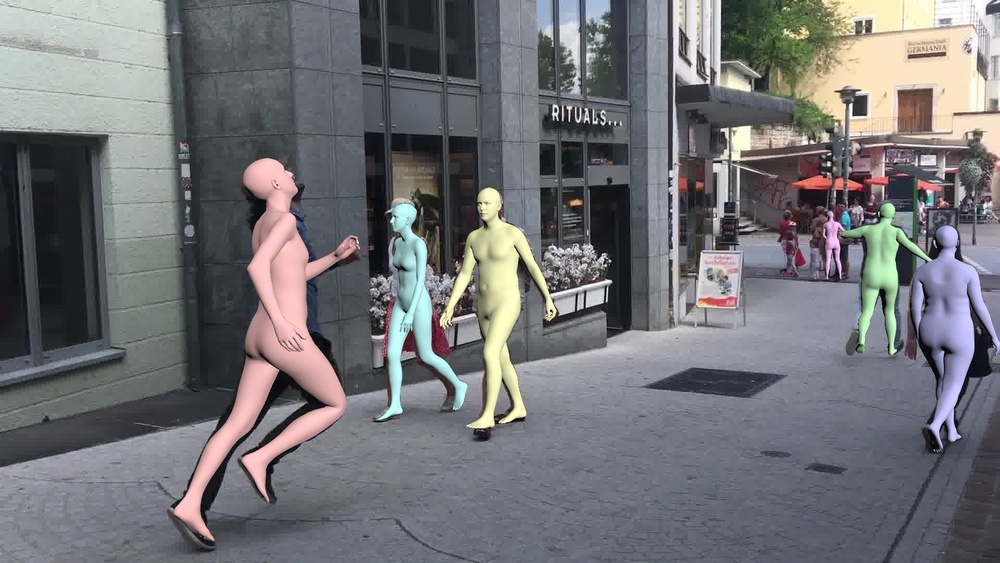}

\includegraphics[width=0.45\linewidth]{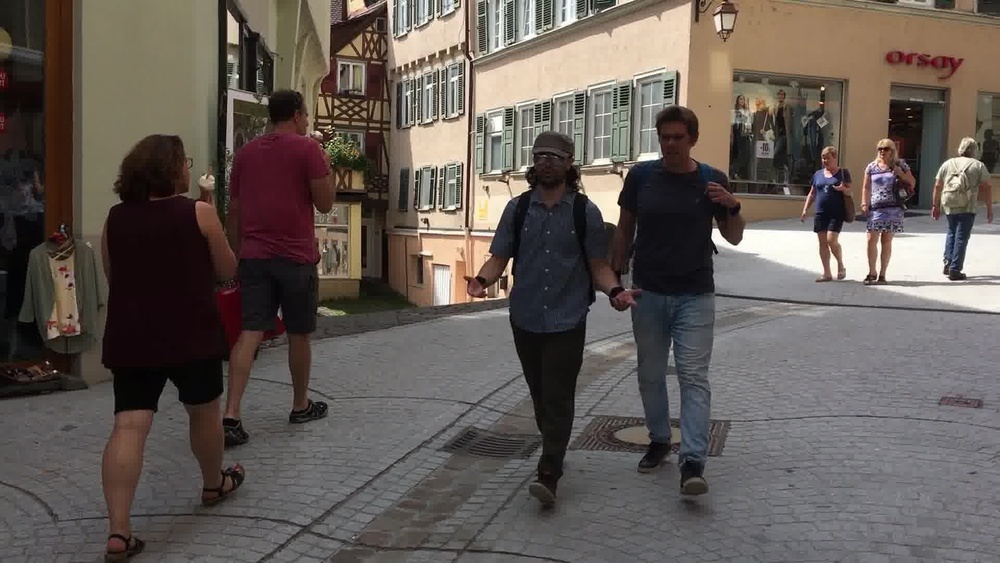}
\includegraphics[width=0.45\linewidth]{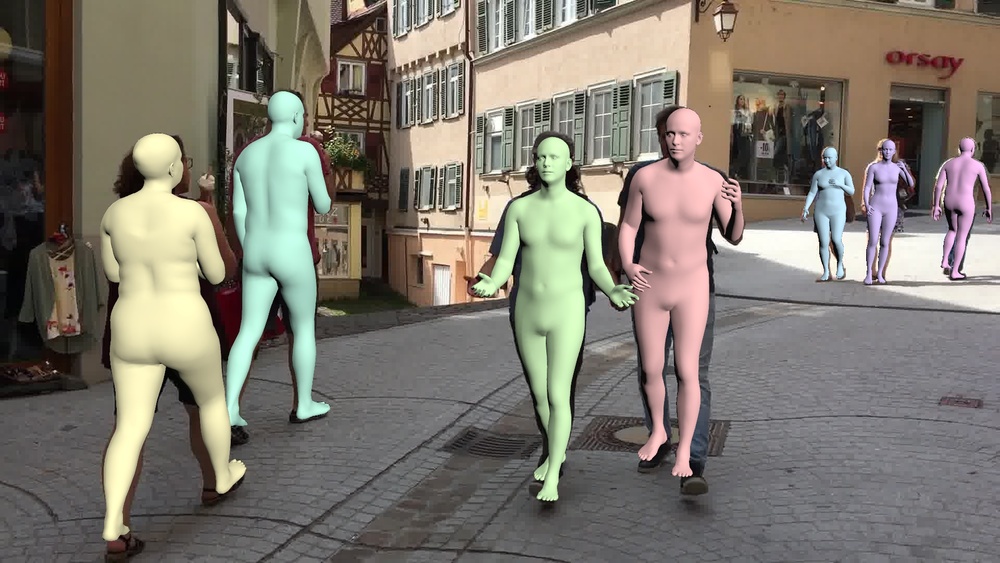}

\includegraphics[width=0.45\linewidth]{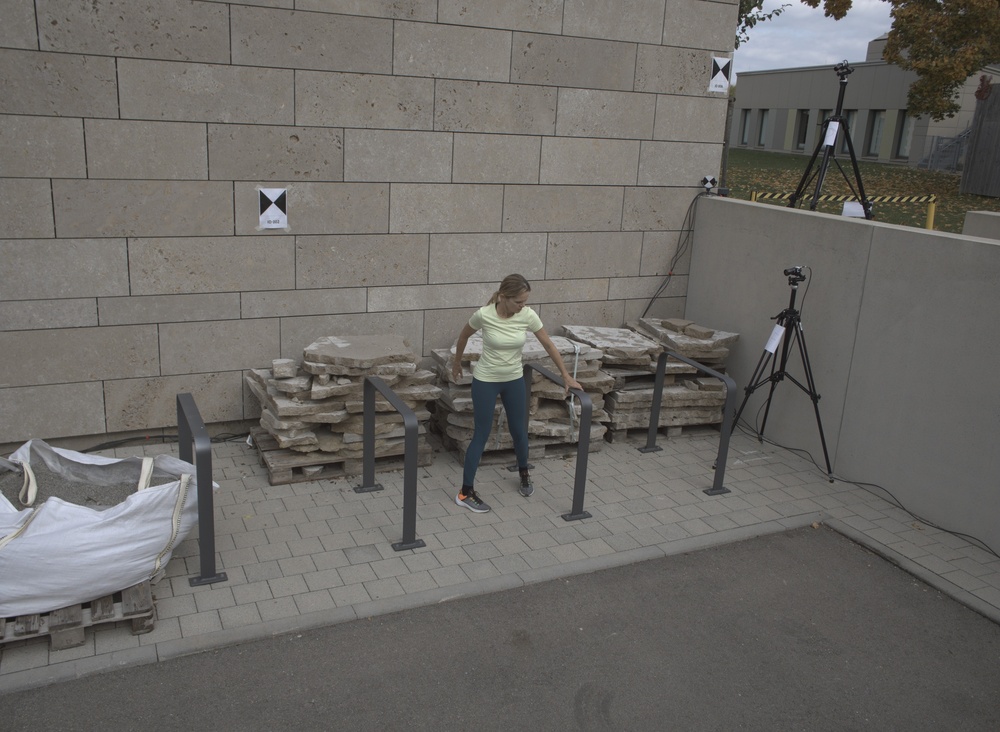}
\includegraphics[width=0.45\linewidth]{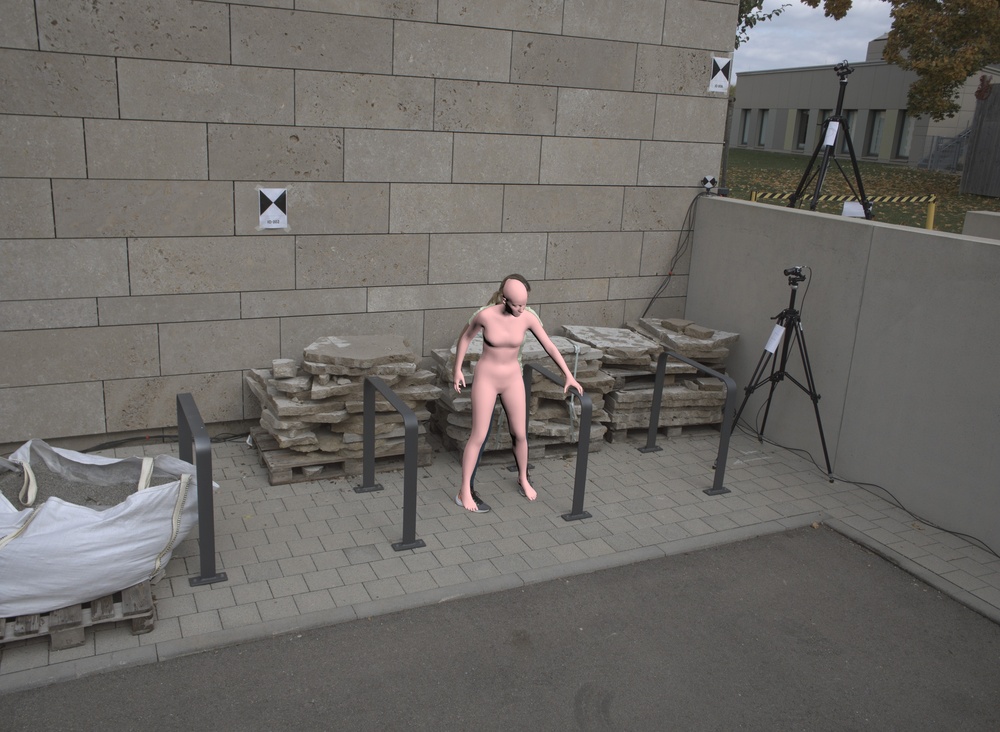}

\includegraphics[width=0.45\linewidth]{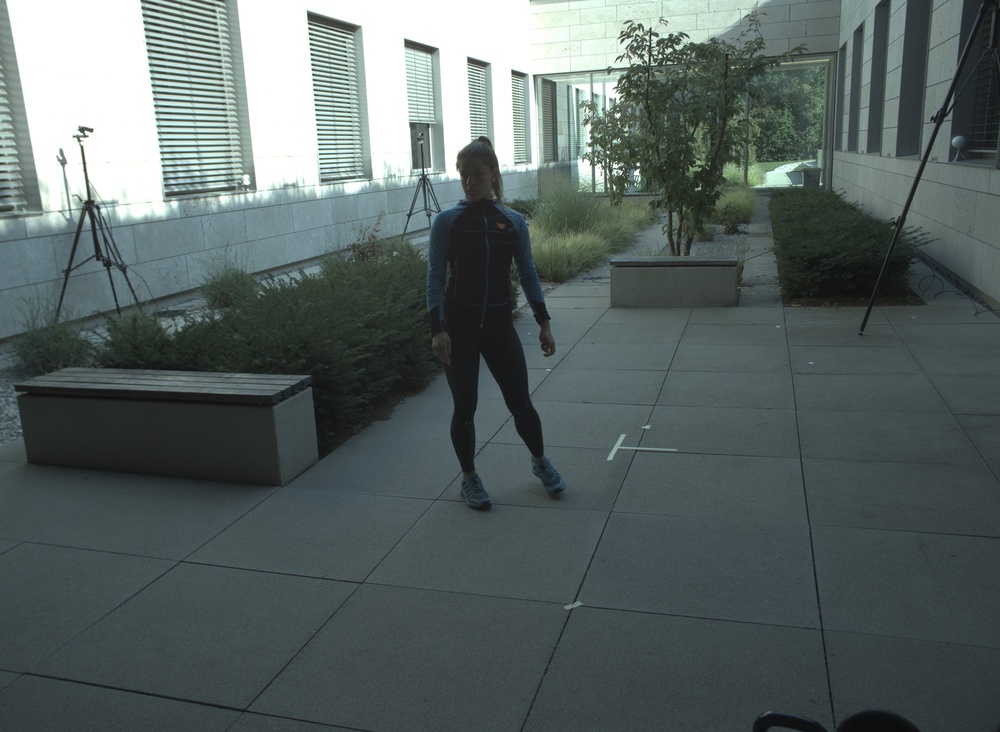}
\includegraphics[width=0.45\linewidth]{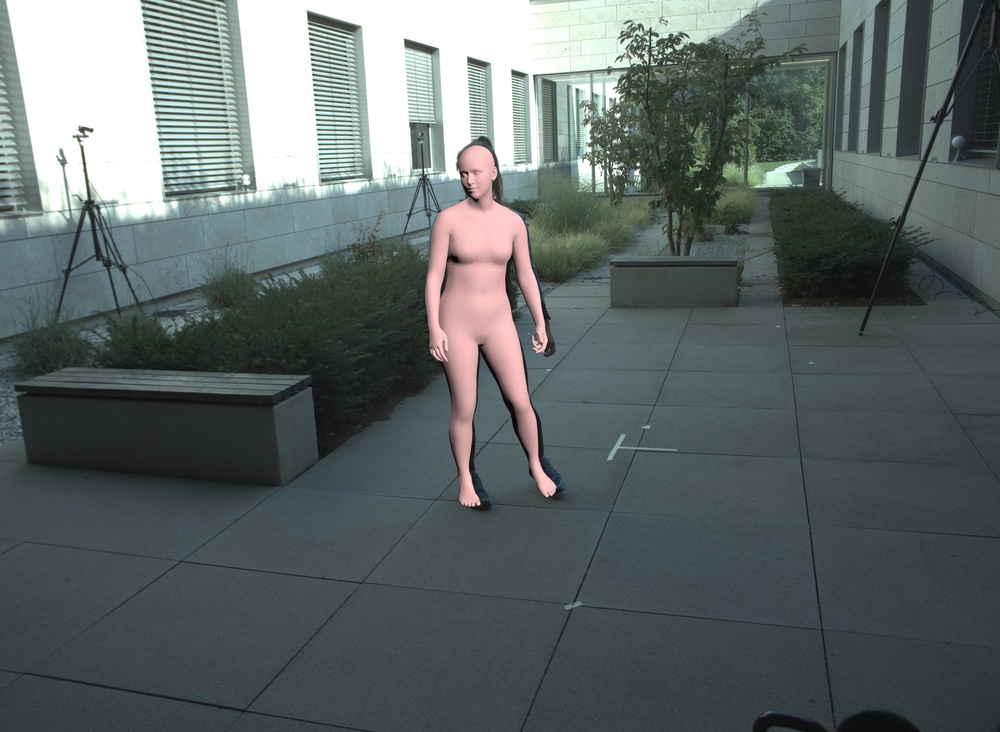}

 \caption{BEDLAM-CLIFF-X results on  3DPW-test (top 2 rows) and RICH-test (bottom 2 rows) images. Note the hand poses and that the body shapes are appropriately gendered.}
    \label{fig:3dpw-rich-qualitative}
\end{figure*}

We also provide qualitative results of BEDLAM-CLIFF-X  on 3DPW and the RICH dataset in \cref{fig:3dpw-rich-qualitative}. 
In this case, we also estimate the \smplx hand poses.
All multi-person results are generated by running the method on individual crops found by a multi-person detector \cite{redmon2018yolov3}.

\newcommand{\figXwidth}{0.37}
\begin{figure*}
\centering
\includegraphics[width=\figXwidth\linewidth]{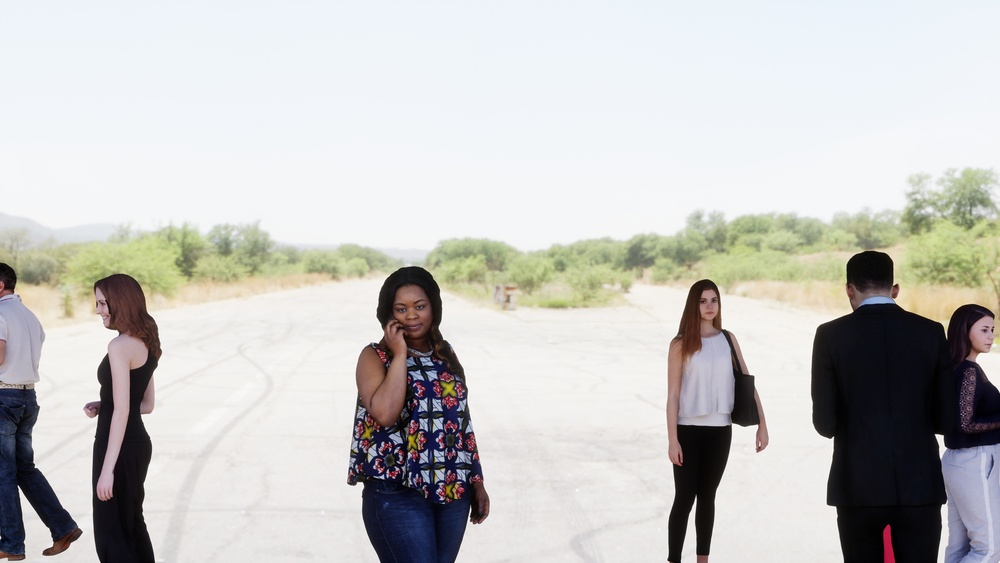}
\includegraphics[width=\figXwidth\linewidth]{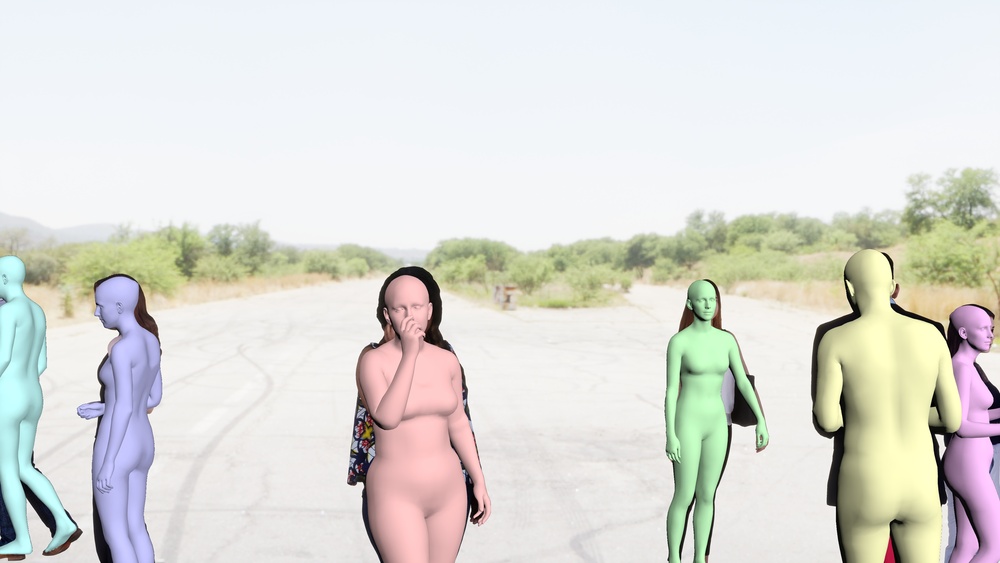}

\includegraphics[width=\figXwidth\linewidth]{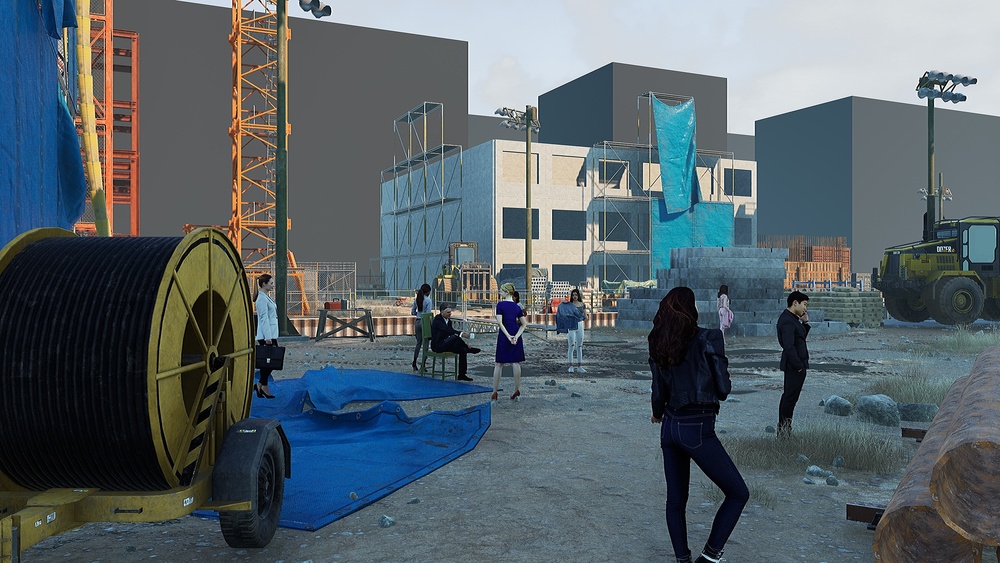}
\includegraphics[width=\figXwidth\linewidth]{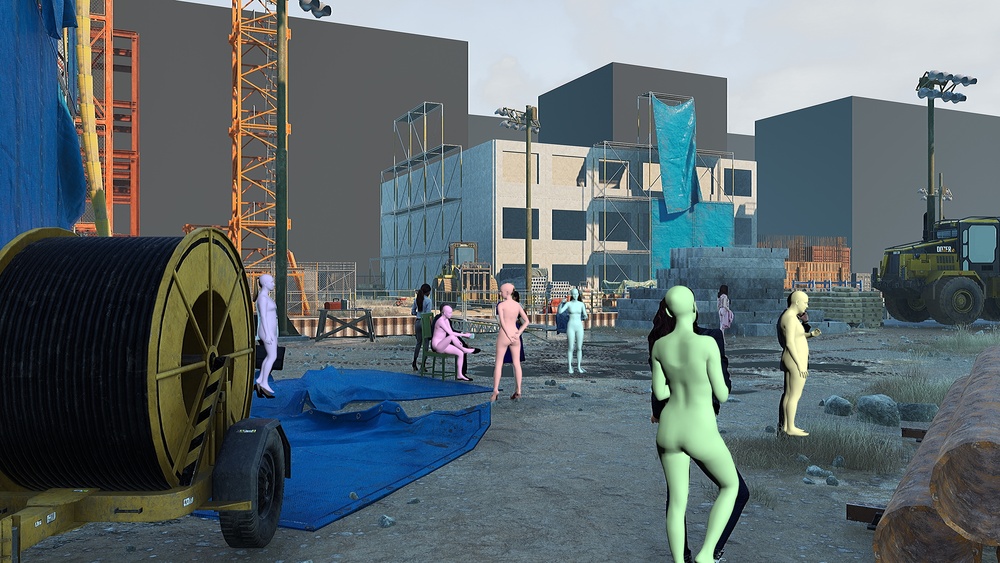}

\includegraphics[width=\figXwidth\linewidth]{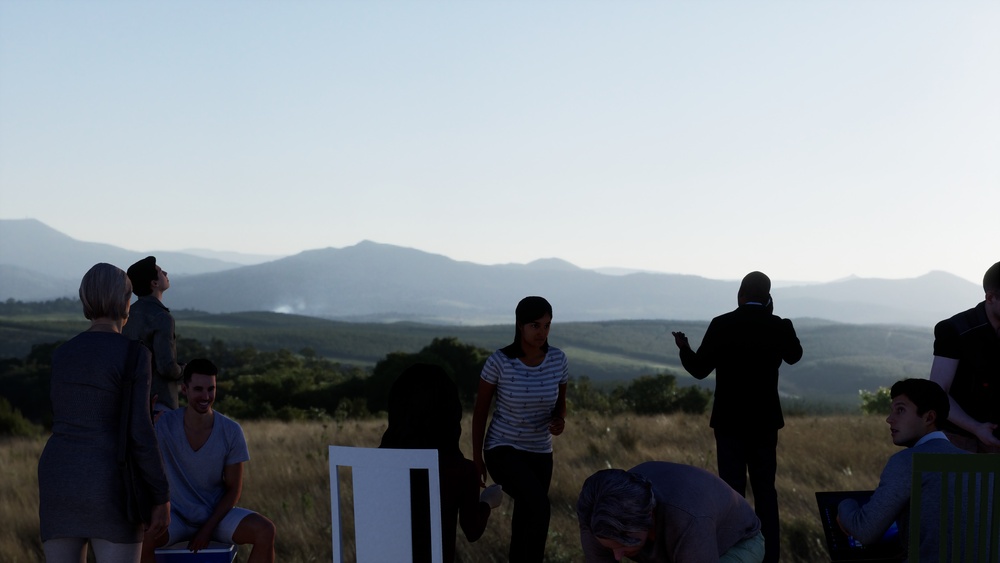}
\includegraphics[width=\figXwidth\linewidth]{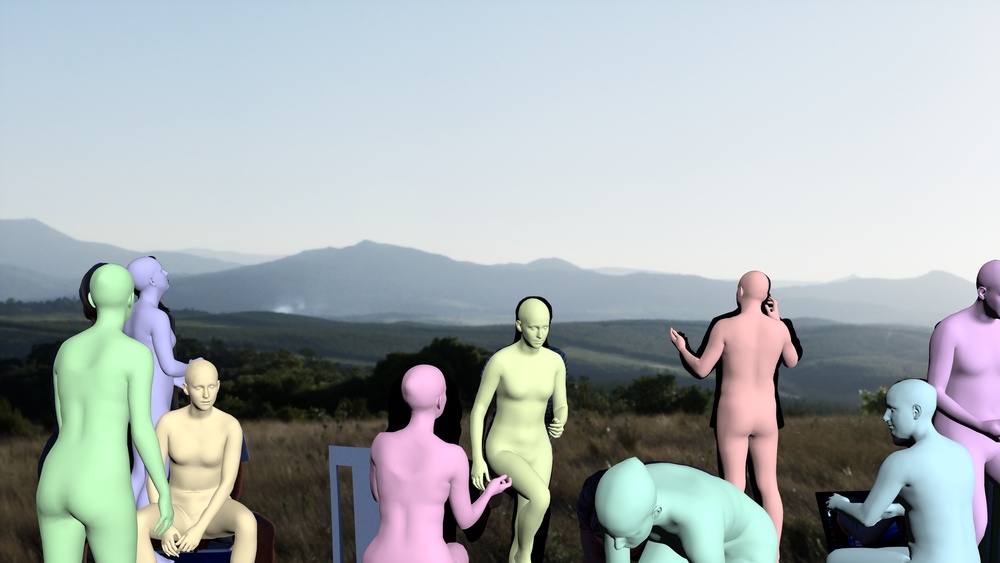}

\includegraphics[width=\figXwidth\linewidth]{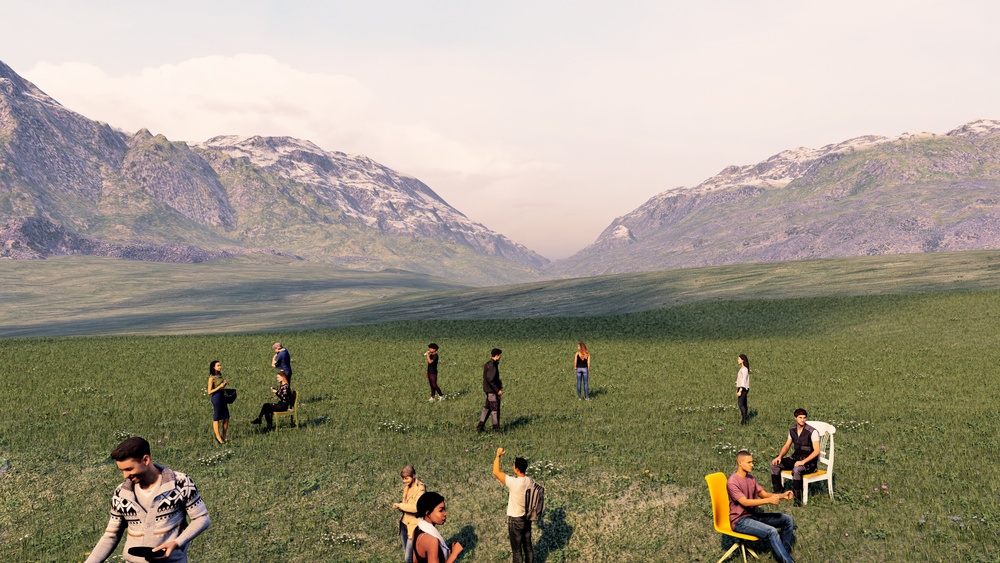}
\includegraphics[width=\figXwidth\linewidth]{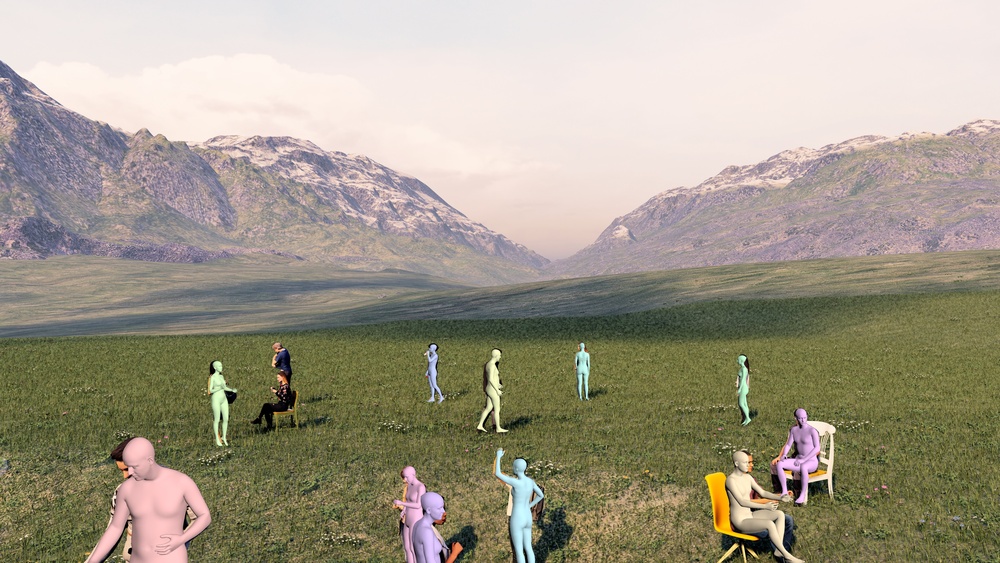}

\includegraphics[width=\figXwidth\linewidth]{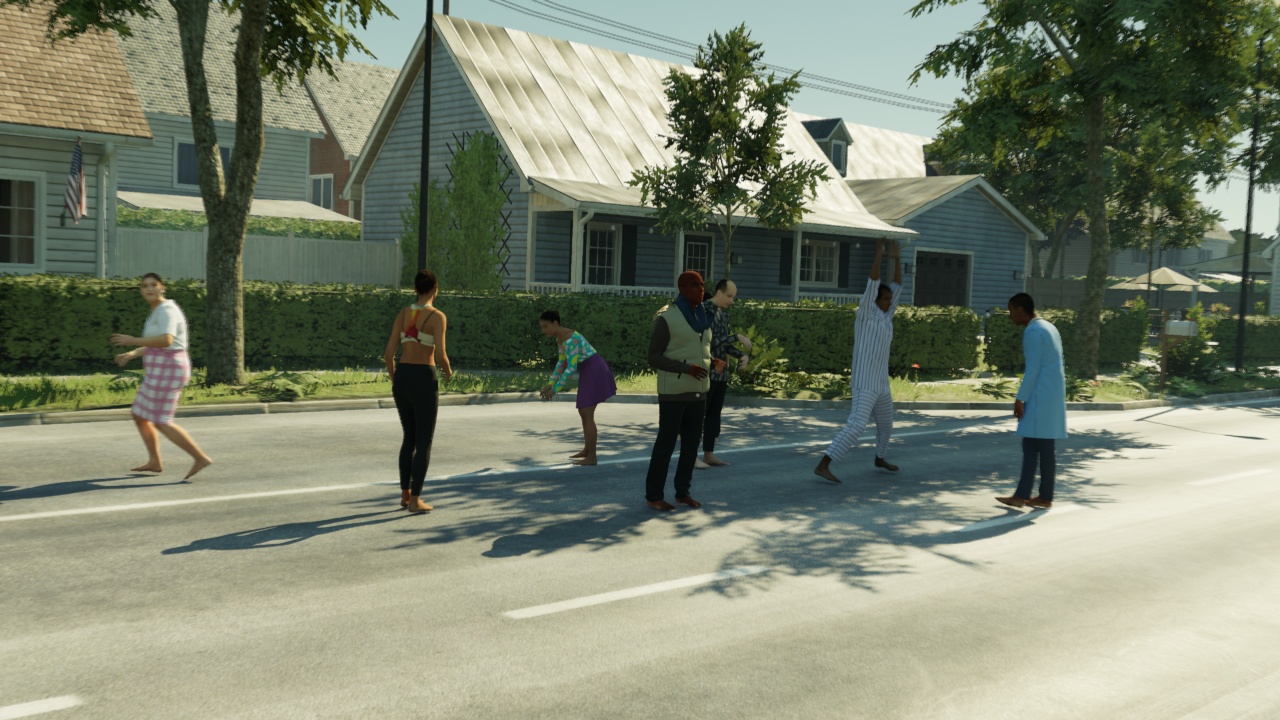}
\includegraphics[width=\figXwidth\linewidth]{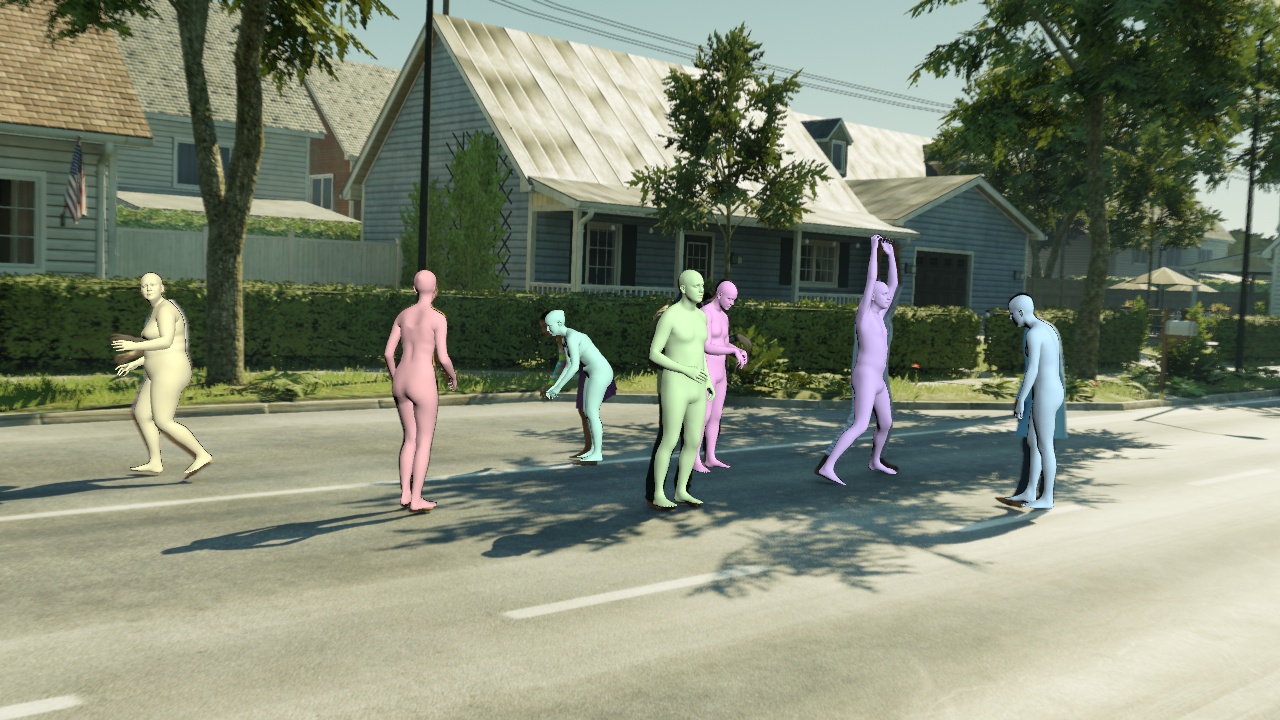}

\includegraphics[width=\figXwidth\linewidth]{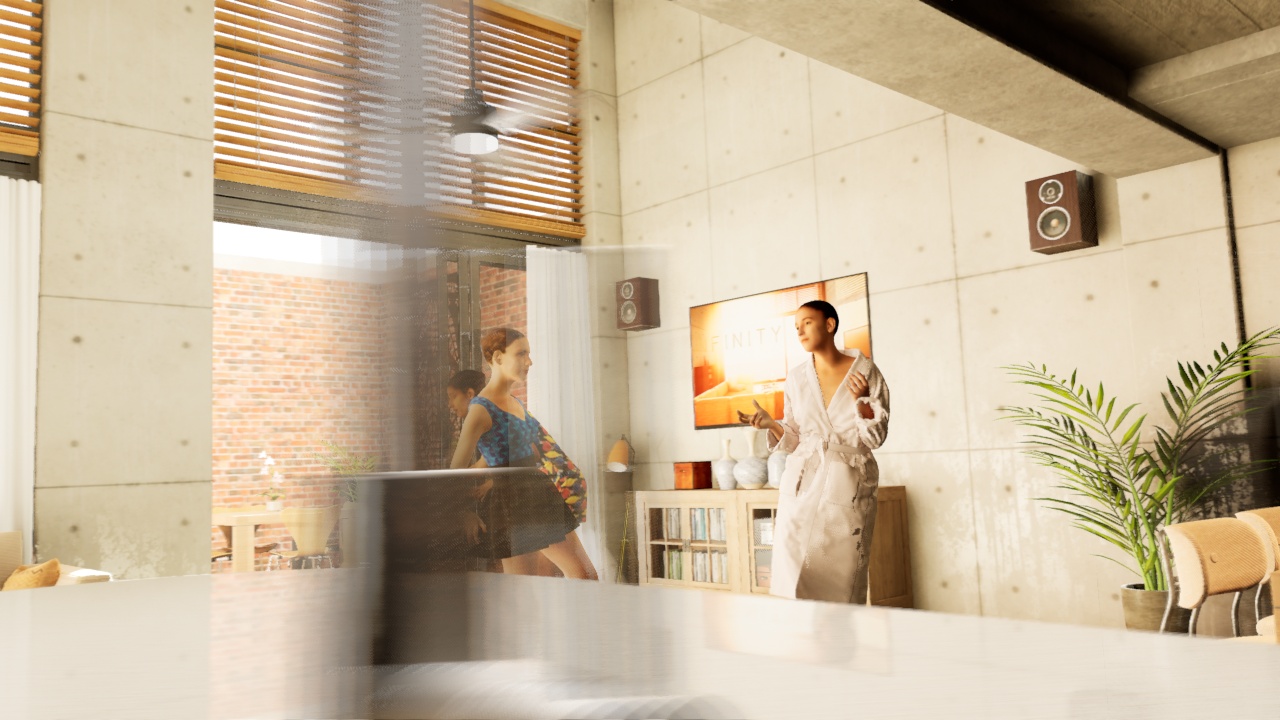}
\includegraphics[width=\figXwidth\linewidth]{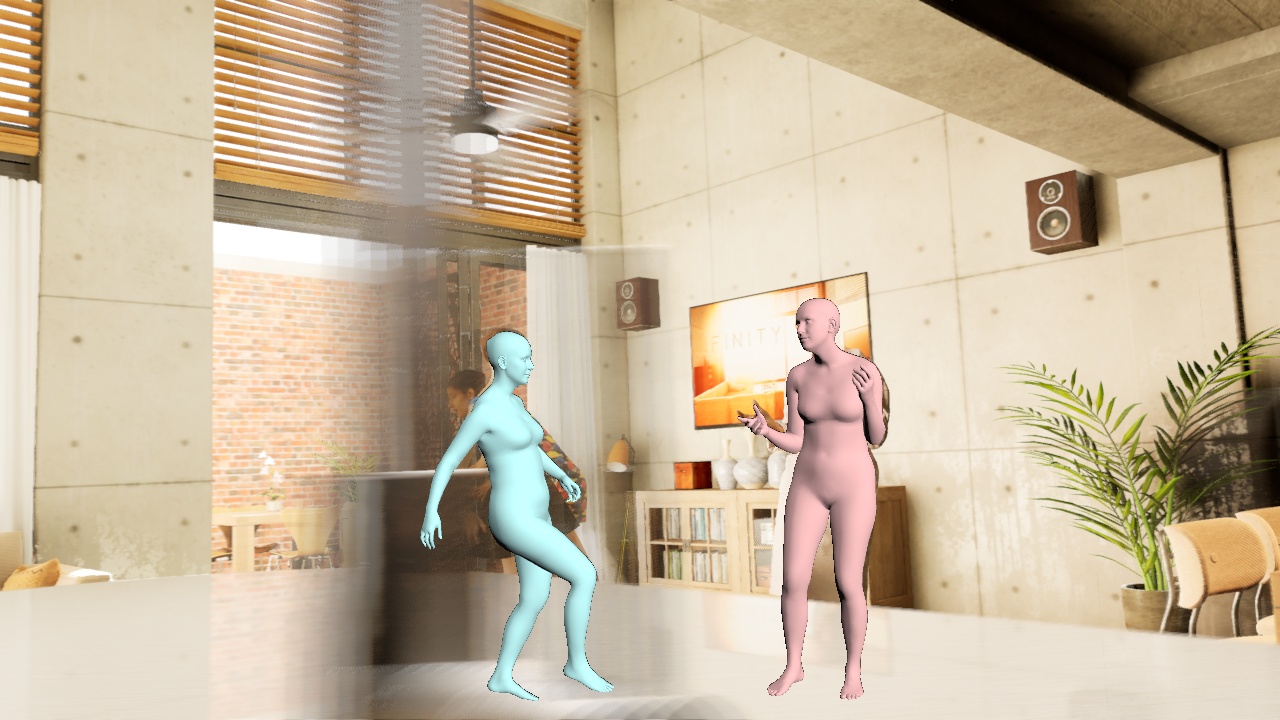}
 \caption{BEDLAM-CLIFF-X results on the AGORA-test (top 4 rows) and the BEDLAM-test images (bottom 2 rows).}
    \label{fig:bedlam-agora-qualitative}
\end{figure*}

\end{document}